\documentclass[11pt]{article}

\usepackage[preprint]{acl}

\usepackage[utf8]{inputenc}
\usepackage[T1]{fontenc}
\usepackage{url}
\usepackage{booktabs}
\usepackage{amsfonts}
\usepackage{amsmath}
\usepackage{amssymb}
\usepackage{nicefrac}
\usepackage{microtype}
\usepackage{xcolor}
\usepackage{graphicx}
\usepackage{algorithm}
\usepackage{algorithmic}
\usepackage{listings}
\usepackage{multirow}
\usepackage{caption}
\usepackage{subcaption}
\usepackage{tikz}
\usepackage{enumitem}

\usetikzlibrary{arrows.meta, positioning, shapes.geometric}
\title{%
  \textbf{PRISM-BN: A Controlled Corpus and Benchmark for Text-to-Parameterized Bayesian Network Extraction}
}

\author{%
  Amartya Bhattacharya$^{1}$, Nikhil Singh$^{1}$,
  Neeti Pokhriyal$^{2}$, Soroush Vosoughi$^{1}$\\
  $^{1}$Dartmouth College \quad $^{2}$RAND Corporation\\
  \texttt{\{amartya.bhattacharya.gr, nikhil.u.singh,}\\
  \texttt{soroush.vosoughi\}@dartmouth.edu}\\
  \texttt{npokhriyal@rand.org}
}
\begin{document}

\maketitle

\begin{abstract}
Probabilistic Graphical Models (PGMs), especially Bayesian Networks (BNs),
expose directed structure and probabilistic parameters, making them natural
symbolic targets for neurosymbolic AI. Yet training
text-to-parameterized-BN systems requires paired text-to-BN resources unavailable
at scale. We introduce \textbf{PRISM-BN}, a controlled corpus of 5{,}054
BN-grounded descriptions paired with discrete reference BNs containing
variables, states, directed edges, root priors, and full multi-parent CPDs
across five domains. The instances are derived from 50 Wikipedia-seeded
backbones, and their probabilities are internally constructed benchmark targets
rather than externally validated causal estimates. PRISM-BN is built with
\textbf{PRISM}, a marginal-first pipeline that elicits marginal and local joint
distributions, analytically recovers normalized CPDs, and constructs locally
reparameterized subgraphs. We define a benchmark with semantic node and state
alignment, conditional structural scoring, and strict full-CPD evaluation.
Across six LLM extractors, Node $F_1$ ranges from 0.56 to 0.83, conditional Edge
$F_1$ from 0.90 to 0.97, and CPD-KL from 1.11 to 3.14. Conditional state and
edge recovery remain consistently strong, whereas strict full-CPD agreement
remains challenging. These trends persist with independently generated GPT-5.5
references, and a human pilot corroborates structural recoverability and
similar probabilistic interpretations. PRISM-BN supports separate evaluation
of structural recovery and probabilistic parameter estimation.

\end{abstract}

\section{Introduction}
\label{sec:intro}

A long-standing goal of neurosymbolic AI is to combine the language
understanding of neural networks with the structured and interpretable
reasoning of symbolic systems~\citep{garcez2023,manhaeve2018deepproblog}.
Probabilistic graphical models (PGMs), and Bayesian Networks (BNs) in
particular, are attractive symbolic targets because they represent directed
dependency structure as a Directed Acyclic Graph (DAG) and attach probability
distributions to each node. This makes model-based inference explicit and
auditable~\citep{pearl1988,koller2009}. A system that reads natural-language
descriptions and outputs fully parameterized BNs would provide a useful
neural-to-symbolic interface for domains where qualitative causal knowledge
and quantitative uncertainty are encoded in text, including economics, policy,
environmental reports, and societal modeling.

Training and evaluating such systems requires paired natural-language
descriptions and fully parameterized BN labels at scale. Existing resources do
not provide this combination. The Bayesian Network Repository~\citep{scutari2010}
contains fewer than 50 curated networks from narrow expert domains, but they
are not paired with text. Recent LLM-based causal extraction methods
\citep{long2023,kiciman2023,ban2023} recover qualitative DAGs, but leave CPDs
unspecified. Conversely, work on eliciting probabilities from LLMs reports
persistent failures, including base-rate neglect, coarse numeric estimates,
prompt sensitivity, and violations of normalization constraints
\citep{wang2025always,nafar2025extracting}. The field therefore lacks a
benchmark that jointly evaluates recovery of variables, states, directed
structure, and probabilistic parameters from language.

We introduce PRISM-BN, a controlled corpus for
text-to-parameterized-BN extraction, and \textbf{PRISM}, the construction
pipeline used to produce it. PRISM-BN contains 5{,}054 natural-language
descriptions paired with fully parameterized discrete reference BNs across five
domains. These examples are derived from 50 larger Wikipedia-seeded source BNs,
which we call backbones. Each released example includes variables, states,
directed edges, root priors, full CPDs, Wikipedia provenance, and a generated
BN-grounded description checked for coverage of variables, non-null states, and
causal edges. The benchmark input is the generated description paired with the
resulting reference BN. PRISM-BN therefore targets controlled extraction rather
than open-ended causal discovery. Its labels are internally consistent benchmark
targets, not externally validated estimates of real-world causal effects.

The core technical idea in PRISM is a \emph{marginal-first} parameterization
strategy. Instead of asking an LLM to produce CPDs directly, PRISM elicits
lower-dimensional probability objects, including unconditional marginals
$P(\text{node}{=}s)$ and local joint distributions over connected variables.
CPDs are then recovered analytically by normalization. For a single-parent edge
$A \rightarrow B$, PRISM computes
\[
P(B{=}b \mid A{=}a) =
\frac{P(A{=}a,B{=}b)}
     {\sum_{b'} P(A{=}a,B{=}b')}.
\]

This separates probability elicitation from probability-table construction.
The LLM provides marginal and joint tables, while the pipeline enforces local
normalization programmatically.

Starting from the 50 backbones, PRISM-BN uses a subgraph re-rooting procedure
to produce the released examples. Connected subgraphs are extracted from each
backbone, retained edges are reparameterized from cached marginal and joint
tables, and nodes whose parents were cut away become roots with cached
unconditional priors. Each induced subgraph is paired with a fresh generated
description, yielding self-contained text--BN examples rather than fragments
with dangling dependencies.

We evaluate six language models on PRISM-BN using a four-phase benchmark:
semantic node alignment, state alignment, directed-edge scoring, and CPD
scoring. State, edge, and CPD metrics are conditional on successful earlier
alignment steps and should not be interpreted as unconditional end-to-end
accuracy. Across the six generators, node $F_1$ ranges from 0.56 to 0.83, with
near-ceiling recall (0.97--0.99) but substantially lower precision (0.40--0.71),
showing that node over-generation is the dominant structural failure mode.
Conditional state $F_1$ ranges from 0.89 to 0.93, conditional edge $F_1$ from
0.90 to 0.97, and CPD-KL from 1.11 to 3.14. All evaluated LLMs underperform
simple calibration reference predictors for CPD estimation, and the best model
on node $F_1$ is not the best model on CPD-KL. These results show that
structural extraction and probabilistic calibration are decoupled.

Our contributions are:
\begin{itemize}[leftmargin=*, nosep, topsep=0pt, partopsep=0pt, parsep=0pt, itemsep=0pt]

    \item \textbf{Dataset.} PRISM-BN, a corpus of 5{,}054 paired
    natural-language descriptions and fully parameterized discrete reference BNs
    across five domains, derived from 50 Wikipedia-seeded source backbones.

    \item \textbf{Construction pipeline.} PRISM, a marginal-first pipeline that
    elicits lower-dimensional marginal and joint distributions and recovers
    normalized CPDs analytically, avoiding direct CPD elicitation.

    \item \textbf{Subgraph expansion.} A re-rooting and CPD-recomputation
    procedure that converts larger source backbones into smaller,
    self-contained text--BN examples.

    \item \textbf{Benchmark.} A four-phase text-to-BN evaluation protocol with
    semantic node and state alignment, conditional structure scoring, and
    CPD-KL evaluation.
\end{itemize}

PRISM-BN is intended as a controlled benchmark for translating natural-language
descriptions into symbolic probabilistic models, not as a factual causal
knowledge base. Its structures and probabilities are internally consistent
reference labels produced by a Wikipedia-seeded construction pipeline.
Accordingly, CPD-KL measures agreement with the PRISM-BN reference
parameterization, not accuracy against external observational or interventional
data.

\section{Related Work}
\label{sec:related}
A growing line of work shows that LLMs contain useful causal knowledge that can
be turned into graph structures. \citet{kiciman2023} report that LLMs match or
exceed classical discovery algorithms on standard pairwise causal benchmarks,
and \citet{long2023} show that GPT-3 can act as a useful auxiliary signal when
scoring edges in medical causal graphs. \citet{ban2023} use LLM priors inside
score-based search. PromptBN \citep{zhang2025promptbn} extracts full DAGs from
variable metadata using meta-prompting, and its ReActBN extension combines LLM
priors with empirical feedback. CausalGraphBench \citep{babakov2025causalgraphbench}
finds that accuracy drops on larger graphs, in part because exhaustive querying
can overemphasize local edges and introduce hallucinated dependencies. All of
this work is qualitative: it extracts edges but assigns no probabilities.
PRISM-BN adds the missing parametric layer, pairing each reference structure
with the CPDs needed for probabilistic inference. Asking LLMs for CPDs directly
exposes well-known failure modes, including bias toward round numbers and
inconsistency under rephrasing~\citep{wang2025always}. EPK
\citep{nafar2025extracting} splits multidimensional CPD queries into
single-state prompts and renormalizes programmatically, while
\citet{wang2025always} train a decoder-based regression model to suppress
numerical bias. PRISM instead avoids direct CPD queries, asking for
lower-dimensional marginal and local joint distributions and recovering CPDs
analytically. Several benchmarks test LLM probabilistic reasoning, but none
provides what is needed to train and evaluate text-to-BN extraction. CLADDER
\citep{jin2023cladder} tests binary reasoning across Pearl's ladder. BLInD
\citep{nafar2025blind} requires exact numeric probabilities over abstract dummy
events. QUITE \citep{schrader2024quite} uses more naturalistic multi-state
networks and Bayesian inference questions. All three test reasoning over
networks that are already given. PRISM-BN fills this gap by providing generated
BN-grounded descriptions together with reference structure and parameters for
end-to-end text-to-BN benchmarking.

\section{The PRISM Pipeline}
\label{sec:pipeline}

\begin{figure*}[t]
    \centering
    \includegraphics[width=.95\textwidth]{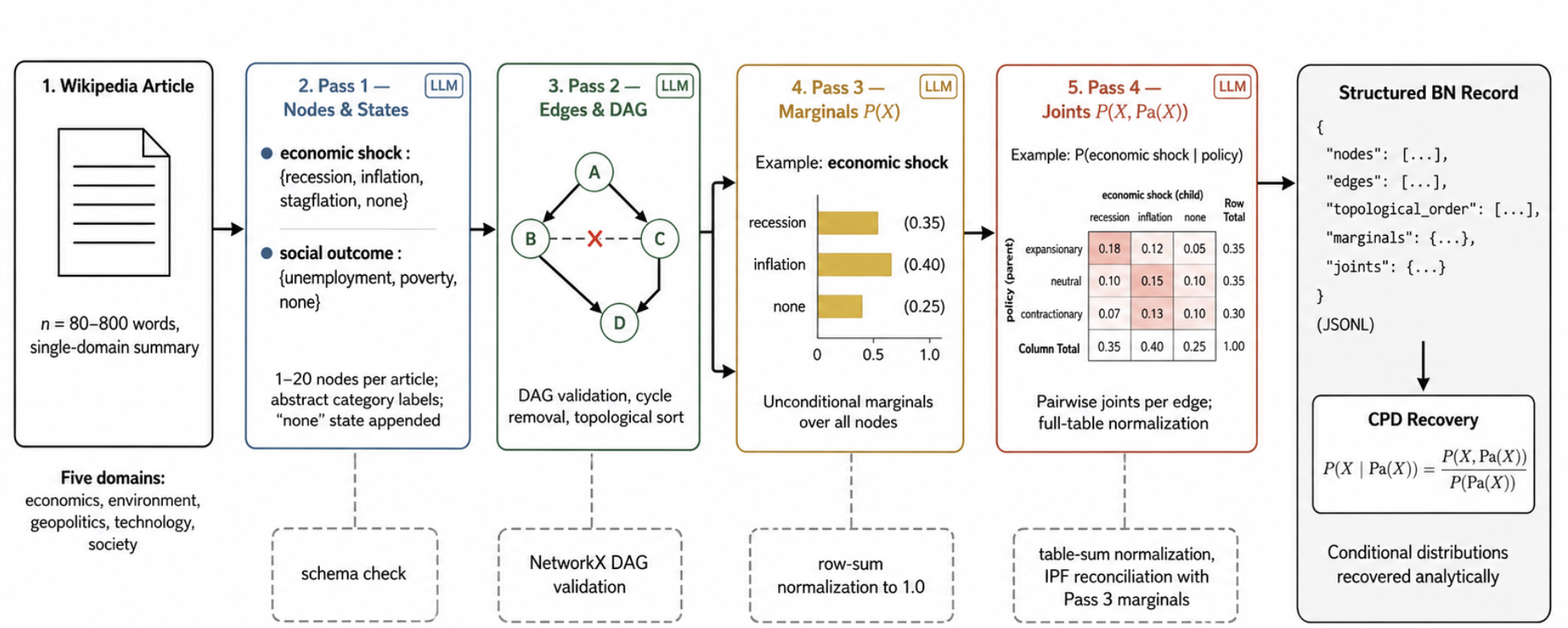}
    \caption{The PRISM marginal-first construction pipeline. Five LLM passes elicit
    nodes and states, edges and DAG, unconditional marginals, pairwise joints,
    and multi-parent joints. Cached marginals and joints are combined to recover
    locally normalized CPDs analytically. Schema and DAG validation are applied
    between passes.}
    \label{fig:pipeline}
\end{figure*}

PRISM converts Wikipedia articles into fully parameterized reference BNs through
five LLM passes, analytic CPD recovery, and coverage-checked natural-language
generation. Figure~\ref{fig:pipeline} shows the data flow. The goal is to
construct internally consistent reference BNs paired with controlled
descriptions. Extraction accuracy from text is evaluated separately in
\S\ref{sec:results}.

\subsection{Backbone Extraction}
\label{sec:extraction}

We define five domains, each seeded with five Wikipedia search categories
chosen to target causally rich articles, such as ``Economic crises'' and
``Deforestation.'' Each retained article is processed by Claude
\texttt{claude-sonnet-4-6} in five passes: node and state extraction, edge
identification with DAG validation, unconditional marginal estimation, pairwise
joint estimation for each edge, and multi-parent joint estimation for children
with two or three parents. The node pass extracts 1--20 candidate causal nodes,
each with 2--6 discrete states plus a mandatory \texttt{none} baseline state.
The edge pass uses NetworkX~\citep{hagberg2008} to remove cycle-inducing edges
as they are added. Marginal and joint tables are renormalized after elicitation,
and the multi-parent joint elicitation (Pass 5) is skipped for nodes with four
or more parents; CPDs for these high-fan-in nodes are instead recovered via the
analytical fallback described in \S\ref{sec:cpd}. The
marginal-first decomposition separates probability elicitation from
probability-table construction: the LLM supplies lower-dimensional marginals
and joints, while the pipeline enforces local normalization analytically.

\subsection{CPD Recovery}
\label{sec:cpd}

\textbf{Single-parent nodes.} CPDs are recovered analytically from pairwise
joint tables via:
\begin{equation}
    P(B{=}b \mid A{=}a) = \frac{P(A{=}a, B{=}b)}{P(A{=}a)},
    \label{eq:bayes_cpd_single}
\end{equation}
where $P(A{=}a)=\sum_b P(A{=}a,B{=}b)$ is the marginal recovered by summing
the pairwise joint over child states.

\textbf{Root nodes.} Root nodes use the Pass-3 marginal directly as their
prior distribution, followed by renormalization.

\textbf{Multi-parent nodes (2--3 parents).} For a child $C$ with parents
$\{A_1,\ldots,A_k\}$ where $k\in\{2,3\}$, the full joint
$P(C{=}c,A_1{=}a_1,\ldots,A_k{=}a_k)$ is elicited directly in Pass 5. The
conditional is then recovered by normalizing the elicited joint:
\begin{equation}
\begin{aligned}
    &P(C{=}c \mid A_1{=}a_1, \ldots, A_k{=}a_k) \\
    &\quad =
    \frac{P(C{=}c,A_1{=}a_1,\ldots,A_k{=}a_k)}
         {\sum_{c'} P(C{=}c',A_1{=}a_1,\ldots,A_k{=}a_k)},
\end{aligned}
\label{eq:multi_cpd}
\end{equation}
making no conditional independence assumptions among the parents. This is
exact with respect to the elicited joint.

For nodes with $k\geq4$ parents, the full joint table contains
$\prod_{i=1}^{k}|S_{A_i}|\times |S_C|$ entries, which grows exponentially in
$k$. Enumerating all entries reliably in a single LLM call becomes both
expensive and error-prone at this scale. We therefore apply a Naive Bayes
factorization as a fallback:
\begin{equation}
\begin{aligned}
    &P(C{=}c \mid A_1{=}a_1, \ldots, A_k{=}a_k) \\
    &\quad \propto
    \frac{\prod_{i=1}^{k} P(C{=}c \mid A_i{=}a_i)}{P(C{=}c)^{k-1}},
\end{aligned}
\label{eq:naive_bayes}
\end{equation}
where each pairwise conditional is recovered from Pass-4 joint tables via
Eq.~\ref{eq:bayes_cpd_single} and the marginal $P(C{=}c)$ is taken from
Pass~3. In practice, only around 0.1\% of the 5{,}054 subgraphs in PRISM-BN
contain a node with four or more parents.

\subsection{From Backbones to Dataset}
\label{sec:subgraph}

The 50 backbones form the basis of the released dataset. For a configurable
size range $[k_{\min}, k_{\max}]$, we enumerate connected node subsets of each
backbone via joint BFS and DFS traversals with deduplication. Given a sampled
subset $S$, edges with both endpoints in $S$ are retained, and nodes whose
parents were cut away become roots with priors taken from the Pass-3 marginals
of their backbone. CPDs for non-root nodes are recomputed from cached marginal
and joint tables using the recovery rules of \S\ref{sec:cpd}. Topological
levels are recalculated, and each subgraph is paired with a freshly generated
natural-language description via the coverage-checked NLG procedure of
\S\ref{sec:nlg}.

\begin{figure*}[t]
    \centering
    \includegraphics[width=.95\textwidth]{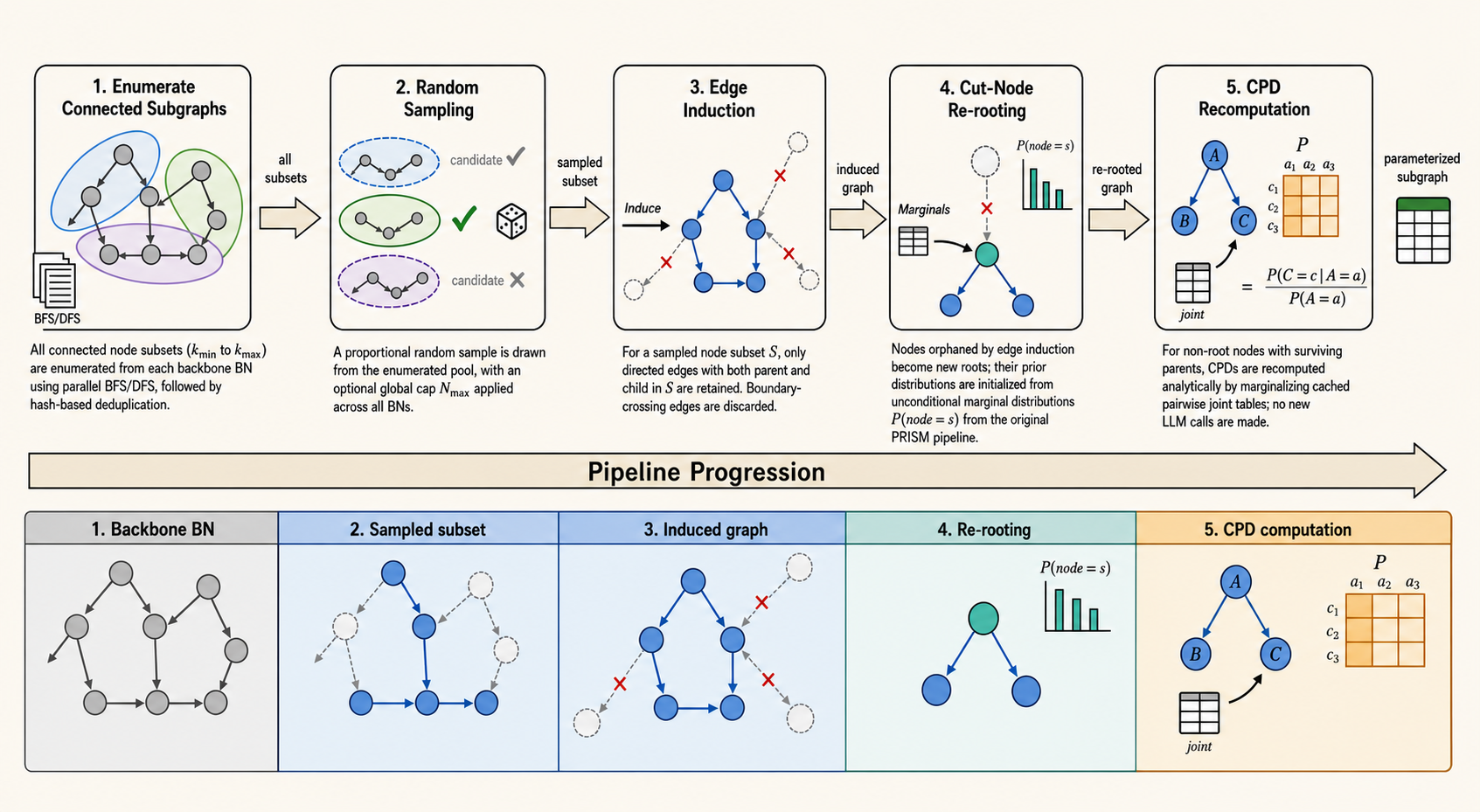}
    \caption{From backbone to released examples: (1) enumeration, (2) random
    sampling, (3) edge induction, (4) cut-node re-rooting using marginal priors,
    and (5) CPD recomputation via cached joint marginalization. Each resulting
    BN is paired with a fresh natural-language description.}
    \label{fig:subgraph-pipeline}
\end{figure*}

\begin{table}[t]
\centering
\small
\caption{Structural comparison of the 50 source backbones and the 5{,}054
released paired examples (means).}
\label{tab:comparison}
\begin{tabular}{lrr}
\toprule
\textbf{Metric} & \textbf{Backbones} & \textbf{Released} \\
\midrule
\# graphs                 &     50 &  5{,}054 \\
\# node records           &    584 & 38{,}820 \\
Mean nodes                &  11.68 &     7.68 \\
Mean edges                &  14.26 &     7.50 \\
Mean depth                &   5.04 &     3.63 \\
Mean avg in-/out-deg.     &   1.16 &     0.97 \\
Mean avg states           &   3.75 &     3.85 \\
Mean text len.\ (words)   & 480.84 &   349.30 \\
\bottomrule
\end{tabular}
\end{table}

\subsection{Natural Language Generation from PGMs}
\label{sec:nlg}

Each parameterized BN is paired with a generated description (generated by Claude Sonnet 4.6). Generation is
subject to three coverage and consistency constraints checked programmatically
after every attempt: every node name and non-\texttt{none} state must appear,
every causal connection must be described with explicit causal language, and
language strength must reflect CPD magnitudes. Generation is retried up to
three times, with violated constraints fed back as explicit feedback. The
generated description is therefore a BN-grounded verbalization rather than a
full table-level transcription of every CPD entry. A random sample of 344
text--BN pairs was manually verified to confirm that the automatic coverage
constraints held in practice. The remaining 4{,}710 pairs were verified
automatically using Claude Opus 4.6 as a checker, with semantically equivalent
node names treated as matches.

\section{Dataset Characteristics}
\label{sec:dataset_characteristics}

PRISM-BN consists of \textbf{5{,}054 paired text--BN examples}, derived from 50
source backbone BNs. Each example is self-contained, with its own generated
description and its own re-rooted priors and CPDs. Table~\ref{tab:comparison}
summarizes the structural differences between the 50 source backbones and the
released corpus. All statistics in this section are computed over the 5{,}054
released examples, which contain 38{,}820 node records in total. Because multiple released examples can originate from the same source backbone,
we release the backbone identifier for every example. Supervised evaluations
should use backbone-disjoint splits to avoid leakage across related subgraphs.
The zero-shot benchmark in \S\ref{sec:results} uses the full corpus because no
model is trained on PRISM-BN.

The released corpus targets moderately deep, sparsely connected causal motifs:
mean node count 7.68 ($\sigma = 1.46$), mean edge count 7.50
($\sigma = 1.99$), mean depth 3.63 ($\sigma = 1.13$), mean in-/out-degree
0.97, and mean state cardinality 3.85. Examples of 7--9 nodes account for
81.0\% of the corpus. Full structural distributions appear in
Figure~\ref{fig:subgraph_overview}, and full text-length distributions appear
in Appendix~\ref{app:dataset_figures}.

\begin{figure*}[t]
    \centering
    \includegraphics[width=\textwidth]{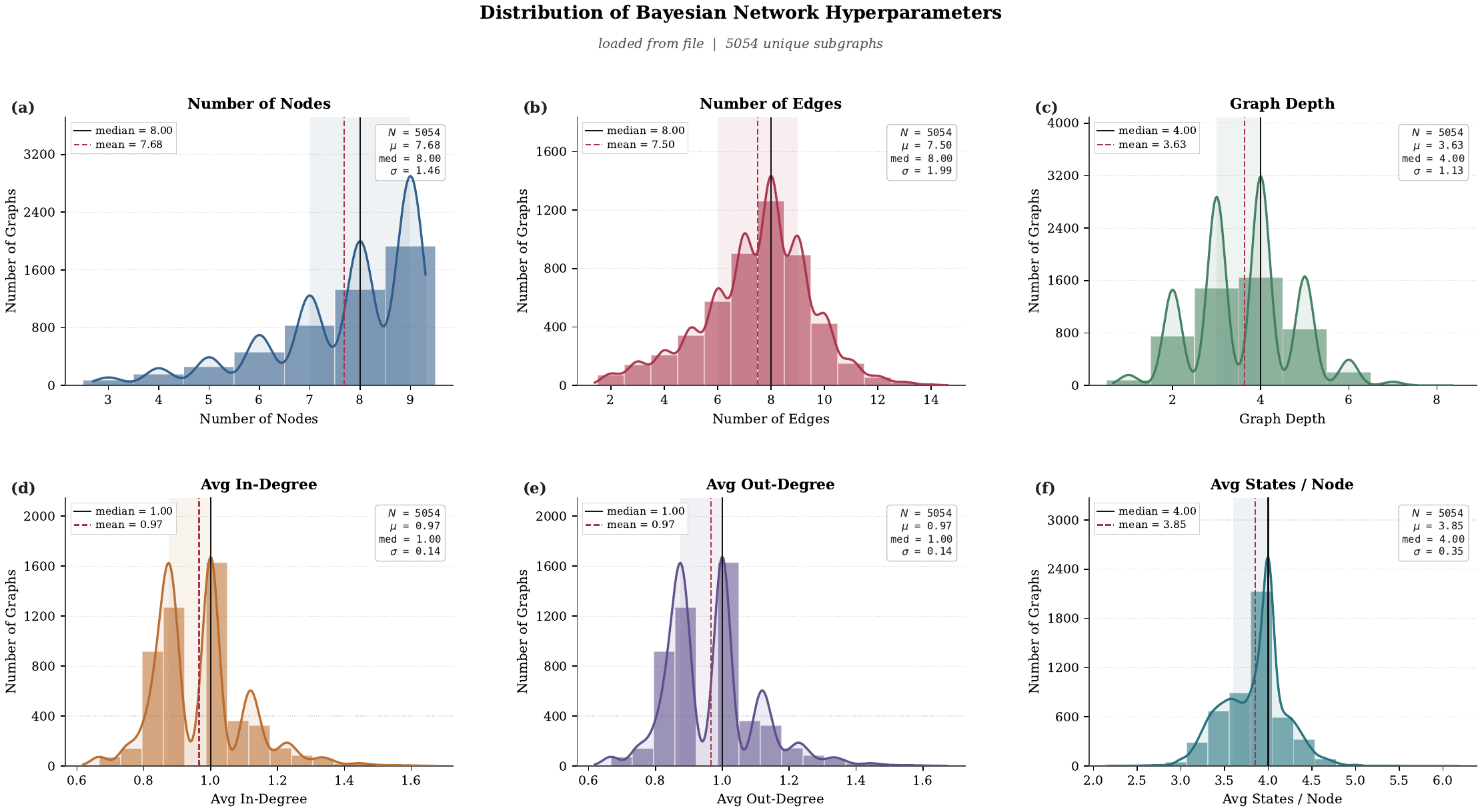}
    \caption{Structural distributions across the 5{,}054 released subgraphs in
    PRISM-BN, generated by the subgraph extraction pipeline
    (Figure~\ref{fig:subgraph-pipeline}) applied to the 50 backbone networks.
    Panels show histograms with kernel density estimates for (a) number of
    nodes, (b) number of edges, (c) graph depth, (d) average in-degree,
    (e) average out-degree, and (f) average states per node.}
    \label{fig:subgraph_overview}
\end{figure*}

\textbf{Domain-level structure.}
Per-domain mean statistics are reported in Table~\ref{tab:domain_subgraph}.
Domain coverage is uneven by construction: Geopolitics (2{,}043) and
Environment (1{,}711) together account for 74.3\% of the corpus, followed by
Society (626), Economics (586), and Technology (88). Despite this imbalance,
structural means among the four largest domains are homogeneous: mean node
count varies only between 7.46 and 7.85, and mean in- and out-degree sits in a
0.92--0.99 band. Technology is the structural outlier on every axis, reflecting
smaller and shallower parent backbones. The largest domain-level effect on
parameterization is state cardinality: Geopolitics has the richest state spaces
(mean 3.99 states/node), while Technology and Environment have the leanest
(3.65 and 3.74).

\begin{table}[t]
\centering
\small
\setlength{\tabcolsep}{3pt}
\caption{Domain-level statistics for the 5{,}054 paired examples (mean values,
computed from the JSON skeleton).}
\label{tab:domain_subgraph}
\begin{tabular}{lrcccc}
\toprule
\textbf{Domain} & \textbf{\#Ex.} & \textbf{Nodes} & \textbf{Edges}
    & \textbf{Depth} & \textbf{States} \\
\midrule
Economics   &   586 & 7.46 & 7.22 & 3.65 & 3.73 \\
Environment & 1{,}711 & 7.70 & 7.68 & 3.45 & 3.74 \\
Geopolitics & 2{,}043 & 7.85 & 7.69 & 3.86 & 3.99 \\
Society     &   626 & 7.65 & 7.08 & 3.48 & 3.85 \\
Technology  &    88 & 5.23 & 4.59 & 2.61 & 3.65 \\
\bottomrule
\end{tabular}
\end{table}

\section{The Text-to-BN Task and Evaluation Pipeline}
\label{sec:eval}

\textbf{Task formulation.}
Let $\mathcal{T}$ be a natural-language description of a BN. In PRISM-BN,
$\mathcal{T}$ is a generated BN-grounded description paired with a reference
parameterized BN. A model $\mathcal{M}$ reads $\mathcal{T}$ and outputs
$(\hat{\mathcal{G}}, \hat{\mathcal{P}})$, where
$\hat{\mathcal{G}} = (\hat{\mathcal{V}}, \hat{\mathcal{E}})$ is a predicted
DAG and $\hat{\mathcal{P}}$ is the predicted set of CPDs. The reference label
$(\mathcal{G}^*, \mathcal{P}^*)$ comes from PRISM-BN, with each example having
3 to 9 nodes, 2 to 14 directed edges, and fully specified CPD matrices whose
columns sum to 1.0. Predicted nodes and states must first be matched to the
reference labels before any structural or probabilistic metric can be computed.

The names a generator produces, such as ``Rain,'' often differ from reference
names, such as ``Rainfall,'' even when they denote the same variable. We use
Llama 3.3 70B as a fixed semantic judge that returns 1 if two names mean the
same thing or are clear paraphrases, and 0 otherwise. Matching is one-to-one:
each predicted name can be matched to at most one reference name, and the same
binary scheme is used for matching states within a node. The judge prompts are
listed in Appendix~\ref{app:judge}.

\textbf{Node identification.}
The judge compares each predicted node name to each reference node name and
returns a binary score. A match counts only if the score is 1. A predicted name
that matches a reference name already taken by another prediction is treated as
a duplicate, not a new match. Let $k$ be the number of correct matches,
$\hat{n}$ the number of predicted nodes, $d$ the number of duplicates, and $n$
the number of reference nodes. The effective predicted count is
$\hat{n}^* = \max(\hat{n}-d,k)$, which prevents duplicates from inflating
precision. Precision and recall are $P=k/\hat{n}^*$ and $R=k/n$, and
node $F_1=2PR/(P+R)$.

\textbf{State enumeration.}
States are scored only for nodes matched in Phase 1; states attached to
spurious predicted nodes are skipped. For each matched node $i$, the judge
returns binary scores between predicted and reference states, with ``None''
always matching ``None.'' Let $m_i$ be the number of state matches out of
$\hat{s}_i$ predicted and $s_i$ reference states. State $F_1$ is
macro-averaged over the $K$ matched nodes:
\begin{equation}
  F_1^{\text{state}} = \frac{1}{K}\sum_{i=1}^{K}
    \frac{2 (m_i/\hat{s}_i)(m_i/s_i)}{(m_i/\hat{s}_i) + (m_i/s_i)}.
\end{equation}
This is a conditional metric: it evaluates state recovery only for nodes that
were successfully aligned in Phase 1.

\textbf{Edge recovery.}
Edges are scored only when both endpoints were matched in Phase 1. Predicted
edges touching an unmatched node are dropped. Remaining predicted node names
are renamed to their matched reference names using the Phase 1 alignment, and
an edge is correct if the resulting $(\text{parent},\text{child})$ pair appears
in the reference edge set. Reversed edges receive no partial credit. Edge $F_1$
is the standard precision--recall harmonic mean over the renamed edge sets. The
spurious edge rate is reported separately as
$(\hat{e}_m-e_c)/\hat{e}_m \times 100\%$, where $\hat{e}_m$ is the number of
mapped predicted edges and $e_c$ the number correctly predicted. Edge $F_1$ is
therefore also conditional on successful node alignment.

\textbf{CPD estimation.}
CPDs are scored only on edges correctly recovered in Phase 3, which implies
that both endpoints were also matched in Phase 1. For each such edge, let
$p_{rc}$ be the reference CPD entry (row $r$, column $c$) and $\hat{p}_{rc}$
the prediction. We apply Laplace smoothing to whichever distribution appears
in the denominator:
$\tilde{p}_{rc}=(\hat{p}_{rc}+\varepsilon)/\sum_{r'}(\hat{p}_{r'c}+\varepsilon)$
with $\varepsilon=10^{-6}$, and analogously $\tilde{q}_{rc}$ for the reverse
direction. KL divergence is computed column-wise and averaged across the $C$
parent-state columns:
\begin{equation}
  \mathrm{KL}_c = \sum_r p_{rc}\log\frac{p_{rc}}{\tilde{p}_{rc}}, 
  \overline{\mathrm{KL}}(p\,\|\,\tilde{p}) =
  \frac{1}{C}\sum_{c=1}^{C}\mathrm{KL}_c.
\end{equation}
Because forward KL is asymmetric, we additionally report a symmetrized variant:
\begin{equation}
  \overline{\mathrm{KL}}^{\mathrm{sym}} =
  \tfrac{1}{2}\bigl(\overline{\mathrm{KL}}(p\,\|\,\tilde{p})+
  \overline{\mathrm{KL}}(\hat{p}\,\|\,\tilde{q})\bigr),
\end{equation}
with the reverse term defined analogously by swapping reference and predicted
CPD columns. The reported CPD-KL and CPD-KL-Sym are the means over all
correctly recovered edges. CPD-KL therefore measures agreement with the
PRISM-BN reference parameterization on aligned and structurally recovered
edges, not agreement with external observational or interventional data.

\section{Evaluation and Results}
\label{sec:results}
\textbf{Models and judge.}
We evaluate six LLMs as text-to-BN generators: Claude Haiku 4.5 (claude-haiku-4-5), DeepSeek-V3 (deepseek-ai/DeepSeek-V3), Gemma 3 12B (google/gemma-3-12b-it), GPT-4o-mini (gpt-4o-mini), Llama 4 Maverick (meta-llama/Llama-4-Maverick-17B), and Qwen3-30B-A3B (Qwen/Qwen3-30B-A3B-Instruct-2507).
We use Llama 3.3 70B
(meta-llama/Llama-3.3-70B-Instruct:groq) as a fixed semantic judge for
all generators. Each generator runs through the four-phase pipeline using
structured prompts (Appendix~\ref{app:prompts}), and metrics are reported as
means across paired examples, stratified by reference graph size.
\begin{figure*}[t]
\centering
\includegraphics[width=.95\textwidth]{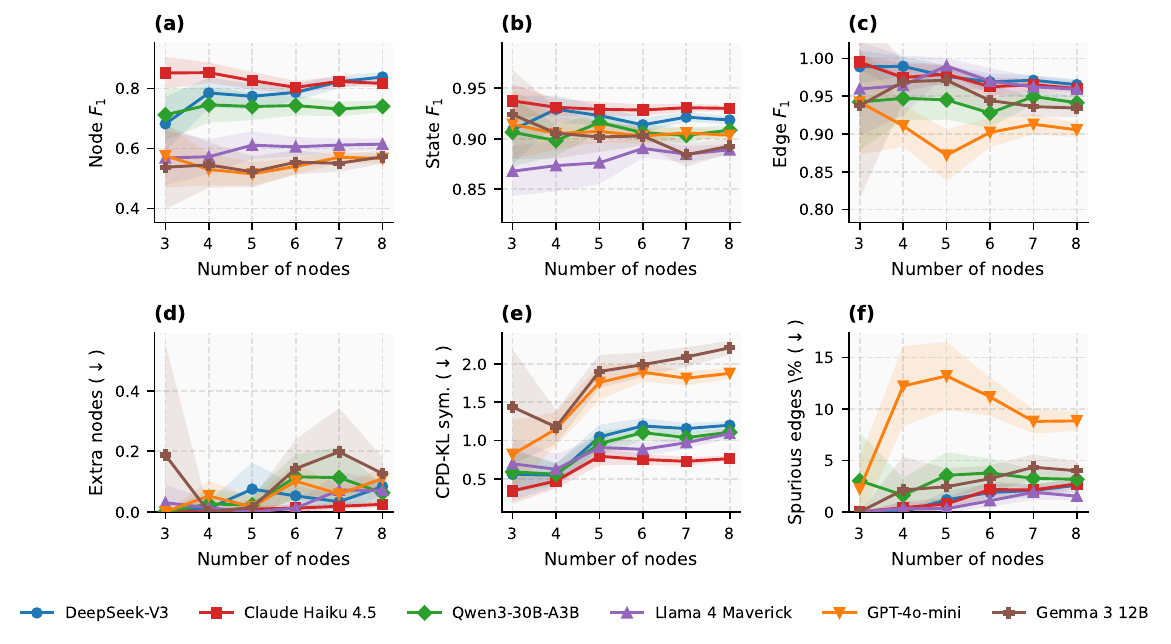}
\caption{Per-metric performance versus reference graph size across six
text-to-BN generators. Lines show per-bin means; shaded bands are confidence
intervals. Panels (a)--(c) report Node, State, and Edge $F_1$; (d) reports
extra predicted nodes; (e) reports symmetric CPD-KL; and (f) reports the
percentage of predicted edges with no reference correspondence. State and
edge metrics are conditional on Phase~1 node alignment. CPD metrics are
conditional on the node, state, and structural alignments required to compare
the complete multi-parent CPD of an eligible child.}
\label{fig:scaling_2x3}
\end{figure*}

Figure~\ref{fig:scaling_2x3} shows that DeepSeek-V3 and Claude Haiku 4.5
achieve Node $F_1$ above 0.81 and conditional Edge $F_1$ above 0.95. Node
recall is near ceiling for all models (0.972--0.987), whereas node precision
varies substantially from 0.395 to 0.711; the complete per-phase precision and
recall results are provided in Appendix~\ref{app:precision_recall}. Thus, node
over-generation is the dominant structural error. Conditional State $F_1$
ranges from 0.886 to 0.931, and conditional Edge $F_1$ ranges from 0.900 to
0.966. Because these metrics evaluate states and edges only after successful
node alignment, end-to-end structural recovery remains bounded by Node $F_1$.
Strict full-CPD reference agreement produces a different model ranking. Claude
Haiku 4.5 obtains the lowest CPD-KL (1.11), despite ranking second on Node
$F_1$, while DeepSeek-V3 is strongest on Node $F_1$ but ranks third on CPD-KL
(1.517). Llama 4 Maverick ranks second on CPD-KL (1.14), followed by
Qwen3-30B-A3B (1.72), GPT-4o-mini (2.25), and Gemma 3 12B (3.14). 

To contextualize the absolute CPD-KL scale, we compare the generators with
three calibration anchors using the same column-wise KL calculation.
\emph{Uniform} is a text-independent anchor that assigns
$1/|S_C|$ to every child state in each scored column. \emph{Marginal} is a
reference-side, parent-ignorant oracle anchor that assigns the reference child
marginal $P(C{=}c)$ to every column while ignoring the parent-state assignment.
\emph{Dirichlet} ($\alpha=1$) independently samples each column from
$\mathrm{Dirichlet}(\mathbf{1})$. A Gaussian-noise sweep and symmetric-KL
variants are reported in Appendix~\ref{app:cpd-baselines}.
All six generators have higher strict reference divergence than the Uniform and Marginal anchors in the reported evaluation. The domain-wise performance has been discussed in Appendix ~\ref{app:per-domain}.
To assess whether the results depend on the model family used to construct the reference BNs, we repeated the evaluation using a separately generated GPT-5.5 reference set under the same construction and evaluation protocol. Despite changes in absolute scores, the relative performance trends across extraction models remain broadly consistent; detailed results are provided in
Appendix~\ref{app:gpt_ref_results}.


\paragraph{Human validation and semantic matching.}
 The manual audit of 344 text--BN pairs verifies coverage of the intended nodes,
states, and edges, but does not validate the Llama-based semantic judge. We
therefore treat judge validation separately. In a small extraction pilot, two
annotators independently reconstructed ten BNs, obtaining Node $F_1$ scores of
0.92/0.95, conditional State $F_1$ scores of 0.95/0.98, conditional Edge $F_1$
scores of 0.95/0.95, and full-CPD KL values of 0.37/0.53. Pairwise agreement was
96\% for nodes, 93\% for states, and 98\% for edges, with an inter-annotator full-CPD KL of 0.14. 
Full validation details are provided in Appendix~\ref{app:human}.

\section{Conclusion}
We present PRISM-BN, a corpus of 5{,}054 controlled text--BN pairs derived from 50 Wikipedia-seeded backbones. Each example pairs a BN-grounded description
with variables, states, directed edges, root priors, and full multi-parent CPDs, constructed through marginal-first probability elicitation and analytical normalization.
Across six LLM extractors, we observe a consistent pattern: node over-generation is the primary structural error, conditional state and edge recovery remains high, and strict full-CPD reference agreement remains more
challenging. The same relative trends persist with independently generated GPT-5.5 references, indicating robustness to the choice of reference generator.
Human evaluation further supports these findings, demonstrating reproducible interpretations. PRISM-BN therefore enables rigorous, separate evaluation of structural recovery and probabilistic parameter estimation. We release the corpus, construction pipeline, and evaluation code.

\section{Limitations}
\label{sec: limitations}
PRISM-BN is a controlled benchmark, not an externally validated causal knowledge
base. Its reference structures, marginals, and joint distributions are produced
by a single construction pipeline seeded by Wikipedia articles, so CPD-KL
scores in \S\ref{sec:results} should be interpreted as agreement with the
PRISM-BN reference parameterization rather than accuracy against observational
or interventional data. The benchmark input is a generated BN-grounded
description, not the original Wikipedia article, so performance measures
controlled text-to-BN extraction rather than open-domain causal discovery.
Semantic matching relies on a single judge model, and human-rater agreement is
not reported in this version. Finally, examples derived from the same source backbone are not independent, so
supervised use of PRISM-BN should follow the backbone-disjoint split guidance in Section 
\S\ref{sec:dataset_characteristics}.
\section{Ethical Considerations}

\textbf{Data provenance and licensing.} PRISM-BN is constructed entirely from publicly available Wikipedia content under CC BY-4.0 and is released under the same license to preserve compatibility with its source. Source articles are encyclopedic entries on abstract causal topics (e.g., ``Economic crises,'' ``Deforestation''), and the generated BN-grounded descriptions verbalize variable names, states, and causal edges rather than information about identifiable individuals. We did not perform an explicit offensive-content scan of the generated descriptions; given the abstract nature of the source categories we judge this risk to be low, but we note it as a residual limitation.

\textbf{Intended use and misuse.} PRISM-BN is intended as a controlled benchmark for training and evaluating text-to-parameterized-BN extraction systems. It is \emph{not} a factual causal knowledge base. The reference structures, marginals, and conditional probability distributions are produced by a single Wikipedia-seeded construction pipeline and represent internally consistent benchmark targets rather than externally validated estimates of real-world causal effects. Accordingly, CPD-KL measures agreement with the PRISM-BN reference parameterization, not accuracy against observational or interventional data. We caution against deploying systems trained or evaluated on PRISM-BN in high-stakes domains such as policy, economics, public health, or environmental decision-making without independent validation against domain-specific evidence. Users should treat outputs of PRISM-BN-trained systems as candidate hypotheses to be verified, not as authoritative causal claims.

\textbf{LLM use in construction and evaluation.} LLMs are central to both the construction pipeline and the benchmark. Reference BNs are produced by an automated pipeline using Claude Sonnet 4.6, with coverage verification by Claude Opus 4.6; the semantic judge is Llama 3.3 70B; and the six evaluated generators are themselves LLMs. This introduces three forms of model-mediated bias that users should keep in mind: (i) reference structures and probabilities may reflect systematic biases of the generating model, (ii) semantic matching depends on a single judge model whose agreement with human annotators is not quantified in this version, and (iii) evaluated models from the same family as the generator may share inductive biases with the reference labels. We discuss these limitations in the \S\ref{sec: limitations} and encourage future work to extend validation with multi-model cross-checks and human inter-rater studies.


\textbf{Use of AI assistants in preparing this work.} We used Claude Opus 4.7 to assist with editing portions of the manuscript and with portions of the code used in figure generation. All design decisions, experimental choices, analyses, and final claims are the authors' own, and all AI-assisted text and code were reviewed and verified by the authors prior to inclusion.

\bibliography{reference}

\appendix
\section{Supplementary Dataset Figures}
\label{app:dataset_figures}

This appendix collects the distributional figures supporting the aggregate
statistics in \S\ref{sec:dataset_characteristics}. Subgraph-level figures
aggregate over 5{,}054 graphs and 38{,}820 node records. Backbone-level figures
aggregate over 50 graphs and 584 node records.
\subsection*{Source Backbones (50 BNs)}

We turn now to the 50 source backbones from which all 5{,}054 subgraphs are
extracted. Backbone-level statistics differ from the subgraph-level statistics
above in scale (larger graphs, longer paths) but follow similar per-domain
patterns.

\begin{figure}[t]
    \centering
    \includegraphics[width=\linewidth]{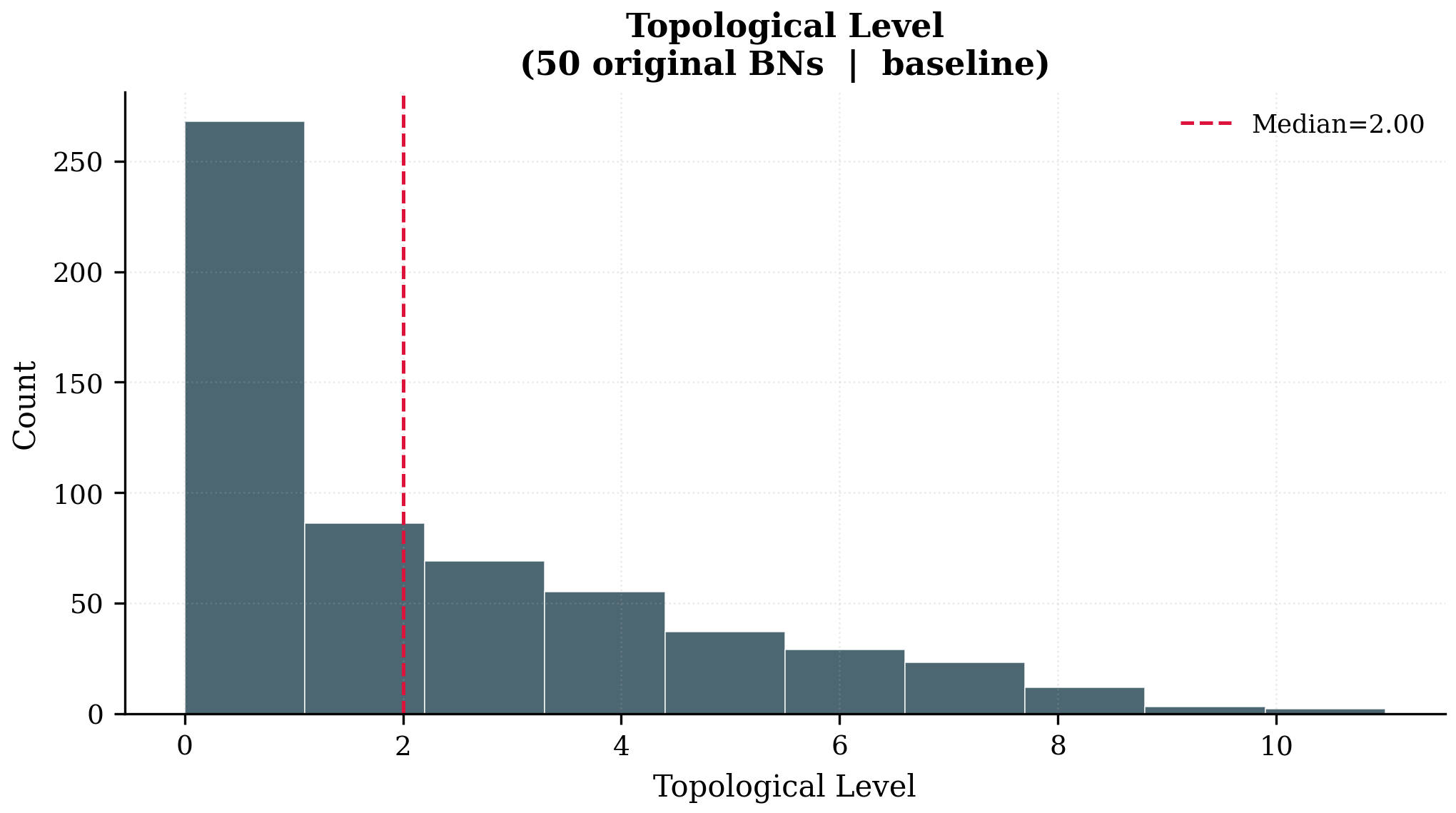}
    \caption{Topological-level distribution across the 584 backbone node
    records (median 2, max 11). The longer tail relative to the subgraph
    corpus reflects backbone depths up to 11.}
    \label{fig:bb_topo_level}
\end{figure}

Figure~\ref{fig:bb_topo_level} reports the topological-level distribution
across the 584 backbone node records. The longer tail relative to the subgraph
corpus reflects backbone depths up to 11, substantially deeper than the
extracted subgraphs, which is exactly what the subgraph augmentation procedure
is designed to flatten.

\begin{figure}[t]
    \centering
    \includegraphics[width=0.75\linewidth]{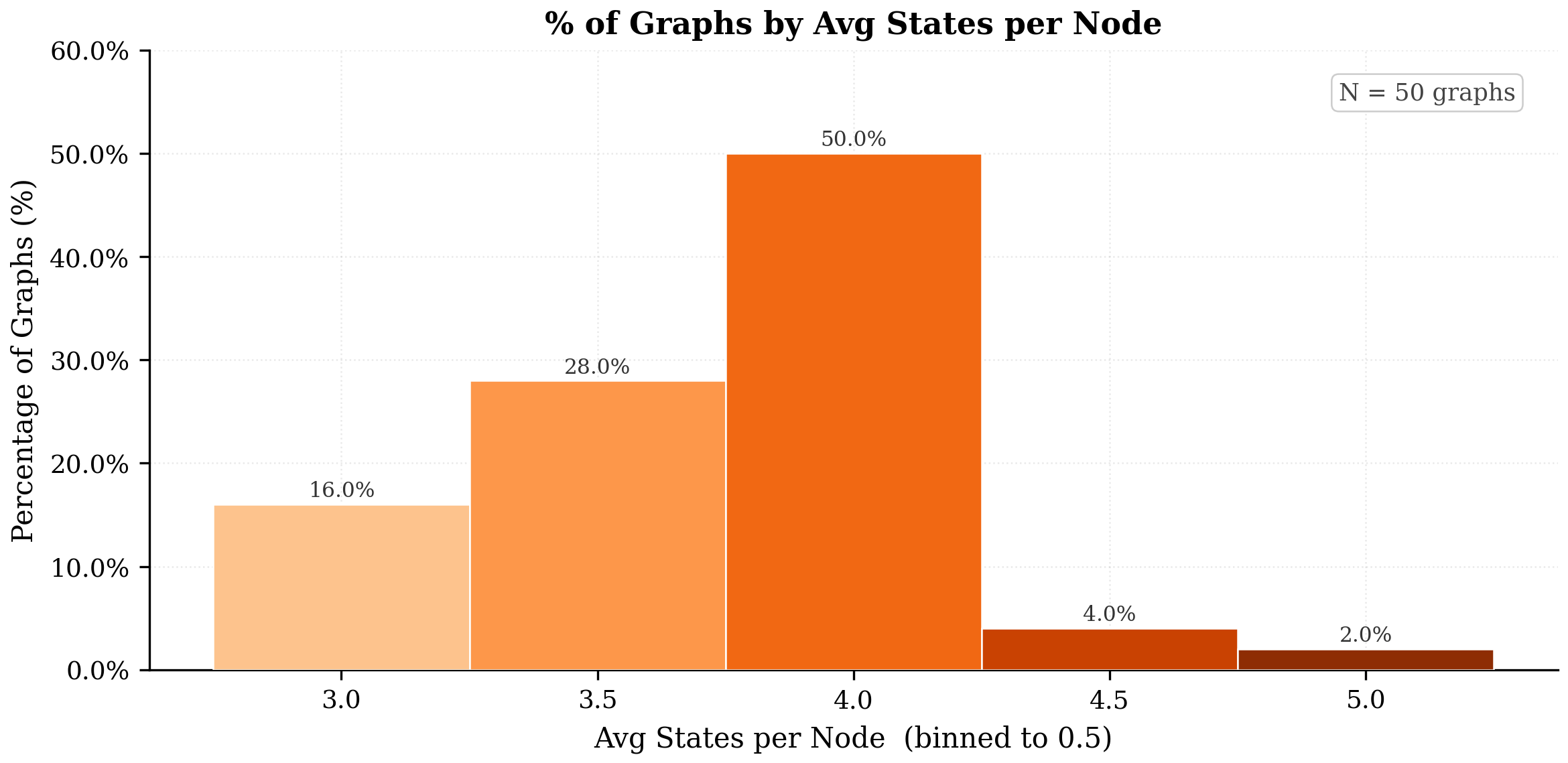}
    \caption{Average states per node across the 50 backbones. The distribution
    is concentrated near 4, similar to the subgraph corpus.}
    \label{fig:bb_avg_states_pct}
\end{figure}

\begin{figure*}[t]
    \centering
    \includegraphics[width=0.75\linewidth]{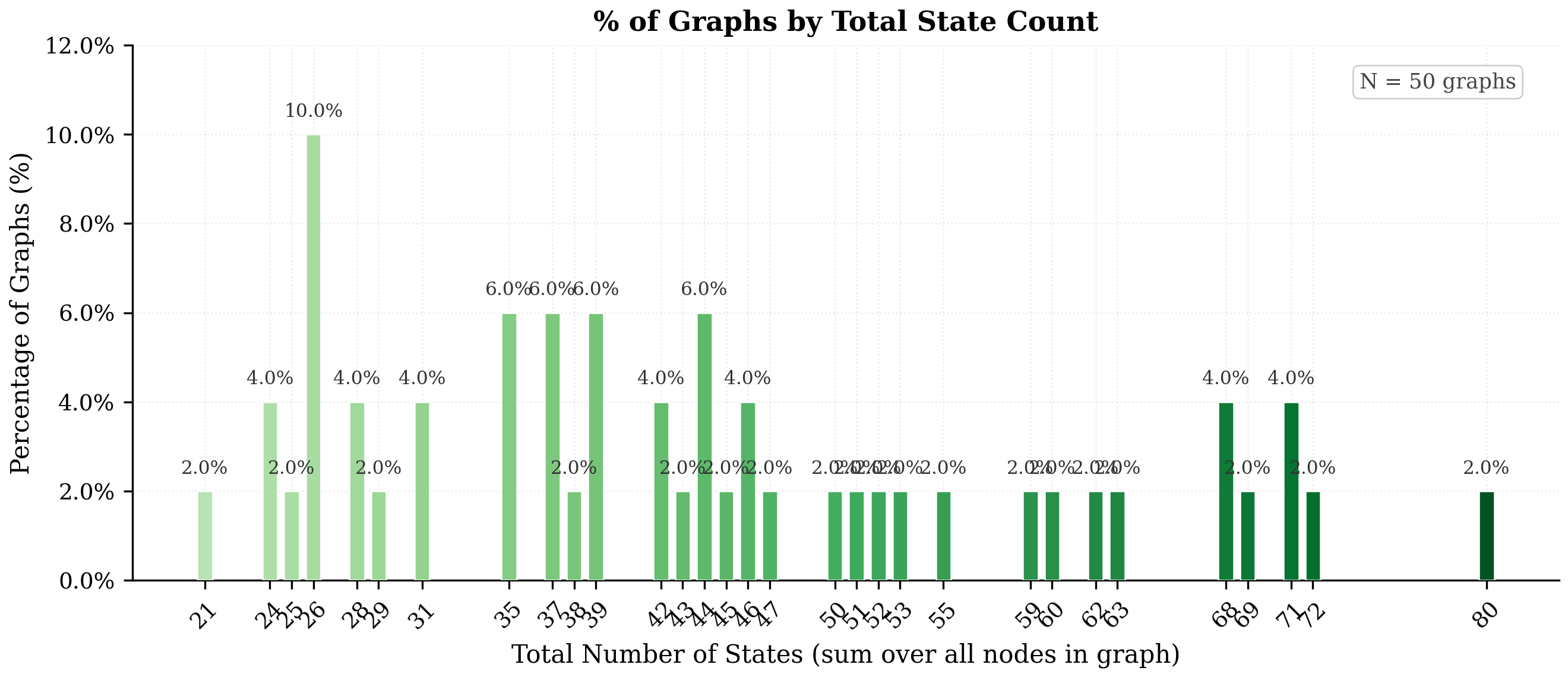}
    \caption{Total states per backbone (summed across all nodes). Total state
    counts span a substantially wider range than in the subgraph corpus,
    reflecting backbone sizes up to 20 nodes.}
    \label{fig:bb_total_states_pct}
\end{figure*}

State cardinality at the backbone level is summarised by
Figures~\ref{fig:bb_avg_states_pct} and~\ref{fig:bb_total_states_pct}. The
average-states distribution is similar to the subgraph corpus (concentrated
near 4), but the total-states distribution per backbone spans a far wider
range, reflecting backbone sizes up to 20 nodes.

\begin{figure*}[t]
    \centering
    \includegraphics[width=0.75\linewidth]{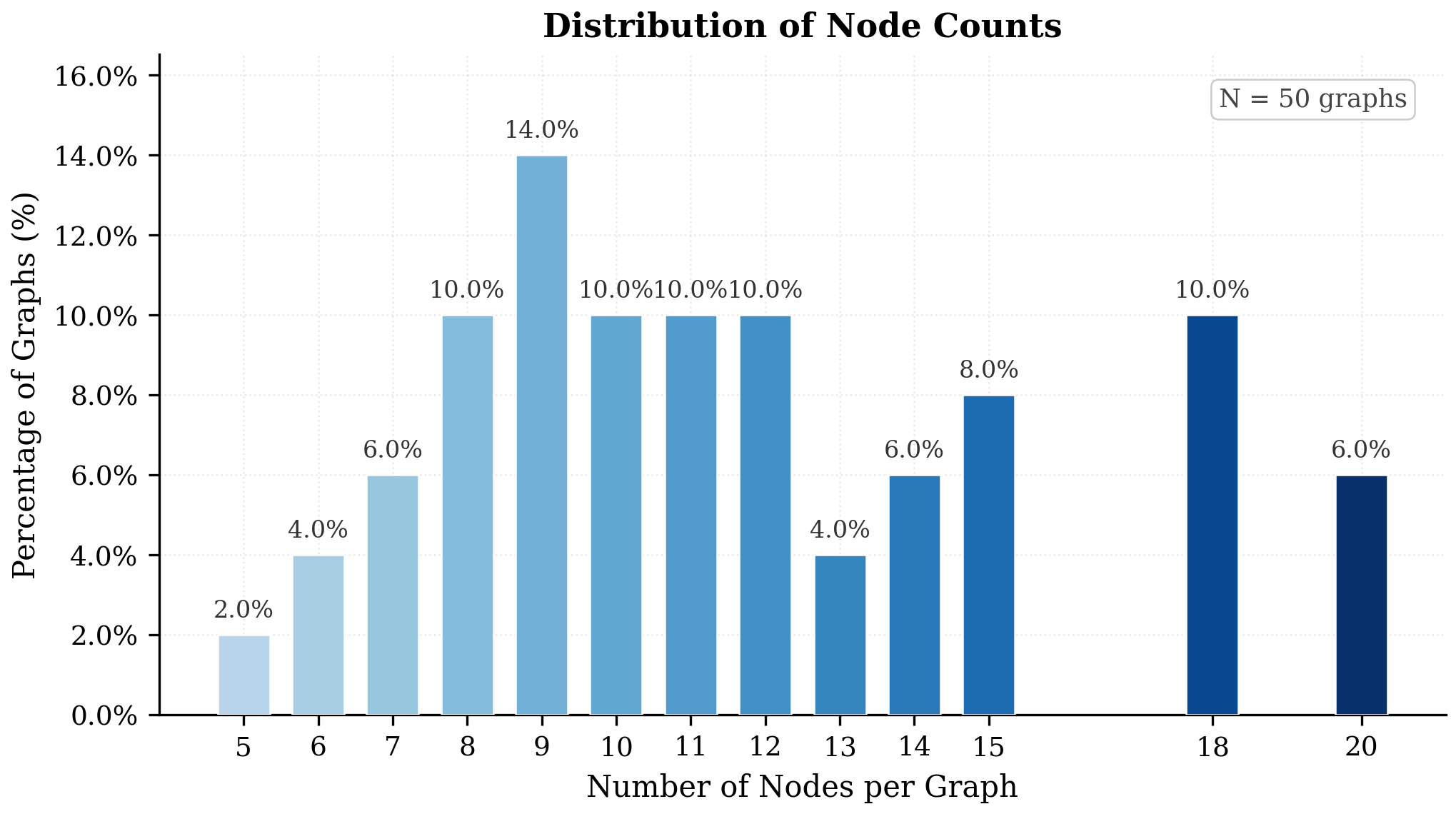}
    \caption{Backbone node-count distribution. Sizes span 5--20 nodes,
    substantially wider than the subgraph corpus.}
    \label{fig:bb_size}
\end{figure*}

\begin{figure*}[t]
    \centering
    \includegraphics[width=0.75\linewidth]{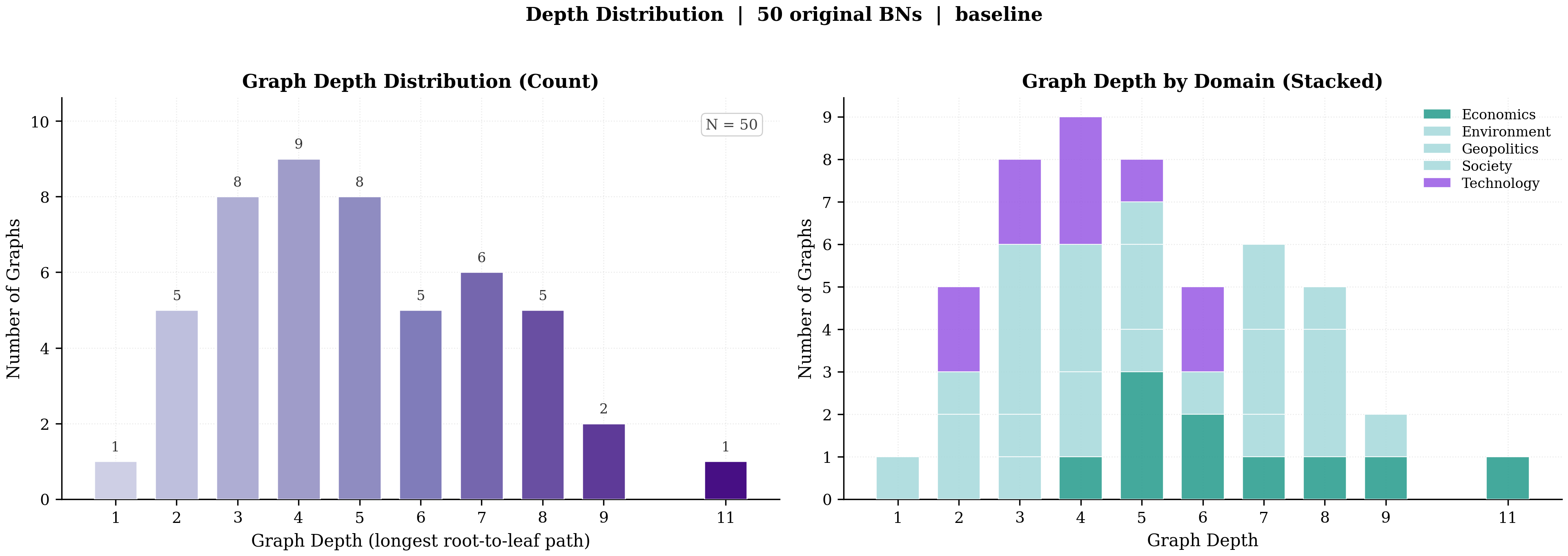}
    \caption{Backbone depth distribution. Depth peaks at 4 with a tail reaching
    11, motivating subgraph augmentation.}
    \label{fig:bb_depth}
\end{figure*}

Backbone size and depth appear in Figures~\ref{fig:bb_size} and
\ref{fig:bb_depth}. Sizes span 5--20 nodes, and depth peaks at 4 with a tail
reaching 11. These wider distributions, relative to the subgraph corpus,
motivate subgraph augmentation as the mechanism for generating a large corpus
of small, tractable examples from a comparatively small set of source graphs.

\begin{figure*}[t]
    \centering
    \includegraphics[width=\textwidth]{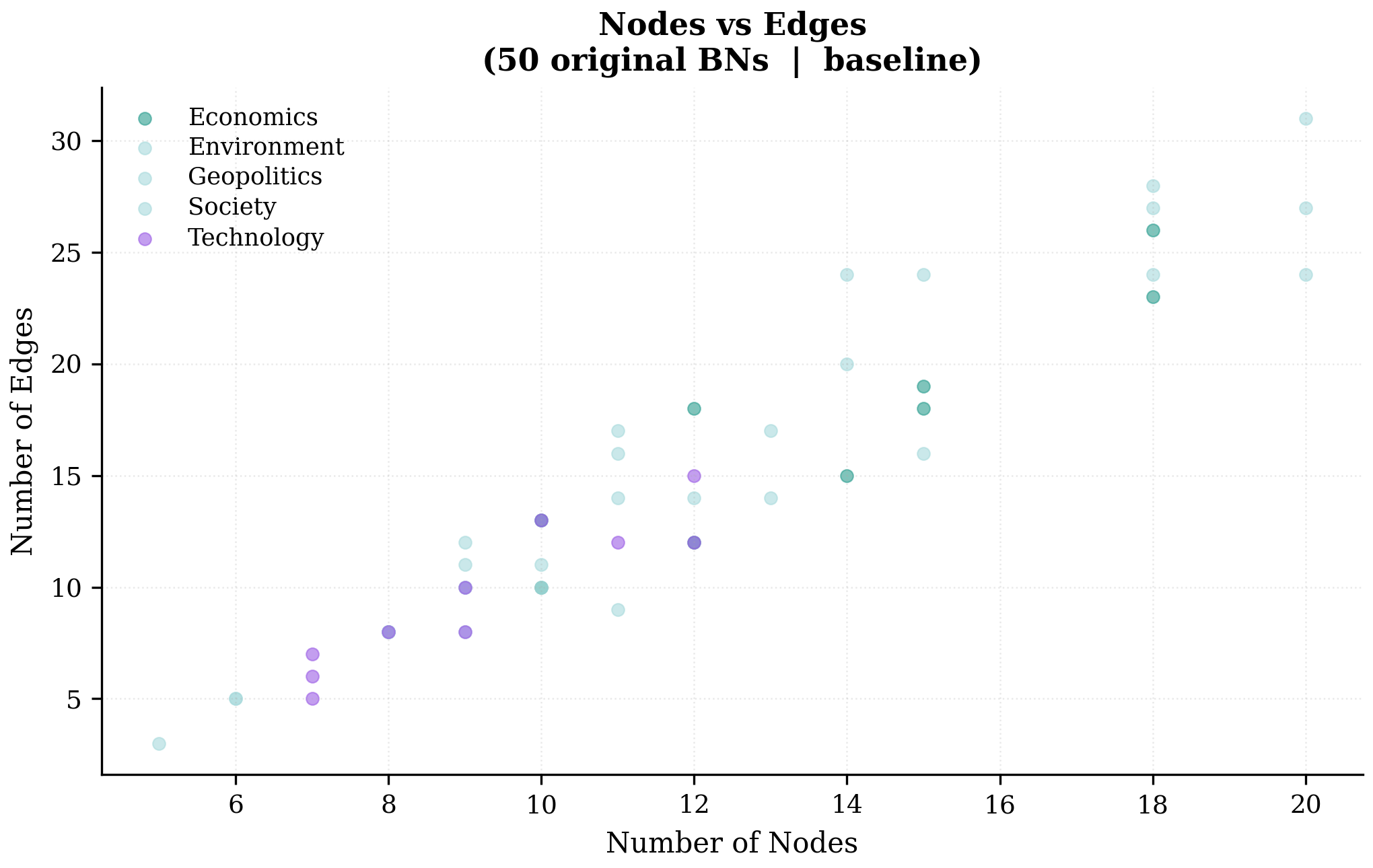}
    \caption{Joint (nodes, edges) distribution for the 50 backbones, coloured
    by domain. Geopolitics and Environment occupy the upper-right; Technology
    is concentrated in the lower-left.}
    \label{fig:bb_nodes_vs_edges}
\end{figure*}

The joint backbone (nodes, edges) distribution by domain appears in
Figure~\ref{fig:bb_nodes_vs_edges}. Geopolitics and Environment dominate the
upper-right (large, edge-dense backbones), while Technology is concentrated in
the lower-left.

\begin{figure*}[t]
    \centering
    \includegraphics[width=0.85\linewidth]{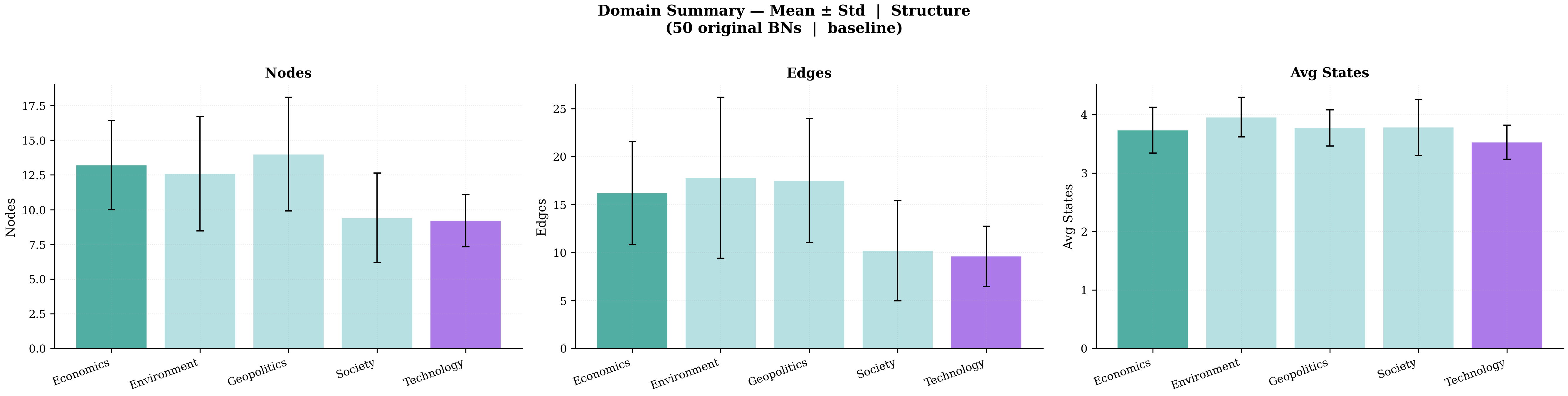}
    \caption{Backbone structural means by domain with $\pm 1\sigma$ bars
    (nodes, edges, depth).}
    \label{fig:bb_domain_bars_structure}
\end{figure*}

\begin{figure*}[t]
    \centering
    \includegraphics[width=0.85\linewidth]{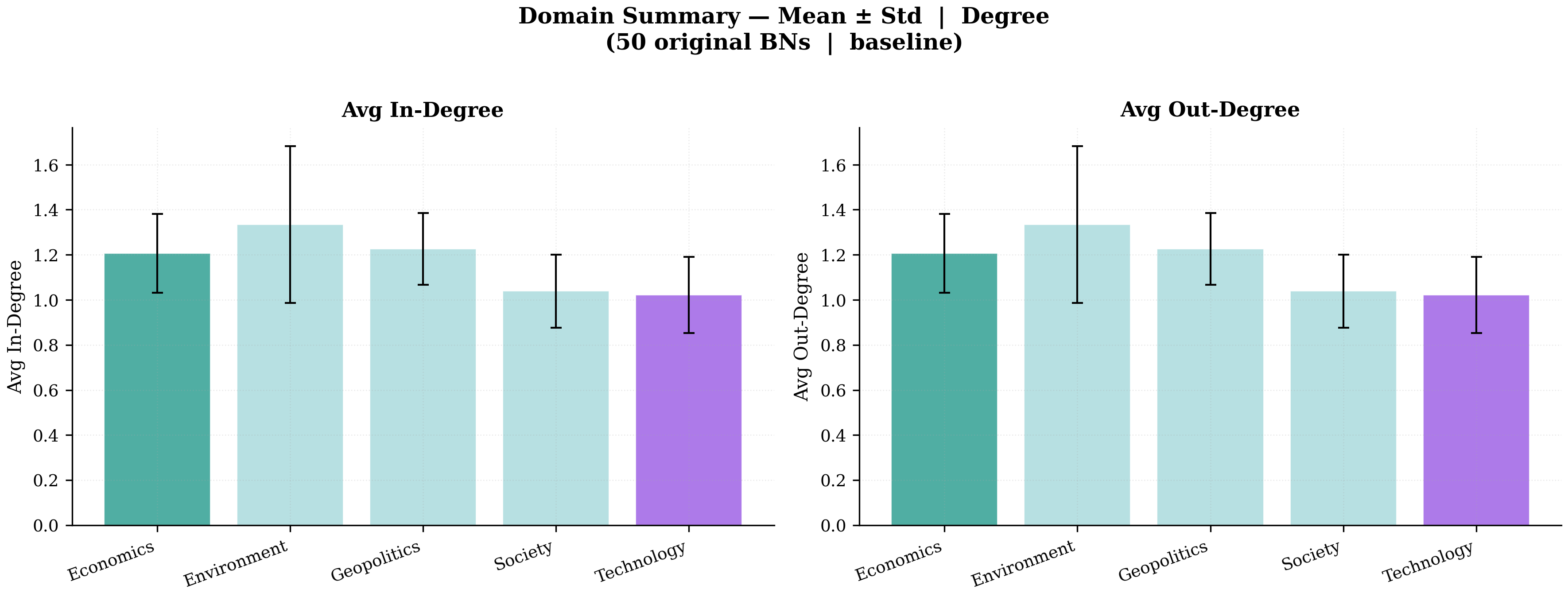}
    \caption{Backbone degree means by domain with $\pm 1\sigma$ bars (in-degree,
    out-degree). Backbone in-/out-degree is uniformly higher than at the
    subgraph level because the subgraph procedure re-roots orphaned parents
    and reduces per-node connectivity.}
    \label{fig:bb_domain_bars_degree}
\end{figure*}

Domain-level structural and degree statistics for the backbones are reported in
Figures~\ref{fig:bb_domain_bars_structure} and~\ref{fig:bb_domain_bars_degree}.
Backbone in-/out-degree is uniformly higher than at the subgraph level because
the subgraph procedure re-roots orphaned parents and reduces per-node
connectivity.

\begin{figure*}[t]
    \centering
    \includegraphics[width=\textwidth]{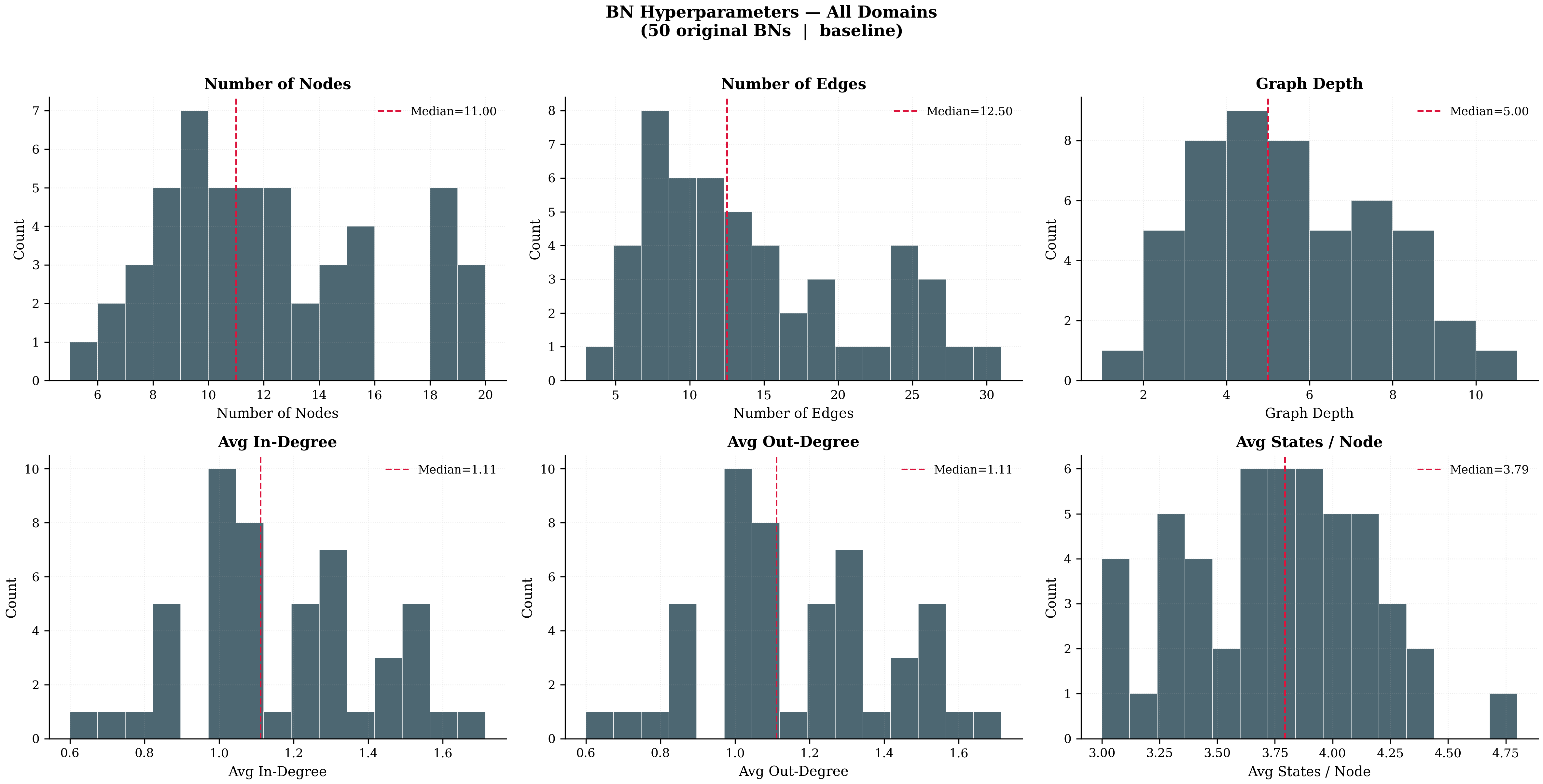}
    \caption{BN-level hyperparameter histograms across the 50 backbones. The
    wider distributions relative to the subgraph corpus motivate subgraph
    augmentation.}
    \label{fig:bb_bn_overview}
\end{figure*}

\begin{figure*}[t]
    \centering
    \includegraphics[width=\textwidth]{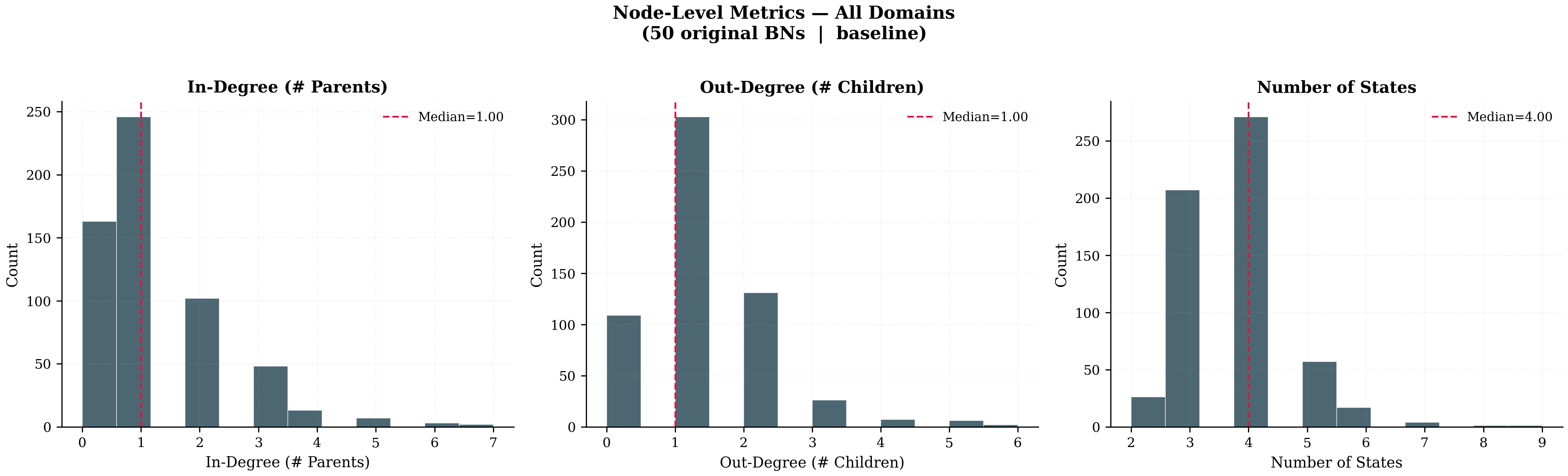}
    \caption{Node-level histograms across the 584 backbone records: in-degree,
    out-degree, and state cardinality. The high-fan-in tail motivates the
    multi-parent CPD treatment in \S\ref{sec:cpd}.}
    \label{fig:bb_node_overview}
\end{figure*}

Figures~\ref{fig:bb_bn_overview} and~\ref{fig:bb_node_overview} present the
overall BN-level and node-level histograms for the 50 backbones. The
substantially wider distributions, particularly in node count and depth, relative
to the subgraph corpus, motivate subgraph augmentation as the dataset's
combinatorial scaling mechanism. The high-fan-in tail in the per-node
in-degree histogram in particular motivates the multi-parent CPD treatment of
\S\ref{sec:cpd}.

\begin{figure*}[t]
    \centering
    \includegraphics[width=\textwidth]{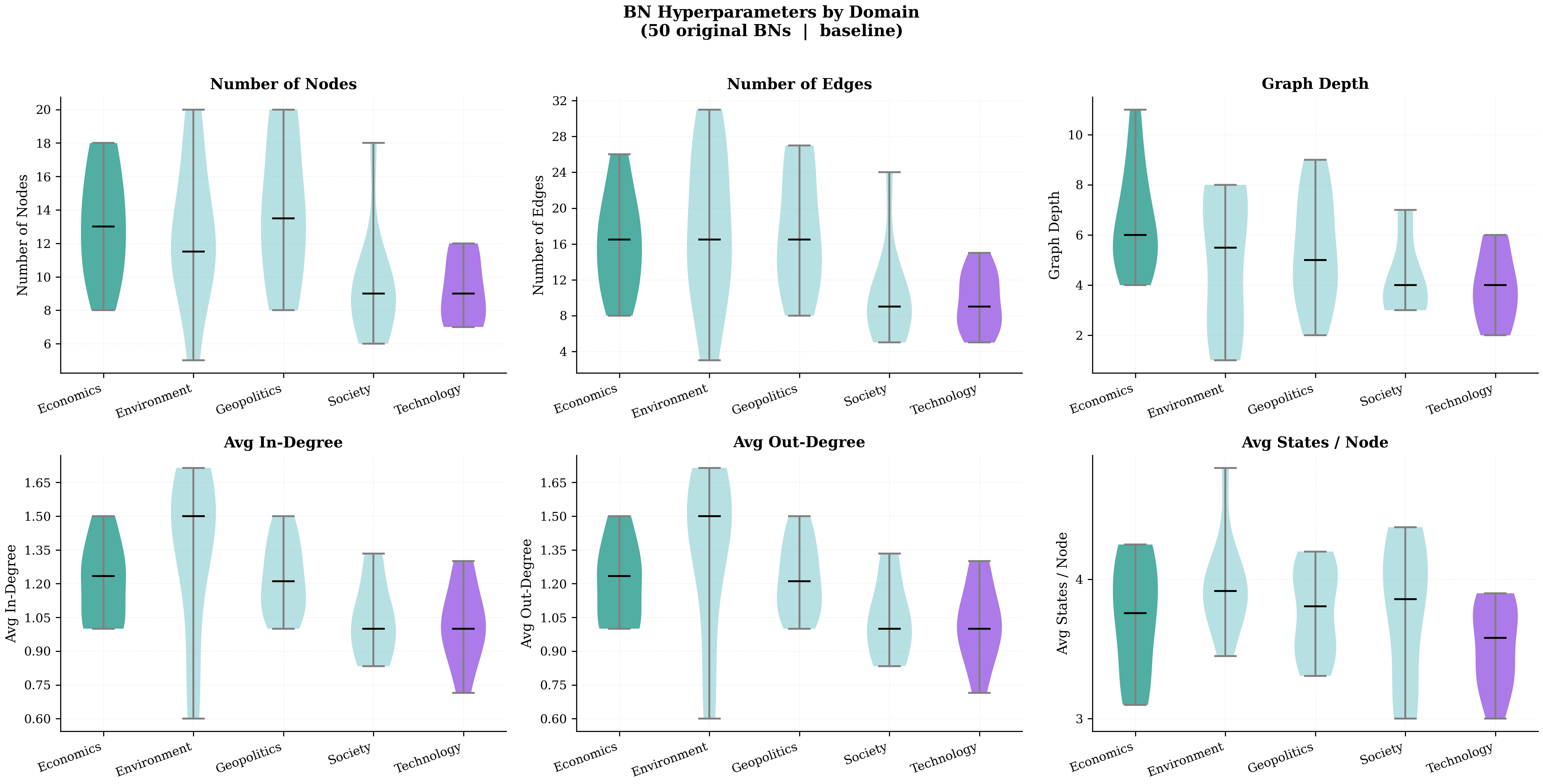}
    \caption{Backbone BN-level distributions, faceted by domain. Geopolitics
    is the structural outlier on the upper end; Society and Technology sit at
    the lower end.}
    \label{fig:bb_bn_by_domain}
\end{figure*}

\begin{figure*}[t]
    \centering
    \includegraphics[width=\textwidth]{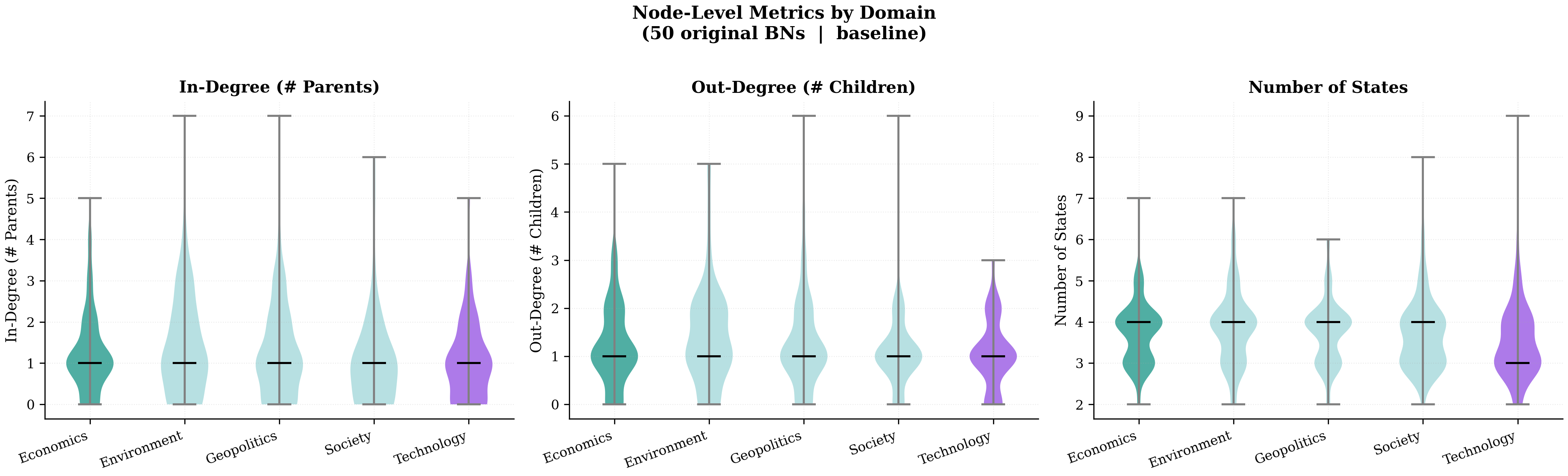}
    \caption{Backbone node-level distributions, faceted by domain. Degree
    distributions are nearly identical across domains; state cardinality is
    modal at 4 throughout.}
    \label{fig:bb_node_by_domain}
\end{figure*}

Finally, Figures~\ref{fig:bb_bn_by_domain} and~\ref{fig:bb_node_by_domain}
show the backbone BN-level and node-level distributions faceted by domain.
Geopolitics is the structural outlier on the upper end (largest, deepest
backbones); Society and Technology sit at the lower end. Degree distributions
remain nearly identical across domains, and state cardinality is modal at 4
throughout.

\subsection*{Released Corpus (5{,}054 Examples)}

\begin{figure*}[t]
    \centering
    \includegraphics[width=\textwidth]{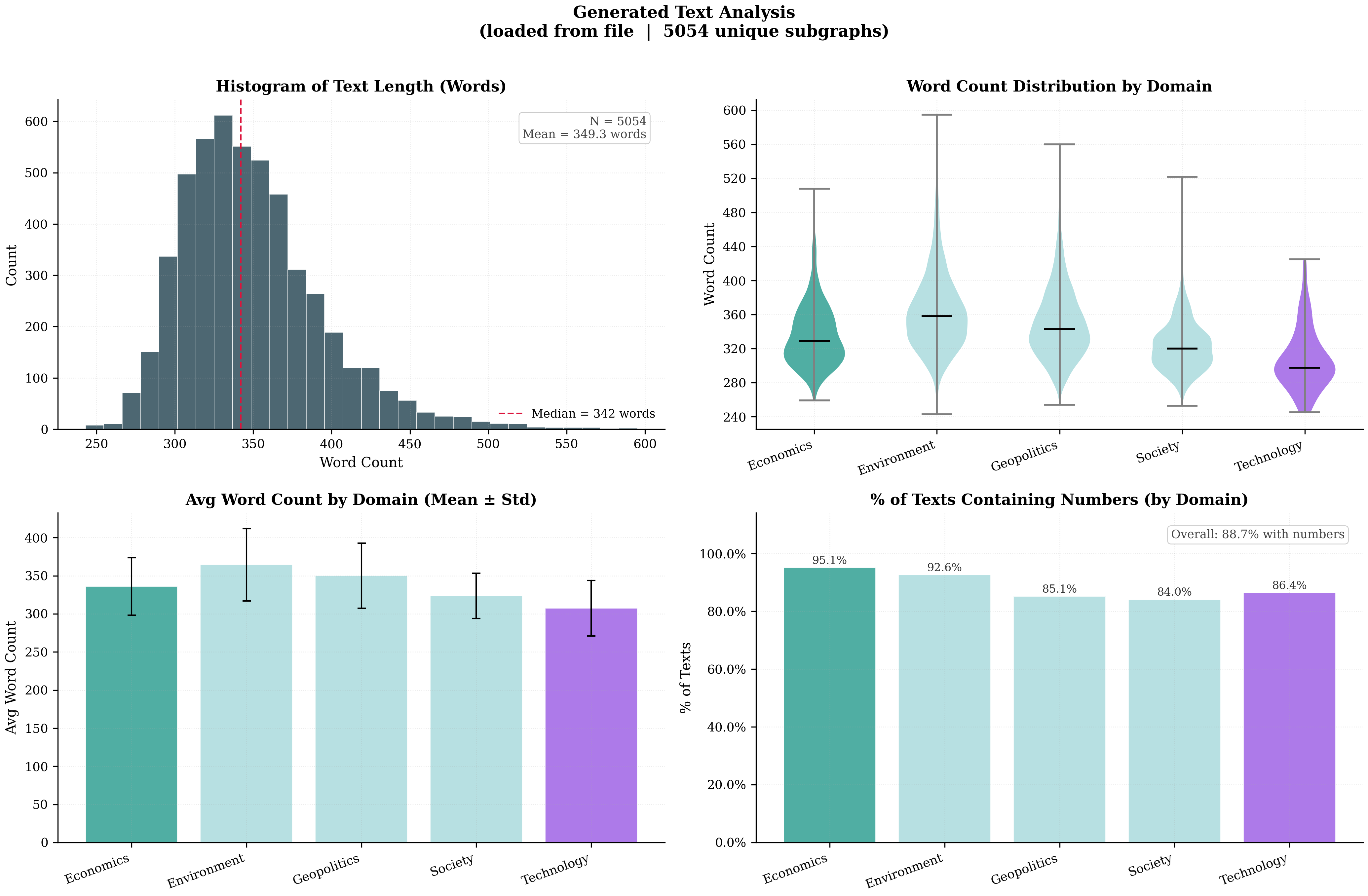}
    \caption{Generated descriptions across the 5{,}054 examples. Top-left:
    overall text-length histogram (mean 349.3 words, median 342, range
    [243, 595]). Top-right: per-domain word-count distributions. Bottom-left:
    per-domain means with $\pm 1$ standard deviation. Bottom-right: fraction of
    texts containing numerical values, ranging from 84.0\% (Society) to 95.1\%
    (Economics) with corpus mean 88.7\%.}
    \label{fig:text_analysis}
\end{figure*}

We begin with the generated descriptions that pair with each subgraph.
Figure~\ref{fig:text_analysis} characterizes the corpus along four axes: text
length, per-domain word-count distributions, per-domain means with variability,
and the fraction of texts containing explicit numerical values. The descriptions
are concentrated tightly around the corpus mean of 349 words, and 88.7\% of
texts contain at least one numerical expression, reflecting the
coverage-checked NLG procedure of \S\ref{sec:nlg}.

\begin{figure*}[t]
    \centering
    \includegraphics[width=\linewidth]{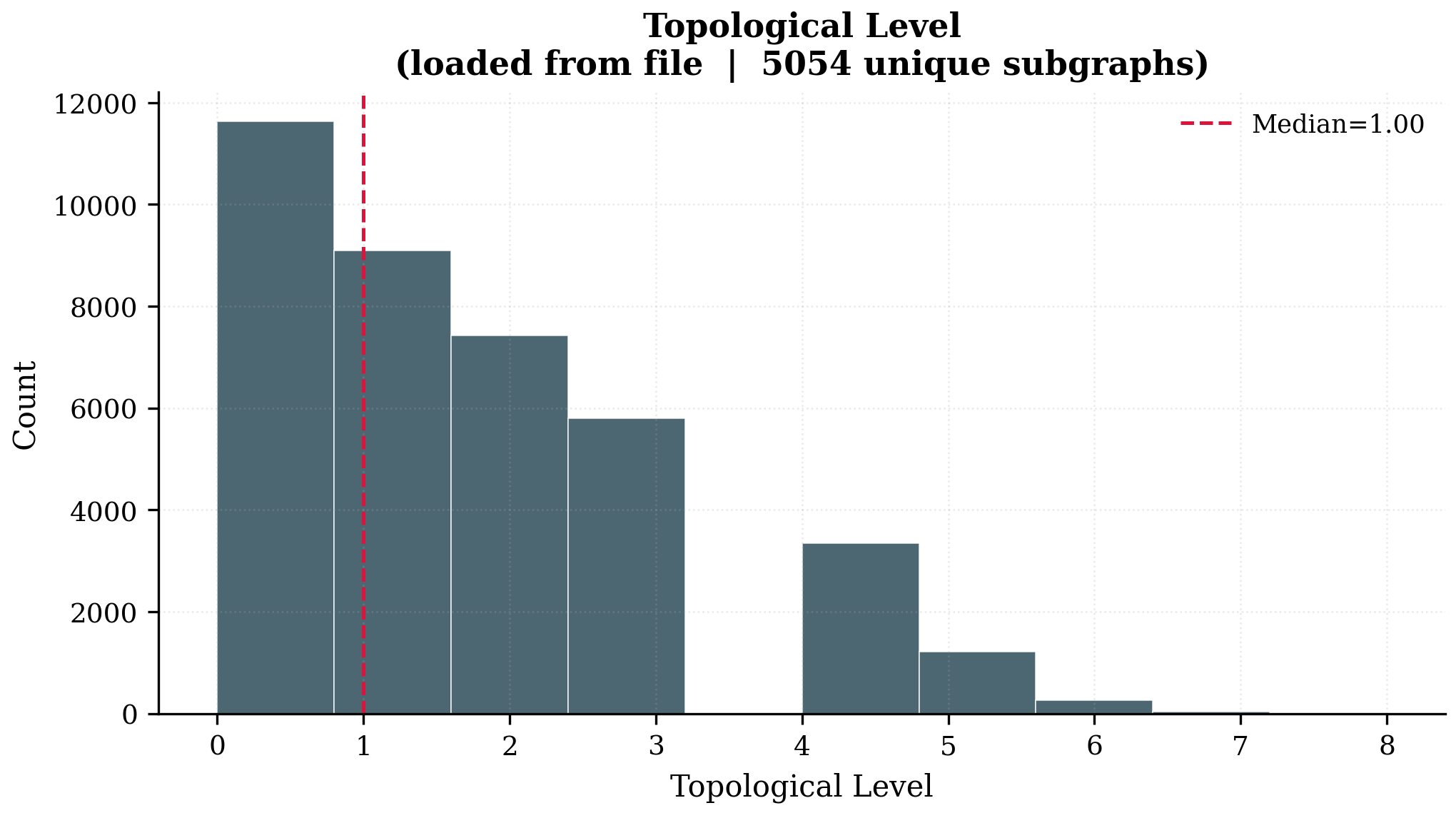}
    \caption{Topological-level distribution across the 38{,}820 node records
    (median 1, max 8). The bulk of nodes occupy the upper causal layers, with a
    long tail extending into deeper layers driven mainly by larger Geopolitics
    subgraphs.}
    \label{fig:topo_level}
\end{figure*}

Structural depth  is summarised in Figure~\ref{fig:topo_level}. The bulk of node
records sit in the upper layers (median topological level 1), with a
long tail extending to depth 8, driven mostly by larger Geopolitics subgraphs.

\begin{figure*}[t]
    \centering
    \includegraphics[width=0.75\linewidth]{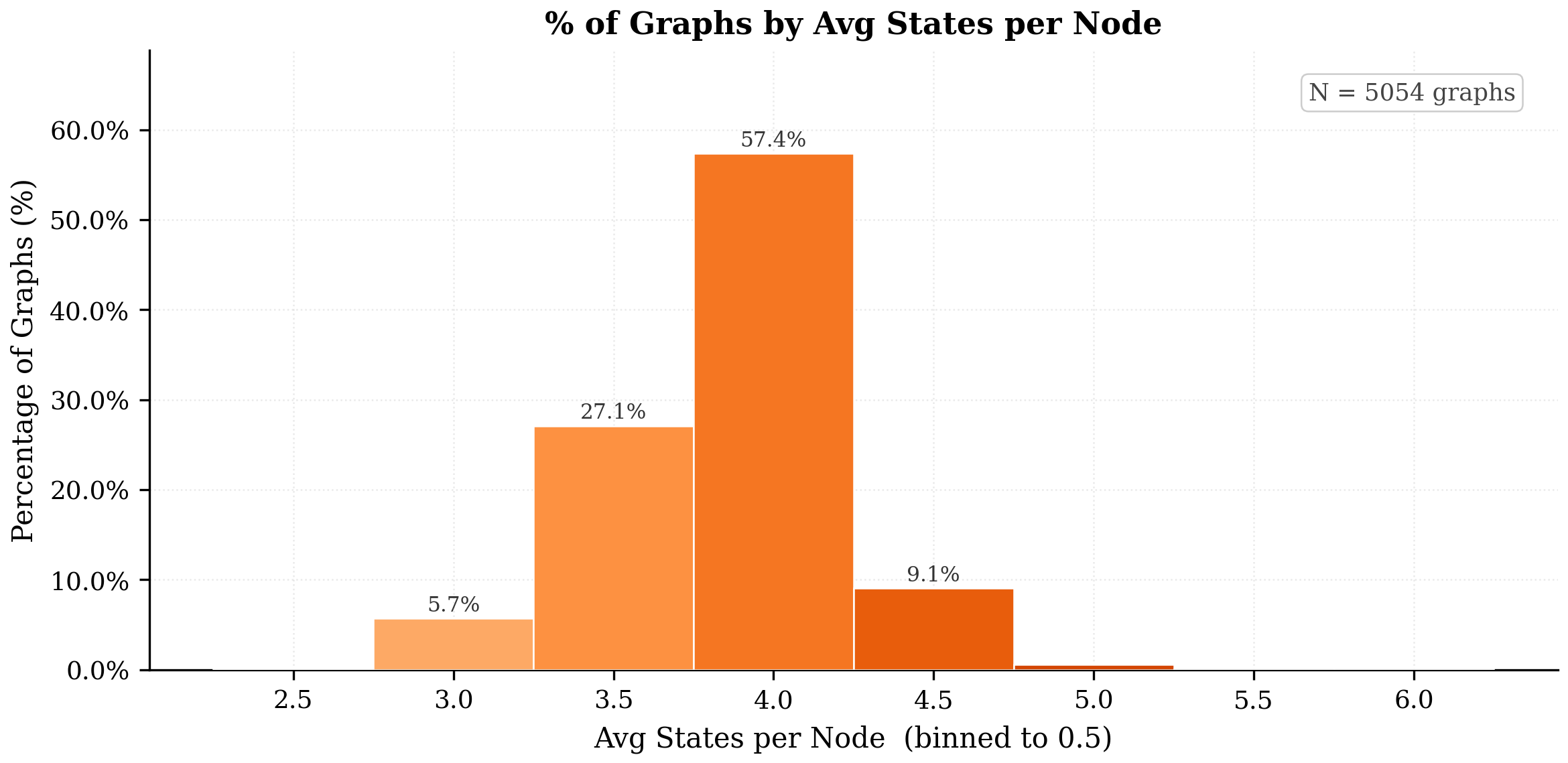}
    \caption{Average states per node across the 5{,}054 subgraphs. The
    distribution is sharply unimodal around 4.}
    \label{fig:avg_states_pct}
\end{figure*}

\begin{figure*}[t]
    \centering
    \includegraphics[width=0.75\linewidth]{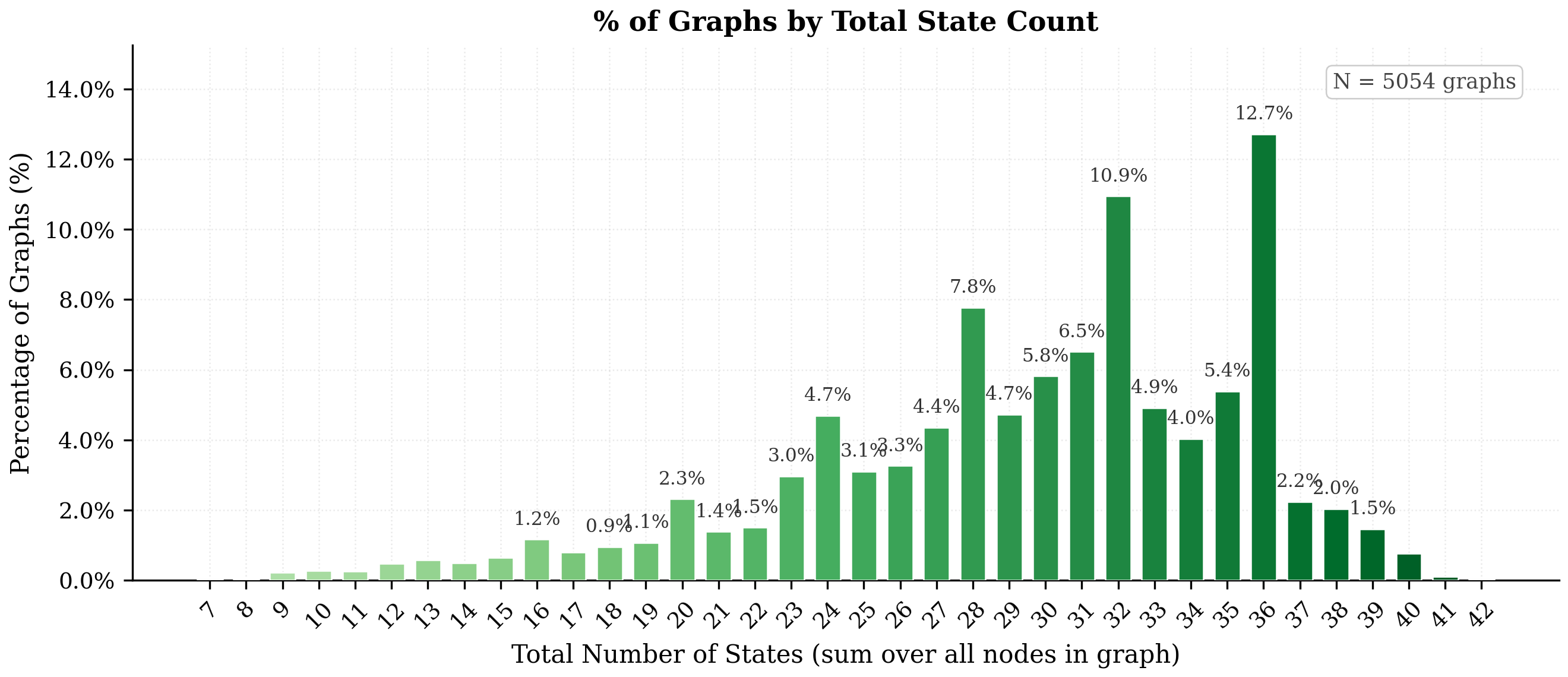}
    \caption{Total states per subgraph (summed across all nodes). The
    distribution peaks sharply at 36, reflecting the dominant 9-node, 4-state
    subgraph class.}
    \label{fig:total_states_pct}
\end{figure*}

State cardinality follows next. Figure~\ref{fig:avg_states_pct} shows the
distribution of average states per node, which is sharply concentrated around
4; Figure~\ref{fig:total_states_pct} shows total state counts per subgraph,
which peak at 36, consistent with the dominant 9-node, 4-state subgraph class.

\begin{figure*}[t]
    \centering
    \includegraphics[width=0.75\linewidth]{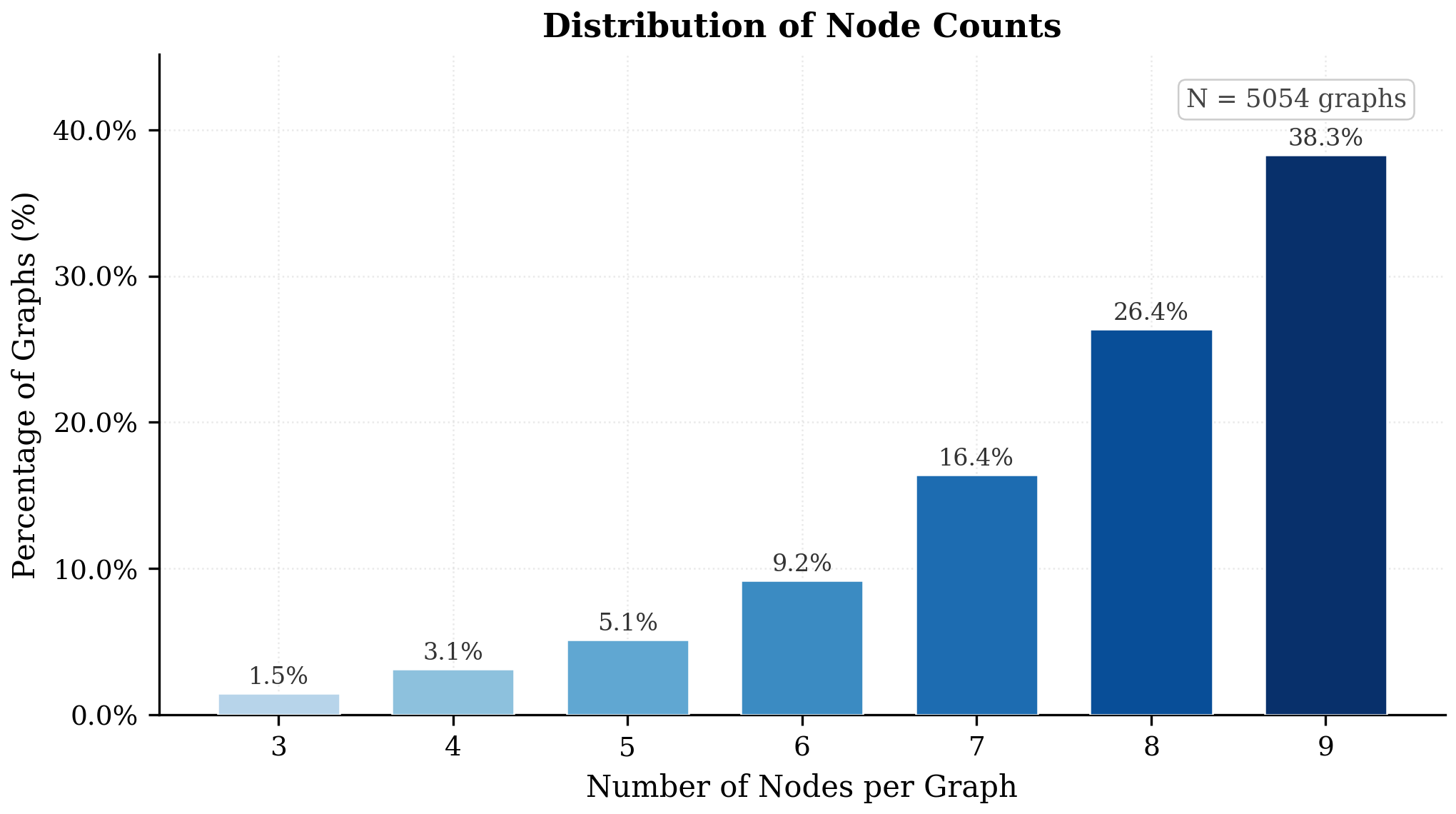}
    \caption{Subgraph node-count distribution. Subgraphs of 7--9 nodes account
    for 81.0\% of the corpus.}
    \label{fig:subgraph_size}
\end{figure*}

\begin{figure*}[t]
    \centering
    \includegraphics[width=0.75\linewidth]{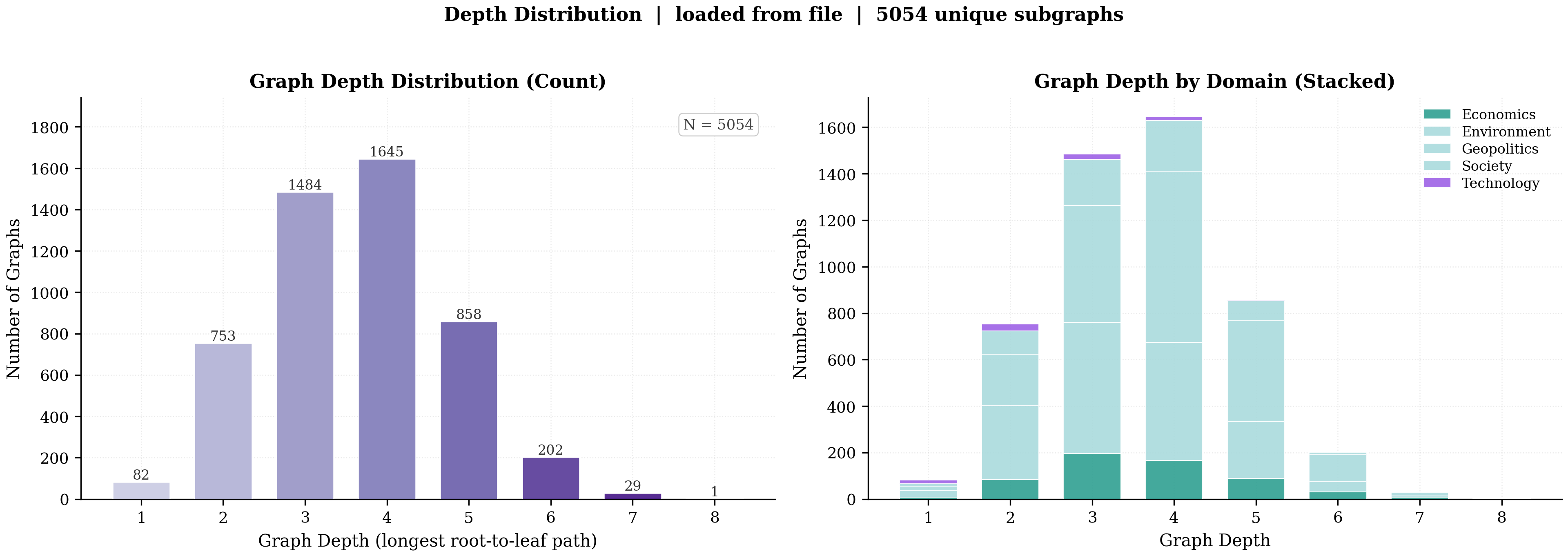}
    \caption{Subgraph depth distribution. Depth peaks at 4, with mass between
    2 and 6 covering the bulk of the corpus.}
    \label{fig:graph_depth}
\end{figure*}

Subgraph size and depth distributions appear in Figures~\ref{fig:subgraph_size}
and~\ref{fig:graph_depth}. Subgraphs of 7--9 nodes account for 81.0\% of the
corpus, and depth peaks at 4. The combination motivates the choice of
moderately-sized, moderately-deep subgraphs as the target regime for
text-to-BN extraction.

\begin{figure*}[t]
    \centering
    \includegraphics[width=\textwidth]{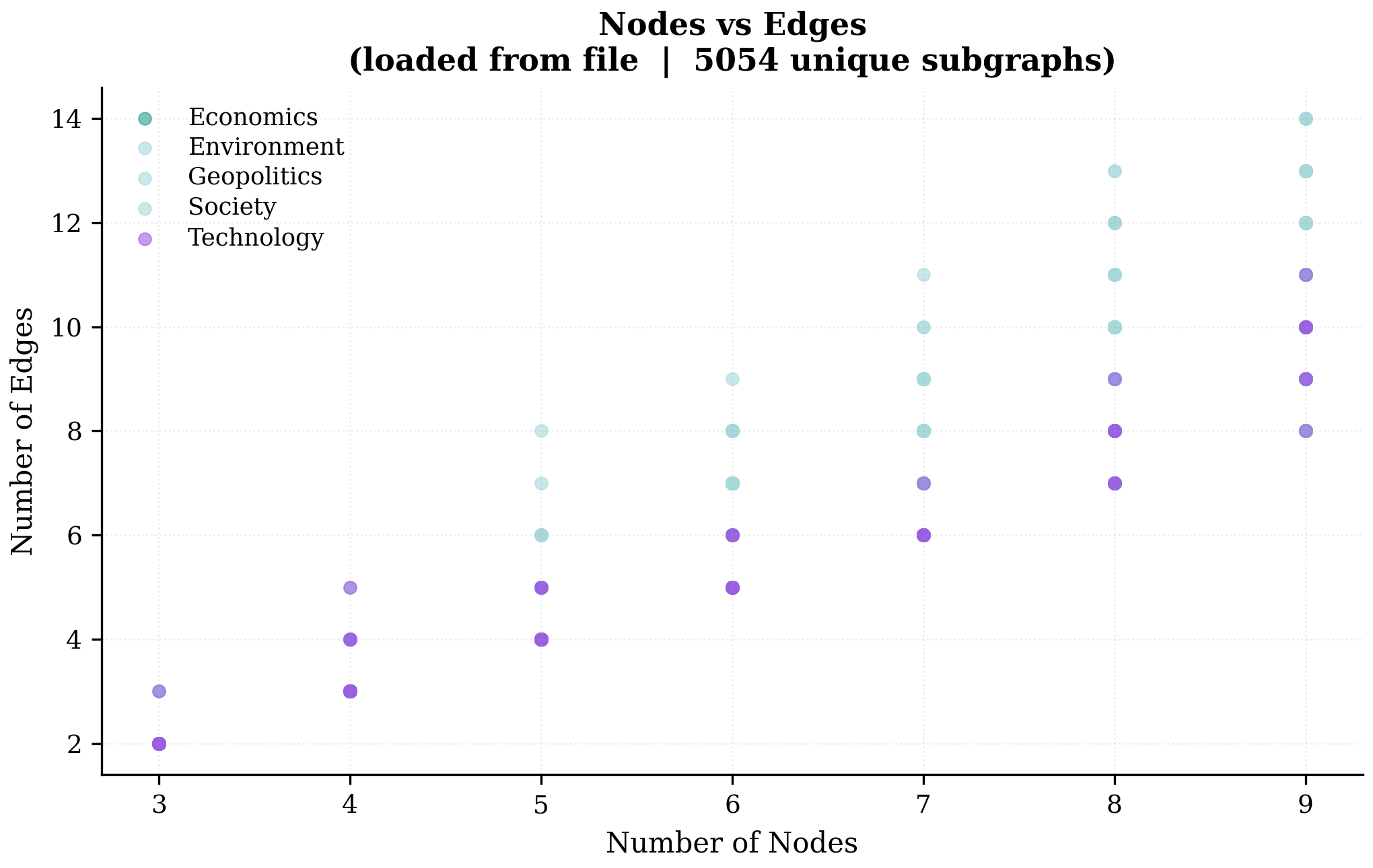}
    \caption{Joint (nodes, edges) distribution coloured by domain. Technology
    occupies the lower-left, while the four larger domains span the full range
    up to 9 nodes and 14 edges. The lower envelope follows the connectedness
    constraint $e \geq n-1$.}
    \label{fig:nodes_vs_edges}
\end{figure*}

The joint relationship between node and edge count, broken out by domain,
appears in Figure~\ref{fig:nodes_vs_edges}. Technology occupies the
lower-left (smaller, sparser subgraphs), while the four larger domains span
the full range up to 9 nodes and 14 edges. The lower envelope follows the
connectedness constraint $e \geq n-1$.

\begin{figure*}[t]
    \centering
    \includegraphics[width=0.85\linewidth]{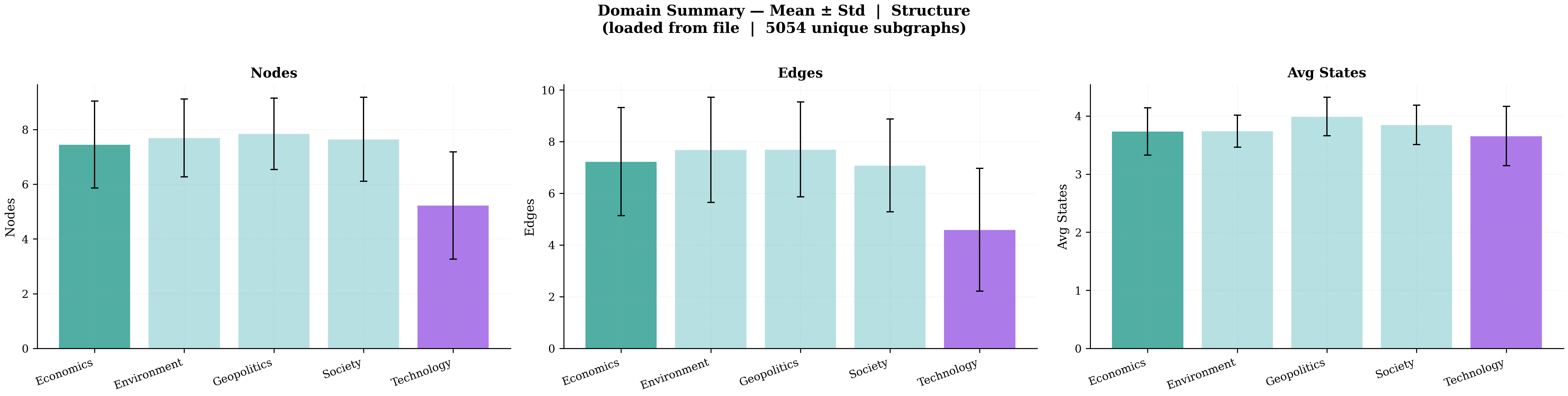}
    \caption{Domain-level structural means with $\pm 1\sigma$ bars (nodes,
    edges, depth). The four largest domains are tightly clustered; Technology
    is the structural outlier on every axis.}
    \label{fig:domain_bars_structure}
\end{figure*}

\begin{figure*}[t]
    \centering
    \includegraphics[width=0.85\linewidth]{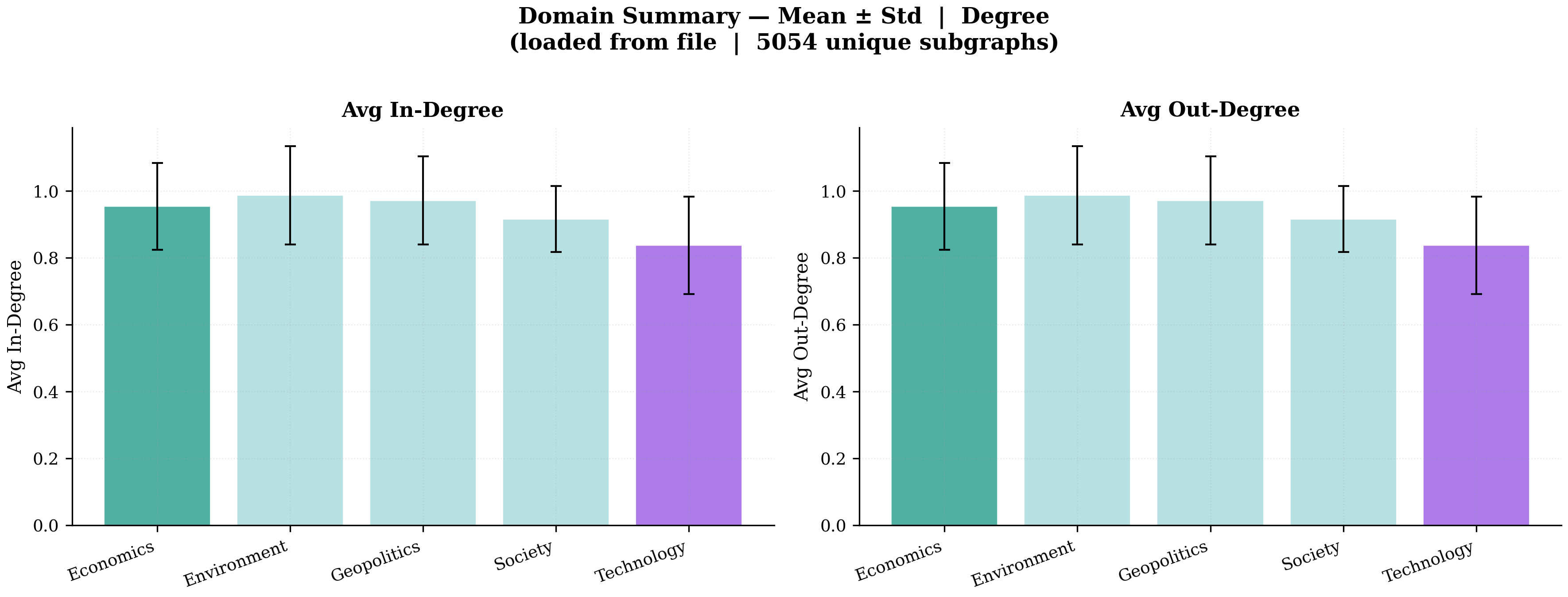}
    \caption{Domain-level degree means with $\pm 1\sigma$ bars (in-degree,
    out-degree). Degree statistics are remarkably uniform across the four
    largest domains, with Technology slightly lower.}
    \label{fig:domain_bars_degree}
\end{figure*}

Domain-level structural means and degree statistics are reported in
Figures~\ref{fig:domain_bars_structure} and~\ref{fig:domain_bars_degree}. The
four largest domains are tightly clustered on all structural dimensions;
Technology is the consistent outlier, with smaller, shallower, and
lower-degree subgraphs.

\begin{figure*}[t]
    \centering
    \includegraphics[width=\textwidth]{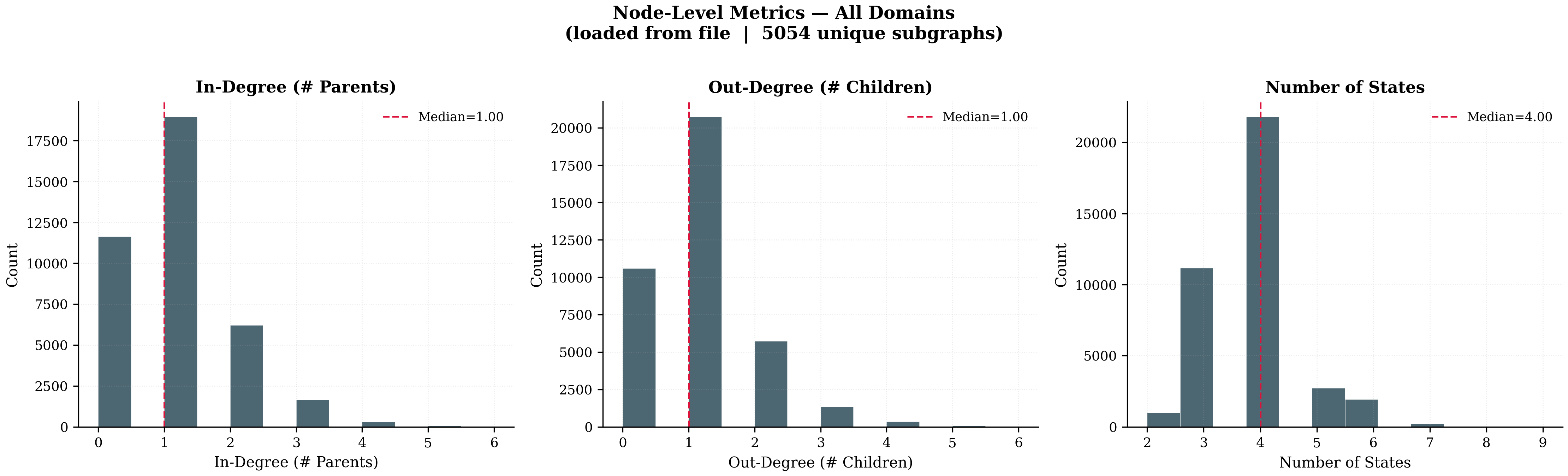}
    \caption{Node-level histograms across the 38{,}820 records: in-degree,
    out-degree, and state cardinality. Roughly 30\% of nodes are roots and
    under 5\% have three or more parents, which permits exact analytic CPD
    recovery for the overwhelming majority of nodes.}
    \label{fig:node_overview_full}
\end{figure*}

Per-node statistics across the full 38{,}820-record corpus appear in
Figure~\ref{fig:node_overview_full}. Roughly 30\% of nodes are roots, and under
5\% have three or more parents, the property that permits exact analytic CPD
recovery for the overwhelming majority of nodes (\S\ref{sec:cpd}).

\begin{figure*}[t]
    \centering
    \includegraphics[width=\textwidth]{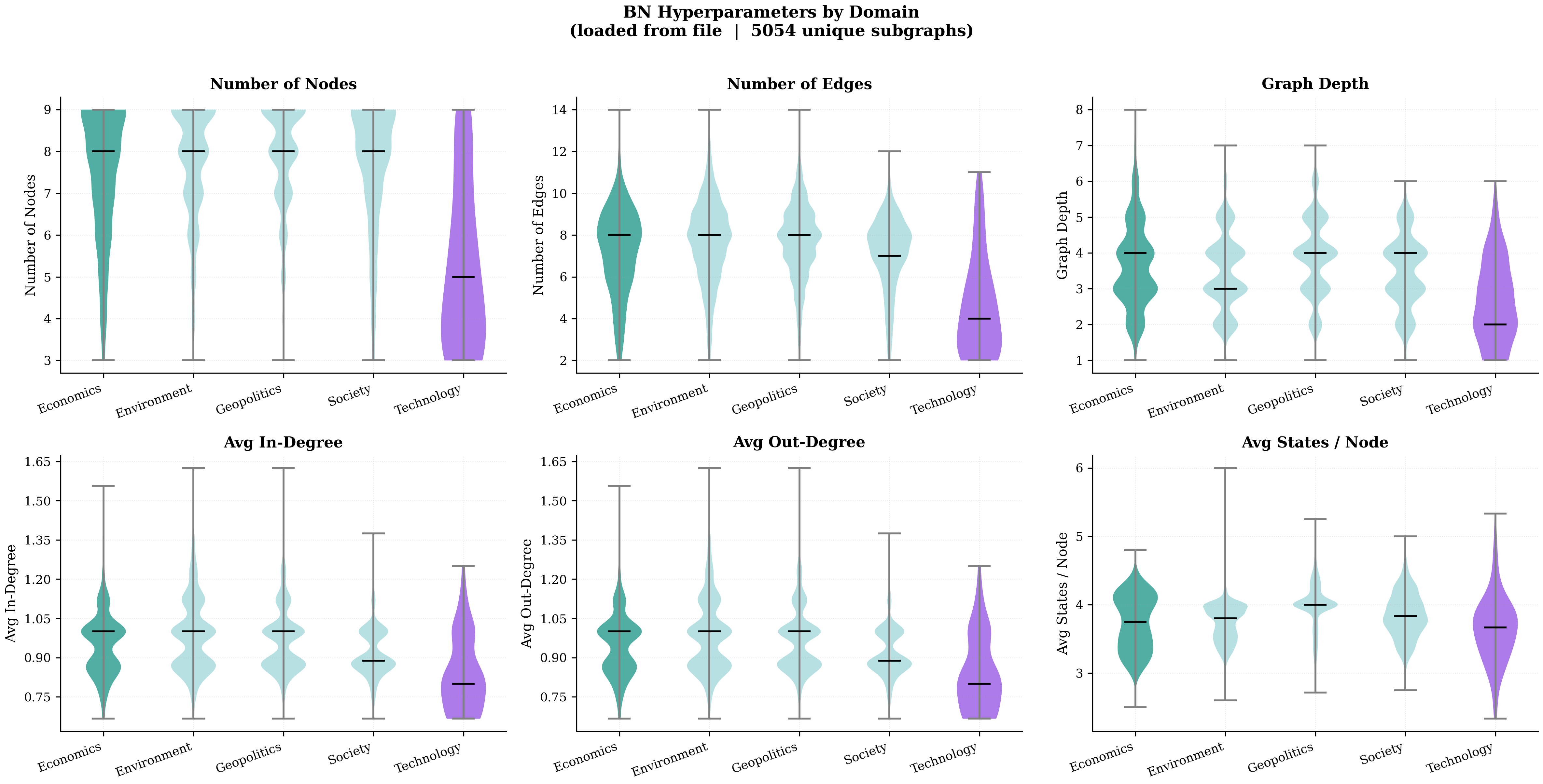}
    \caption{BN-level hyperparameter distributions, faceted by domain. The four
    largest domains are nearly indistinguishable on nodes, edges, and degree,
    with visible domain effects on depth and average states. Technology
    distributions are flatter and shifted toward smaller values throughout.}
    \label{fig:bn_by_domain}
\end{figure*}

\begin{figure*}[t]
    \centering
    \includegraphics[width=\textwidth]{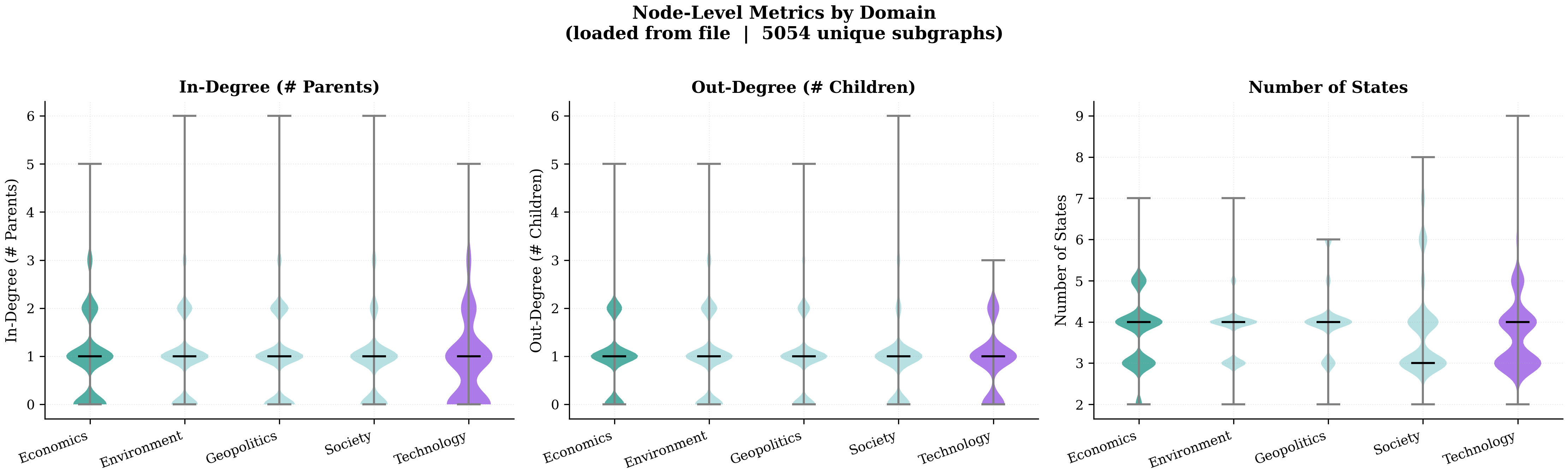}
    \caption{Node-level metric distributions, faceted by domain. Degree
    distributions are nearly identical across the four largest domains. State
    cardinality is most tightly modal at 4 for Environment and Geopolitics and
    most variable for Technology.}
    \label{fig:node_by_domain}
\end{figure*}

Figures~\ref{fig:bn_by_domain} and~\ref{fig:node_by_domain} present the same
BN-level and node-level distributions faceted by domain. The four largest
domains are nearly indistinguishable on nodes, edges, and degree, with
visible differences only on depth and average states. Technology distributions
are flatter and shifted toward smaller values throughout.

\section{Prompt Templates}
\label{app:prompts}

All generators receive the same structured templates and must return only valid
JSON. Malformed responses are retried up to three times before the example is
skipped.

\textbf{Phase 1: Node extraction.} \textit{System:} ``You are an expert at
identifying nodes (variables) in Bayesian Networks from natural language.
Extract only the node names. Return ONLY valid JSON.'' \textit{User:} ``Extract
all Bayesian Network nodes (variables) from this text. Text: \texttt{\{text\}}.
Return ONLY: \texttt{\{"nodes": ["node1", \ldots]\}}. Use exact names from
the text. Do not include states, only node names.''

\textbf{Phase 2: State extraction.} \textit{System:} ``You are an expert at
identifying states of BN nodes from natural language. Always include a `None'
state for every node. Return ONLY valid JSON.'' \textit{User:} ``For each node,
extract all possible states. Text: \texttt{\{text\}}. Nodes:
\texttt{\{nodes\}}. Return ONLY valid JSON with node-state lists.''

\textbf{Phase 3: Edge extraction.} \textit{System:} ``You are an expert at
identifying directed causal relationships between BN nodes. Return ONLY valid
JSON.'' \textit{User:} ``Identify directed causal edges ($A \to B$ means $A$
causes $B$). Text: \texttt{\{text\}}. Available nodes and states:
\texttt{\{node\_state\_list\}}. Only use listed node names; only include edges
supported by the text.''

\textbf{Phase 4: CPD extraction.} \textit{System:} ``You are an expert at
extracting CPDs from natural language. Rows = child states, columns = parent
states. Each column must sum to 1.0. Return ONLY valid JSON.'' \textit{User:}
``Extract the CPD matrix. Parent \texttt{\{parent\}} states (columns):
\texttt{\{parent\_states\}}. Child \texttt{\{child\}} states (rows):
\texttt{\{child\_states\}}.'' Phase 4 is invoked only for edges correctly
predicted in Phase 3, isolating numerical estimation error from structural
error.

\section{Judge Prompt Templates}
\label{app:judge}

The semantic judge receives no information about which generator produced a
prediction. Scoring is binary (1 = same meaning or clear paraphrase; 0 = no
match) with one-to-one matching enforced.

\textbf{Node judge.} ``Match predicted BN node names to reference node names.
Scoring: 1 = same meaning or clear paraphrase, 0 = no match. One-to-one only.
Return ONLY valid JSON.''

\textbf{State judge.} ``Match predicted states to reference states for a BN
node. \texttt{None} always maps to \texttt{None}. Scoring: 1 = same meaning or
clear paraphrase, 0 = no match. One-to-one only. Return ONLY valid JSON.''

\section{Implementation Details}
\label{app:supplementary}

\textbf{CPD recovery.} For a single-parent edge $A \rightarrow B$,
\[
P(B{=}b\mid A{=}a)=\frac{P(A{=}a,B{=}b)}{\sum_{b'}P(A{=}a,B{=}b')}.
\]
For a child $C$ with parents $A_1,\ldots,A_k$, where $k\in\{2,3\}$,
\begin{equation*}
\begin{aligned}
&P(C{=}c\mid A_1{=}a_1,\ldots,A_k{=}a_k) \\
&\quad =
\frac{P(C{=}c,A_1{=}a_1,\ldots,A_k{=}a_k)}
{\sum_{c'}P(C{=}c',A_1{=}a_1,\ldots,A_k{=}a_k)}.
\end{aligned}
\end{equation*}
For $k\geq4$, PRISM uses the Naive Bayes fallback
\begin{equation*}
\begin{aligned}
&P(C{=}c\mid A_1{=}a_1,\ldots,A_k{=}a_k) \\
&\quad \propto
\frac{\prod_{i=1}^{k}P(C{=}c\mid A_i{=}a_i)}{P(C{=}c)^{k-1}}.
\end{aligned}
\end{equation*}

\textbf{Metric definitions.} Precision, recall, and $F_1$ use
$P=n_{\mathrm{corr}}/n_{\mathrm{pred}}$, $R=n_{\mathrm{corr}}/n_{\mathrm{ref}}$,
and $F_1=2PR/(P+R)$, with 0 returned for undefined ratios. For Phase 1,
$n_{\mathrm{pred}}$ is reduced by semantic duplicates before computing
precision. State $F_1$ is macro-averaged over matched nodes. CPD-KL is computed
column-wise with $\varepsilon=10^{-6}$ Laplace smoothing; the per-edge score is
the mean across aligned parent-state columns. We also report
$\tfrac{1}{2}(\mathrm{KL}(p\|\tilde p)+\mathrm{KL}(\hat p\|\tilde q))$ as a
symmetric variant.

\textbf{Fallback scoring.} If the judge fails to return valid JSON after three
retries, state matching falls back to exact case-insensitive string comparison.
Node and edge matching have no fallback; affected examples are flagged and
excluded from aggregate statistics.

\textbf{Model access.} Claude Haiku 4.5 was accessed via the Anthropic API.
DeepSeek-V3, Gemma 3 12B, Qwen3-30B-A3B, Llama 4 Maverick, and the Llama 3.3
70B judge were accessed via the HuggingFace Inference API on the Novita,
Featherless-AI, and Groq backends. GPT-4o-mini was accessed via the OpenAI API.

\section{Per-Phase Precision and Recall}
\label{app:precision_recall}

Table~\ref{tab:precision_recall} reports the underlying precision and recall
values for the Node, State, and Edge $F_1$ scores summarized in
\S\ref{sec:results}. Node recall is near ceiling across all six generators
(0.972--0.987), while node precision varies by almost a factor of two
(0.395--0.711), confirming that node over-generation is the dominant structural
failure mode. State and edge metrics are conditional on Phase~1 node alignment.

\begin{table*}[t]
\centering
\small
\caption{Per-phase precision and recall on PRISM-BN. Node recall is near
ceiling for all models; node precision drives most differences in node $F_1$.
State and edge metrics are conditional on Phase~1 node alignment.}
\label{tab:precision_recall}
\begin{tabular}{l ccc ccc ccc}
\toprule
& \multicolumn{3}{c}{\textbf{Node}} & \multicolumn{3}{c}{\textbf{State}}
& \multicolumn{3}{c}{\textbf{Edge}} \\
\cmidrule(lr){2-4} \cmidrule(lr){5-7} \cmidrule(lr){8-10}
\textbf{Model} & P & R & $F_1$ & P & R & $F_1$ & P & R & $F_1$ \\
\midrule
DeepSeek-V3      & 0.711 & 0.985 & 0.826 & 0.861 & 0.982 & 0.918 & 0.972 & 0.963 & 0.966 \\
Claude Haiku 4.5 & 0.699 & 0.978 & 0.816 & 0.883 & 0.983 & 0.931 & 0.965 & 0.953 & 0.959 \\
Qwen3-30B-A3B    & 0.586 & 0.977 & 0.737 & 0.846 & 0.978 & 0.907 & 0.950 & 0.929 & 0.939 \\
Llama 4 Maverick & 0.451 & 0.987 & 0.618 & 0.809 & 0.979 & 0.886 & 0.964 & 0.957 & 0.960 \\
GPT-4o-mini      & 0.408 & 0.972 & 0.573 & 0.845 & 0.970 & 0.903 & 0.889 & 0.918 & 0.902 \\
Gemma 3 12B      & 0.395 & 0.974 & 0.566 & 0.826 & 0.972 & 0.893 & 0.945 & 0.920 & 0.932 \\
\bottomrule
\end{tabular}
\end{table*}

\begin{figure*}[t]
\centering
\begin{minipage}{0.32\textwidth}\centering
  \includegraphics[width=\linewidth]{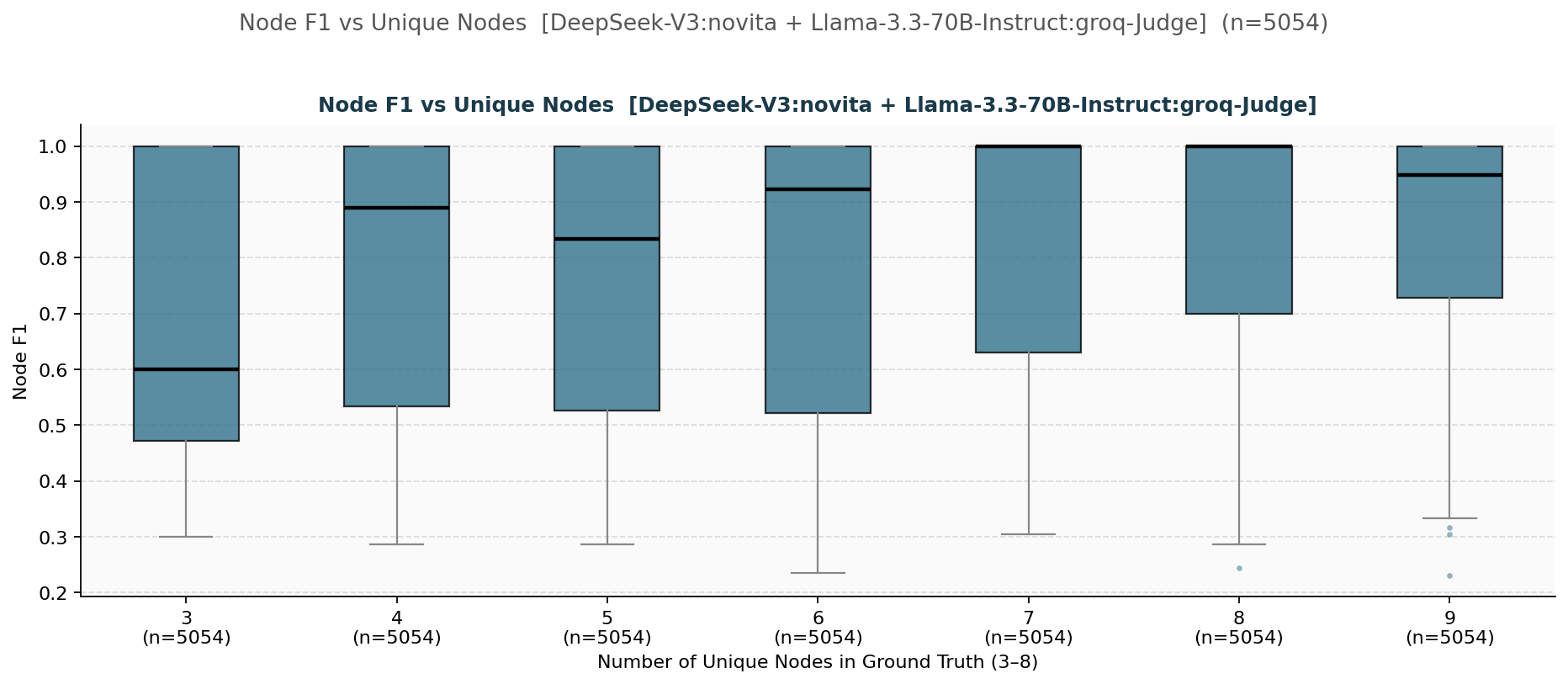}
  \caption*{DeepSeek-V3}\end{minipage}
\begin{minipage}{0.32\textwidth}\centering
  \includegraphics[width=\linewidth]{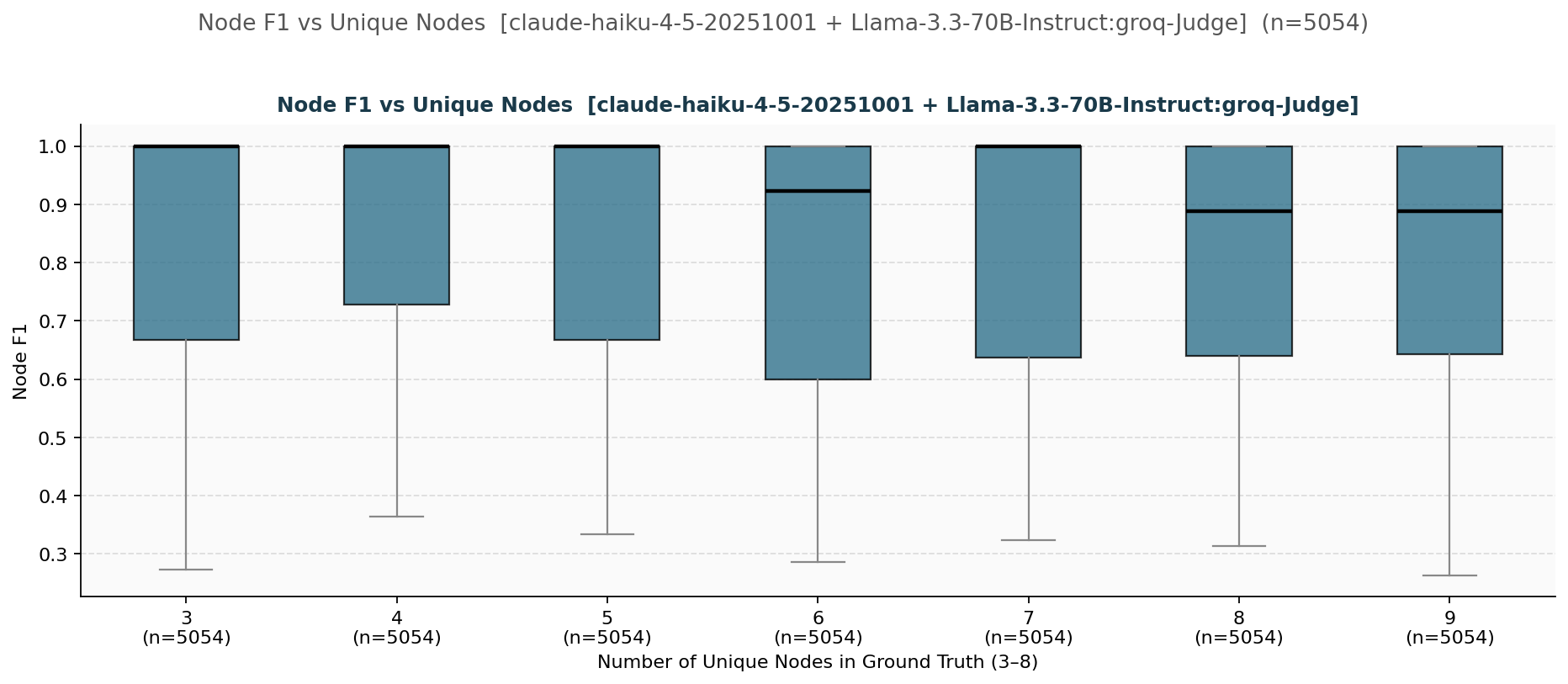}
  \caption*{Claude Haiku 4.5}\end{minipage}
\begin{minipage}{0.32\textwidth}\centering
  \includegraphics[width=\linewidth]{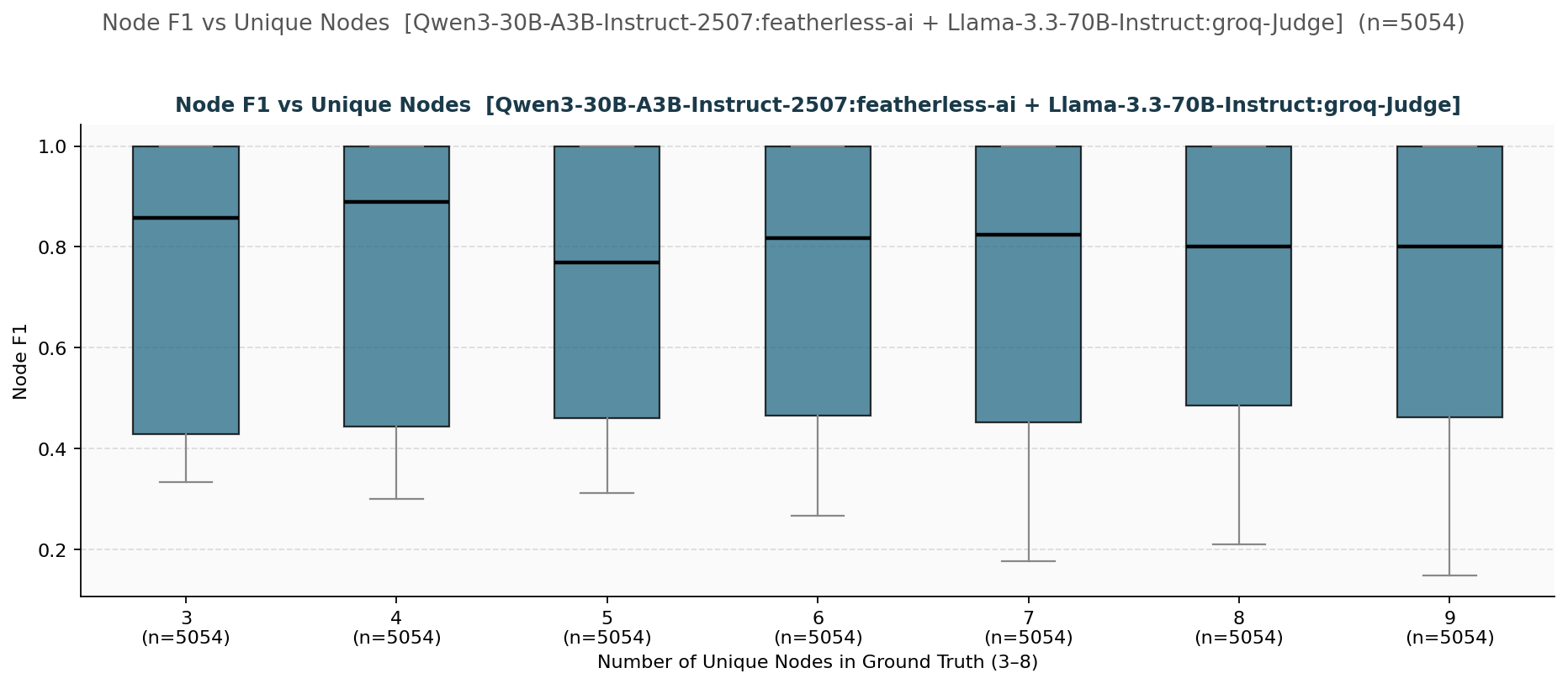}
  \caption*{Qwen3-30B-A3B}\end{minipage}\\[0.5em]
\begin{minipage}{0.32\textwidth}\centering
  \includegraphics[width=\linewidth]{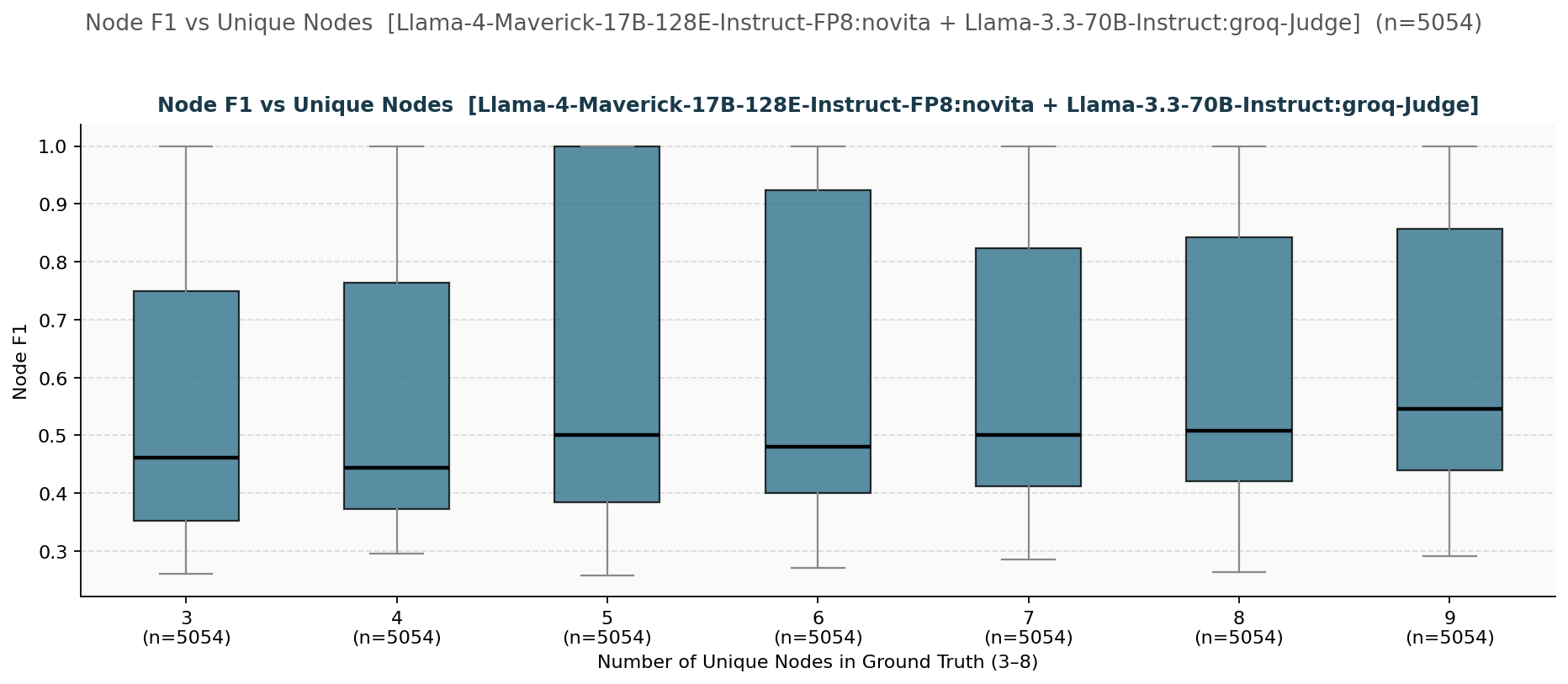}
  \caption*{Llama 4 Maverick}\end{minipage}
\begin{minipage}{0.32\textwidth}\centering
  \includegraphics[width=\linewidth]{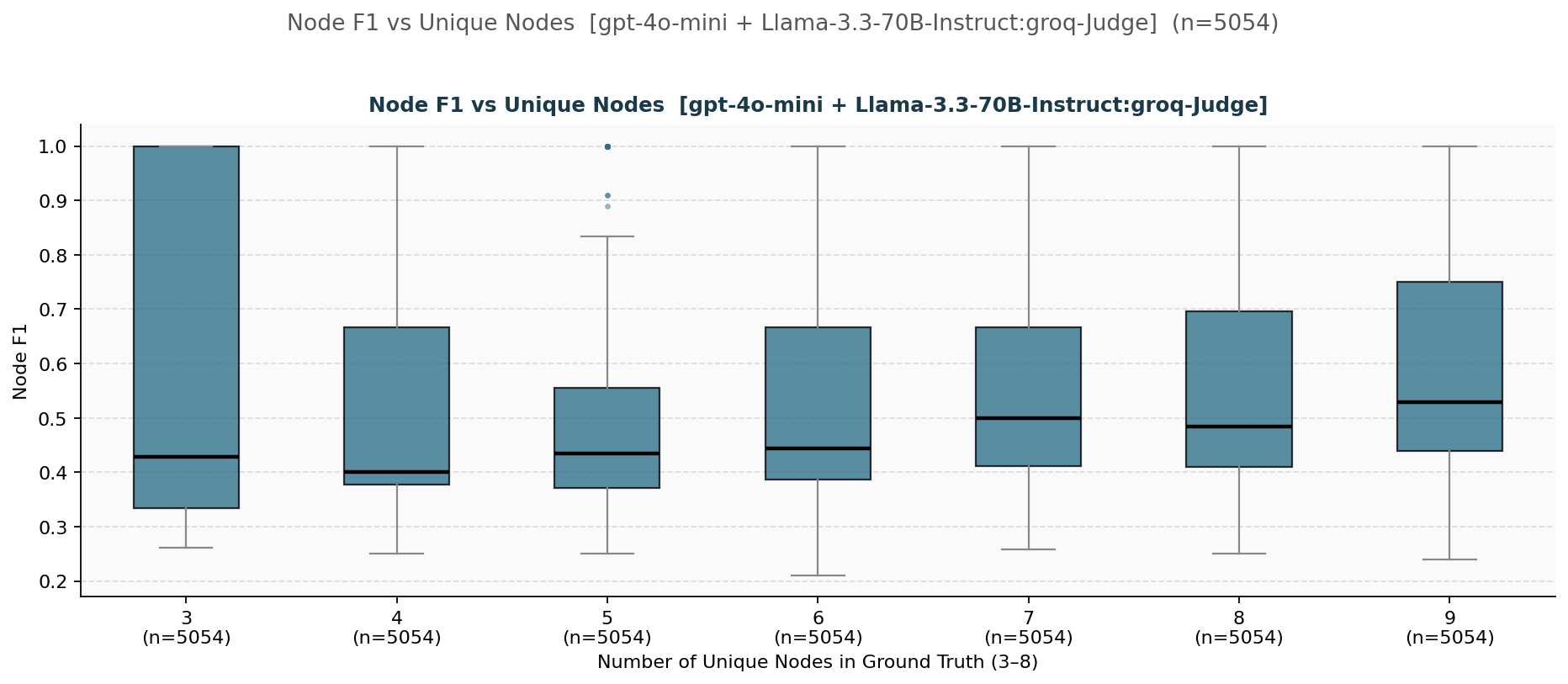}
  \caption*{GPT-4o-mini}\end{minipage}
\begin{minipage}{0.32\textwidth}\centering
  \includegraphics[width=\linewidth]{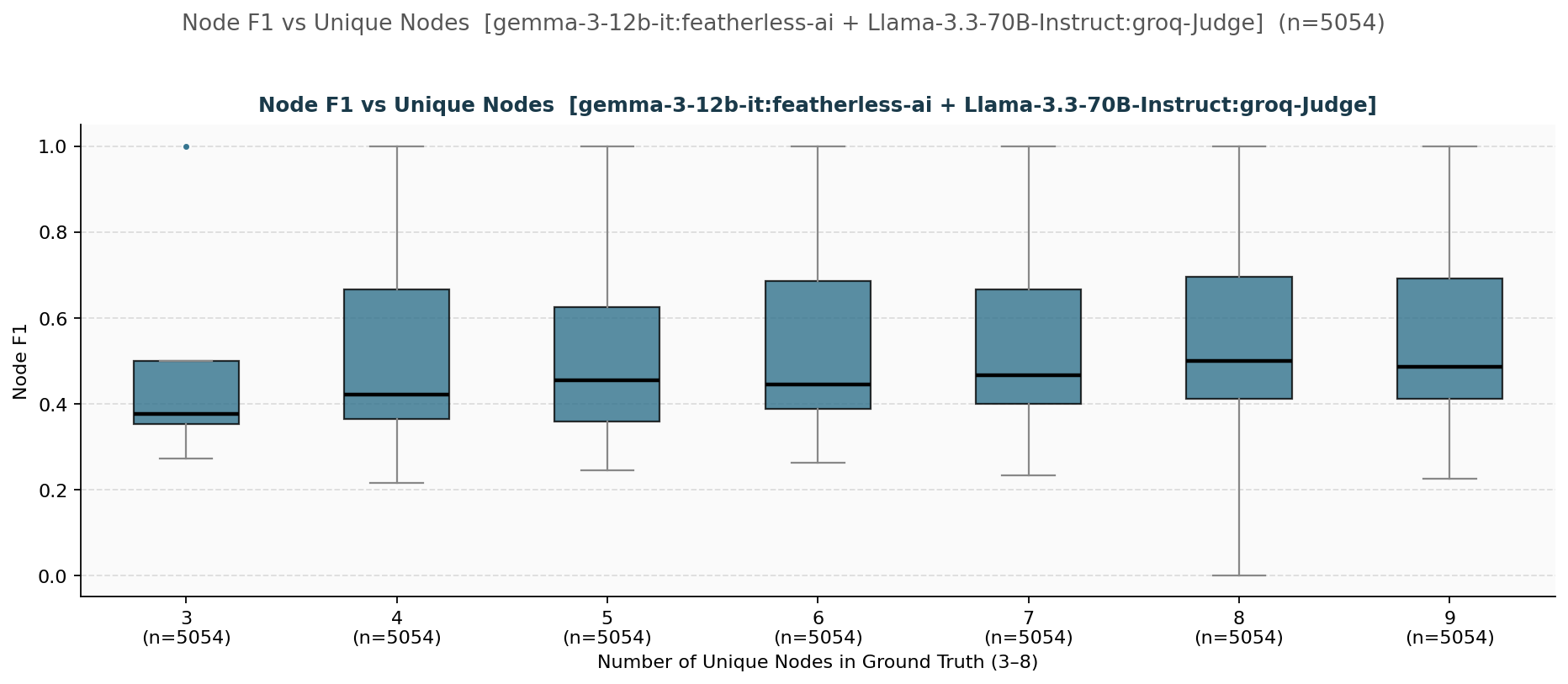}
  \caption*{Gemma 3 12B}\end{minipage}
\caption{Per-model node $F_1$ vs. reference node count. DeepSeek-V3 and Claude
Haiku 4.5 maintain the tightest, highest distributions across the full range.
Qwen3-30B-A3B sits a step below them with similar consistency. Llama 4
Maverick, GPT-4o-mini, and Gemma 3 12B exhibit wider IQRs and lower medians as
node count grows, consistent with the precision-driven failure mode discussed
in \S\ref{sec:results}.}
\label{fig:appendix_node_f1}
\end{figure*}

\begin{figure*}[t]
\centering
\begin{minipage}{0.32\textwidth}\centering
  \includegraphics[width=\linewidth]{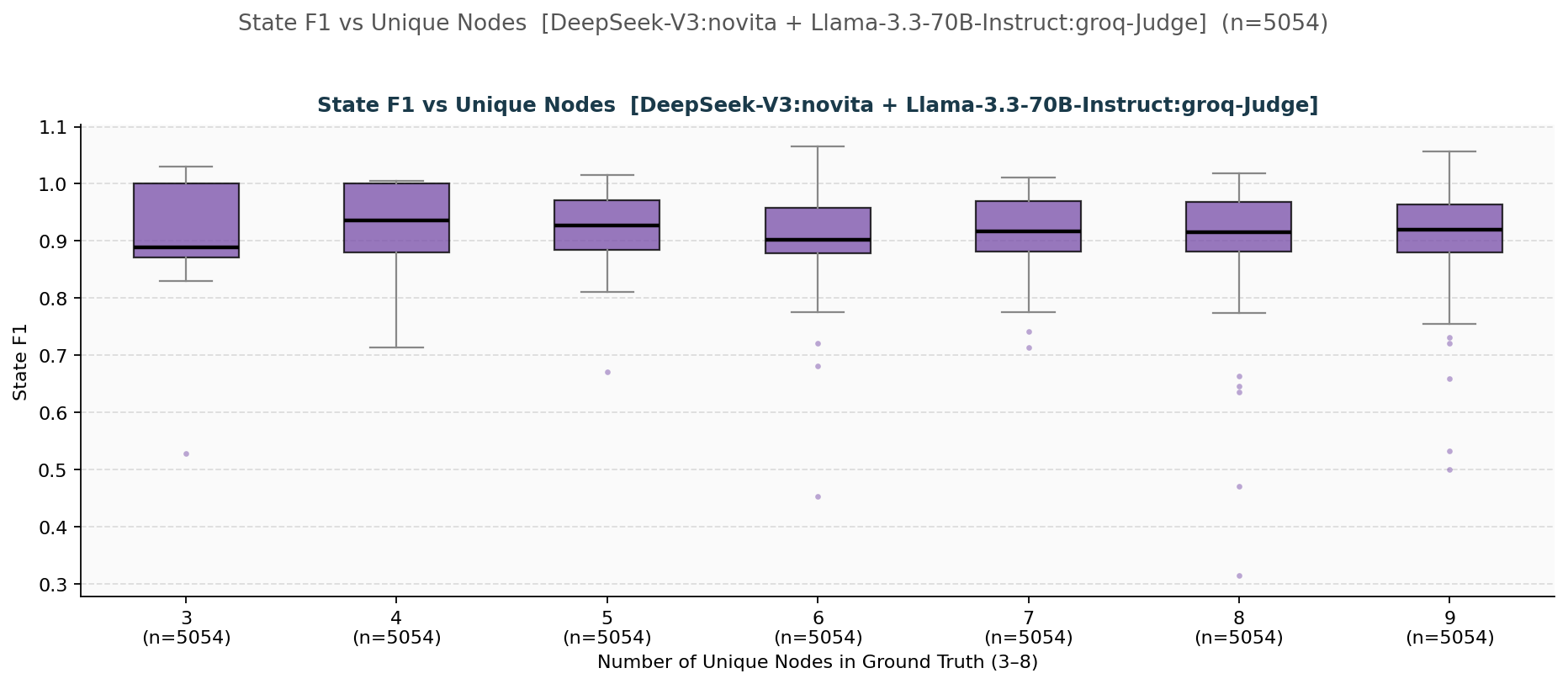}
  \caption*{DeepSeek-V3}\end{minipage}
\begin{minipage}{0.32\textwidth}\centering
  \includegraphics[width=\linewidth]{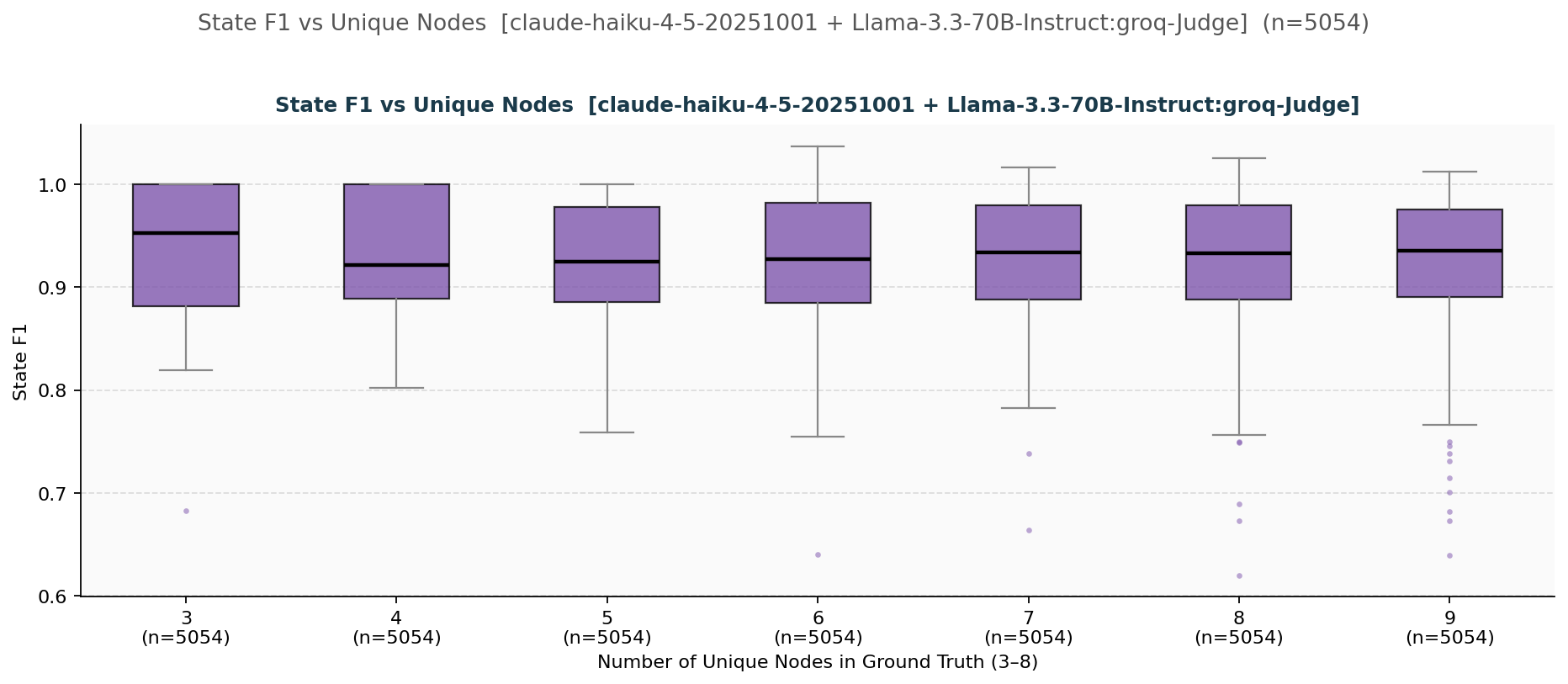}
  \caption*{Claude Haiku 4.5}\end{minipage}
\begin{minipage}{0.32\textwidth}\centering
  \includegraphics[width=\linewidth]{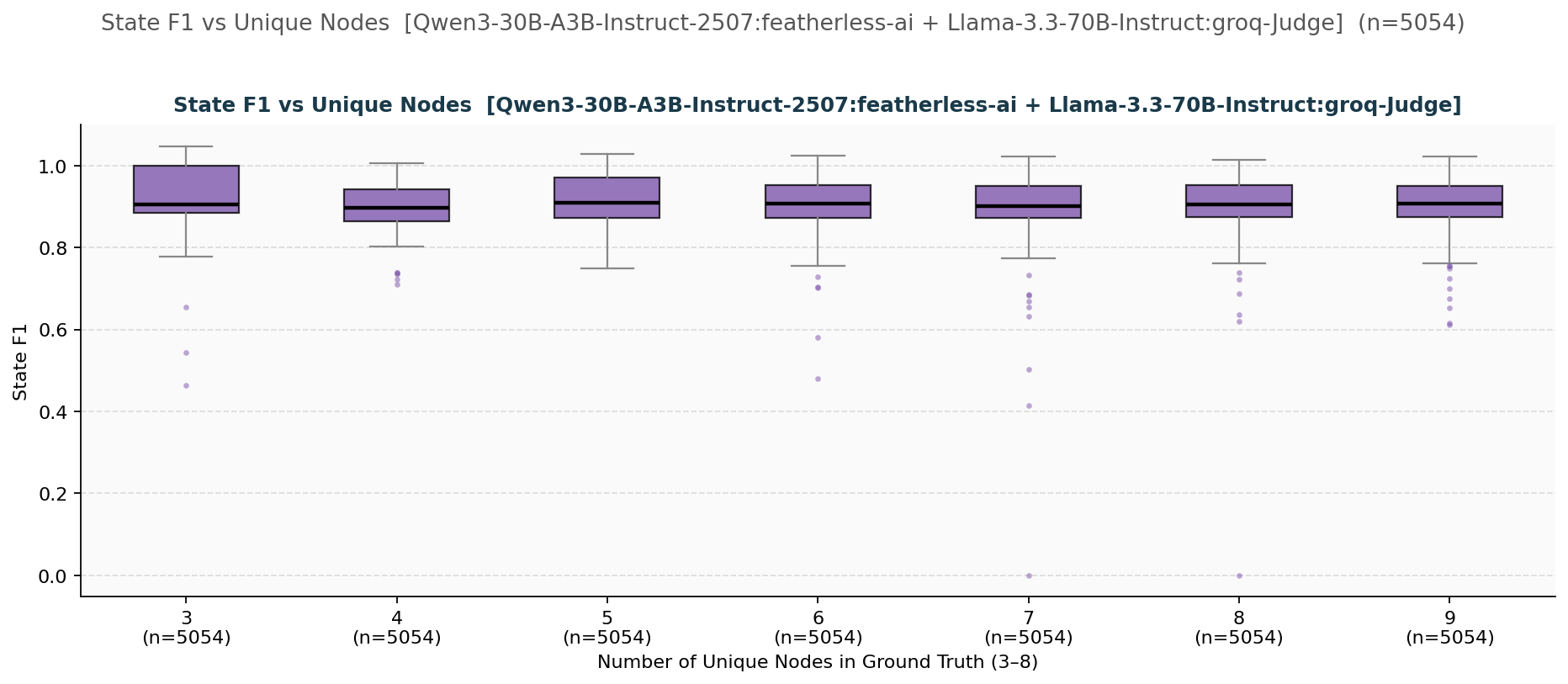}
  \caption*{Qwen3-30B-A3B}\end{minipage}\\[0.5em]
\begin{minipage}{0.32\textwidth}\centering
  \includegraphics[width=\linewidth]{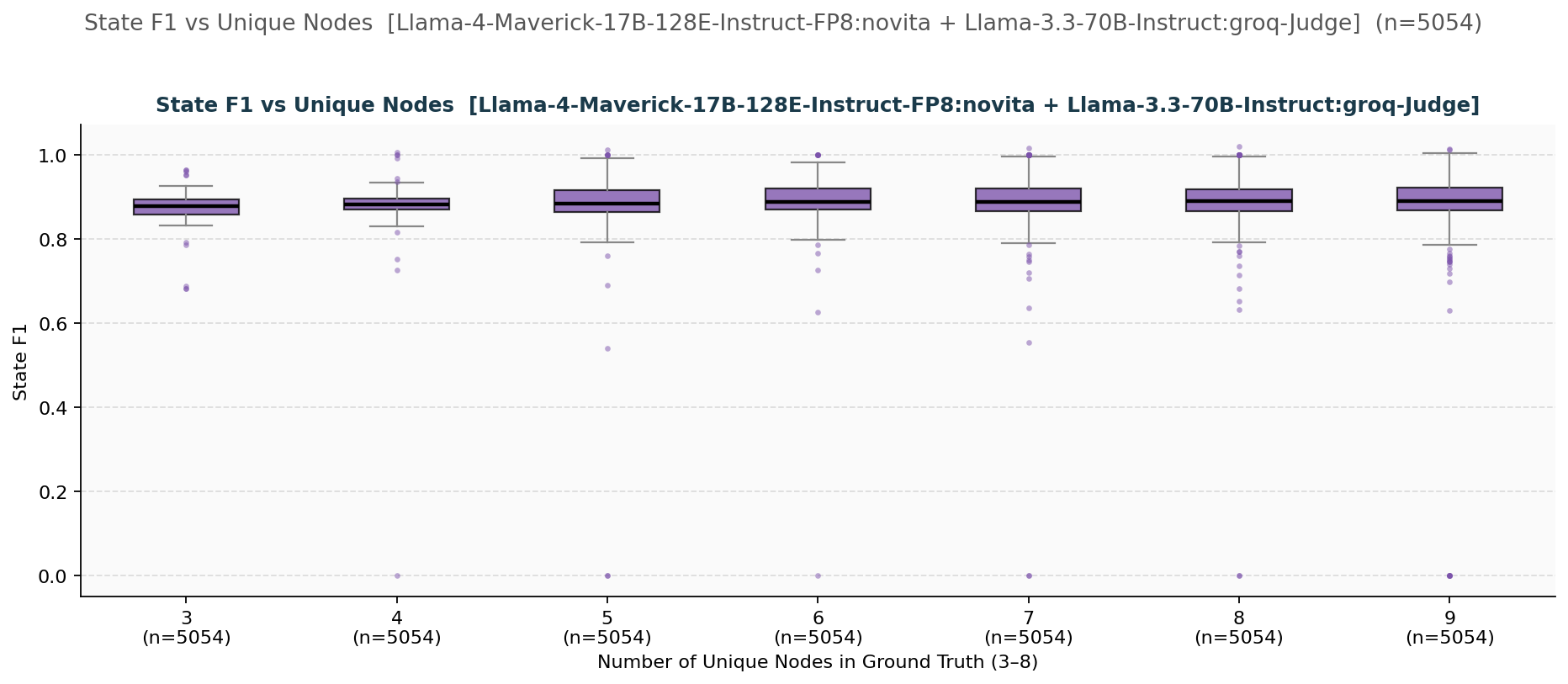}
  \caption*{Llama 4 Maverick}\end{minipage}
\begin{minipage}{0.32\textwidth}\centering
  \includegraphics[width=\linewidth]{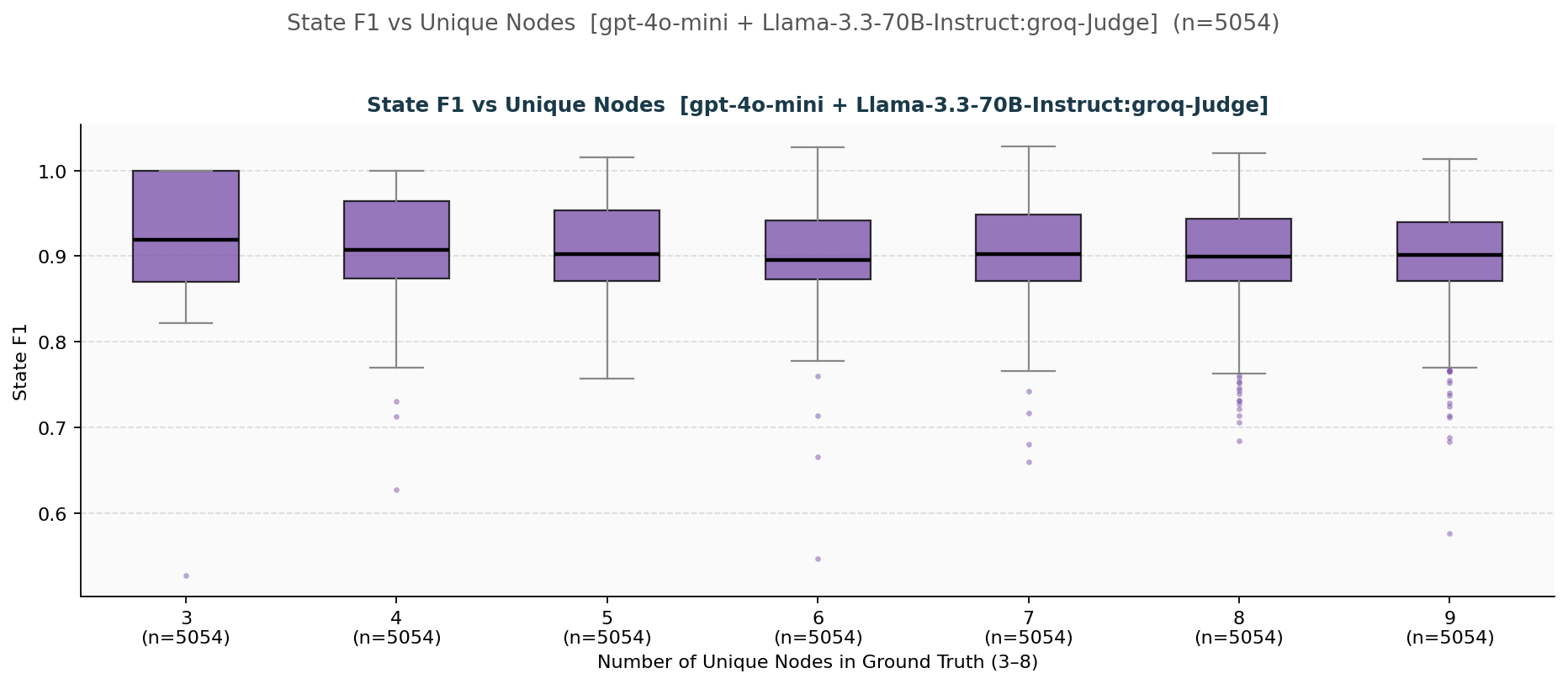}
  \caption*{GPT-4o-mini}\end{minipage}
\begin{minipage}{0.32\textwidth}\centering
  \includegraphics[width=\linewidth]{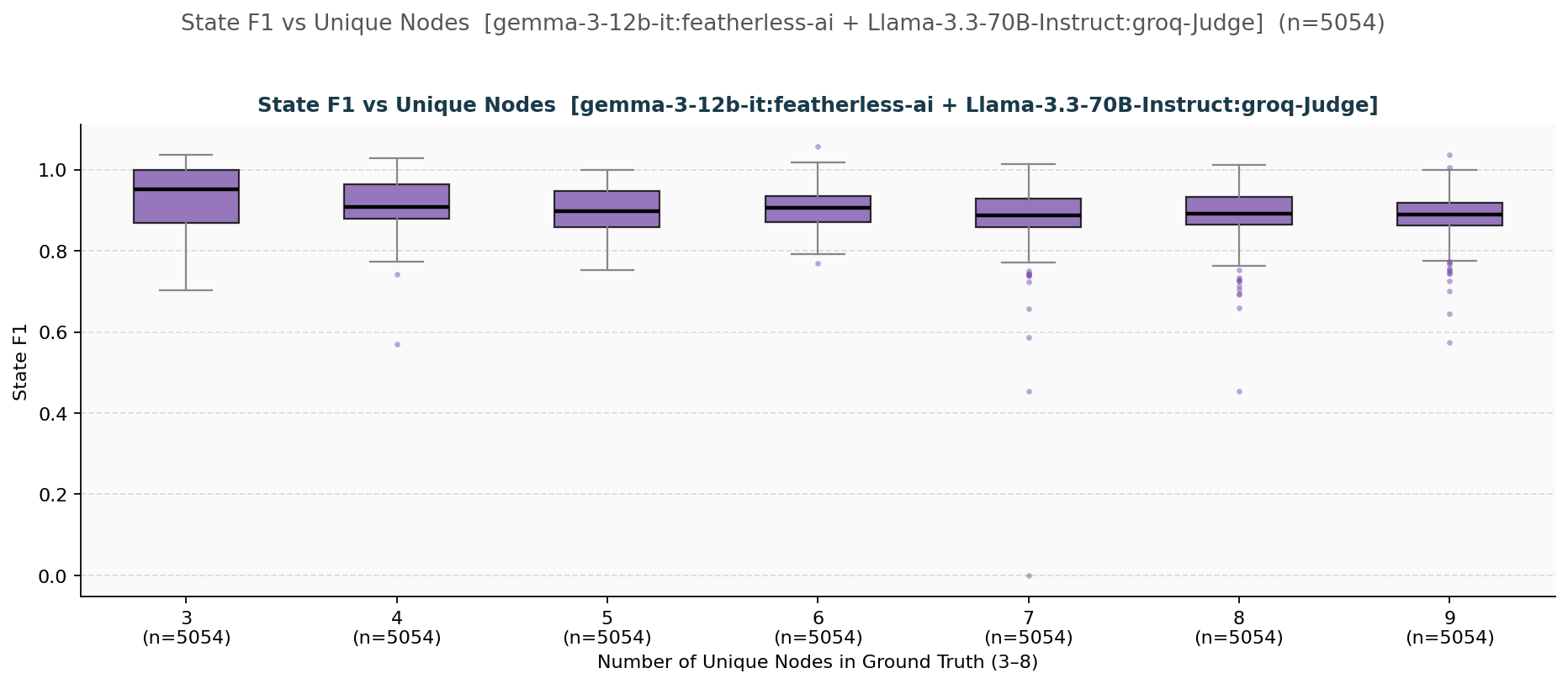}
  \caption*{Gemma 3 12B}\end{minipage}
\caption{Per-model state $F_1$ vs. reference node count. State $F_1$ is
uniformly high across all models and sizes, supporting the observation that
state enumeration is the easiest phase conditional on Phase 1 alignment.}
\label{fig:appendix_state_f1}
\end{figure*}

\begin{figure*}[t]
\centering
\begin{minipage}{0.32\textwidth}\centering
  \includegraphics[width=\linewidth]{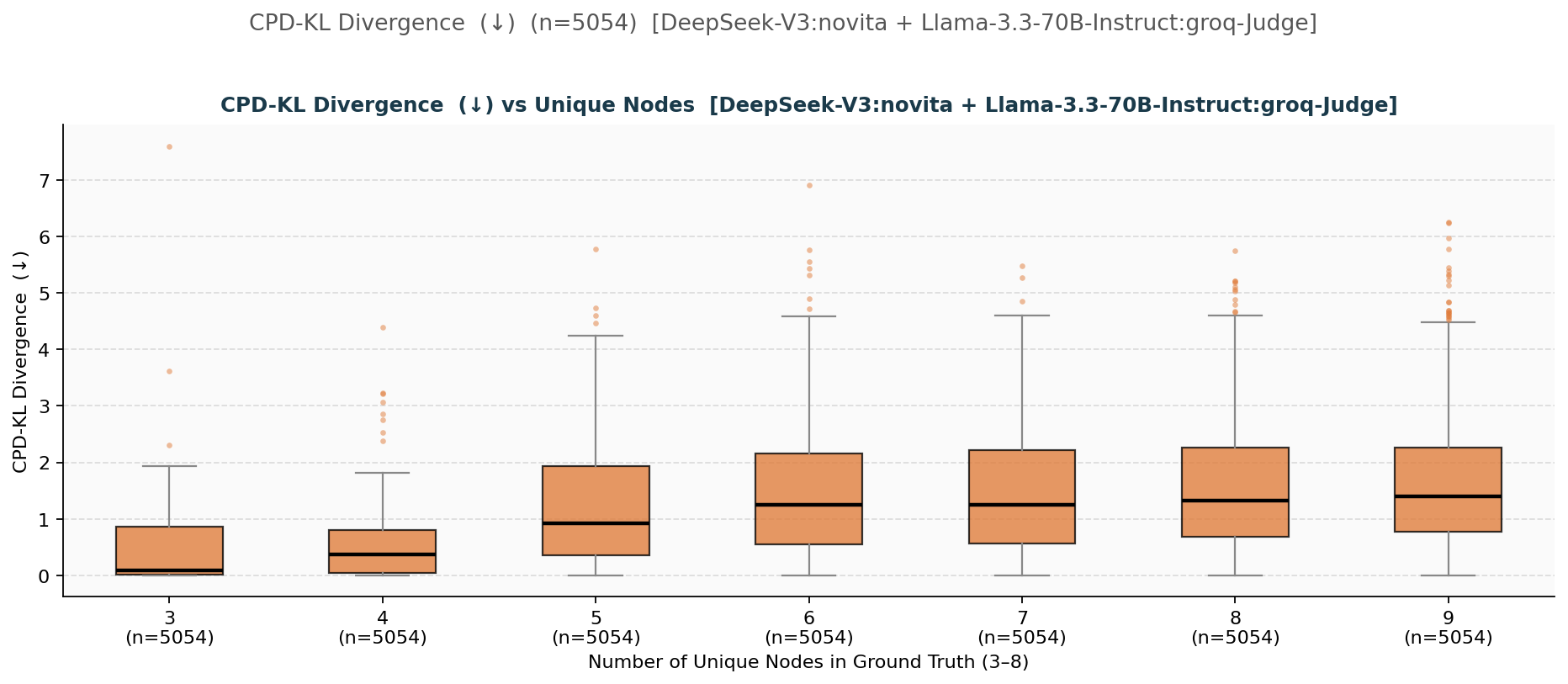}
  \caption*{DeepSeek-V3}\end{minipage}
\begin{minipage}{0.32\textwidth}\centering
  \includegraphics[width=\linewidth]{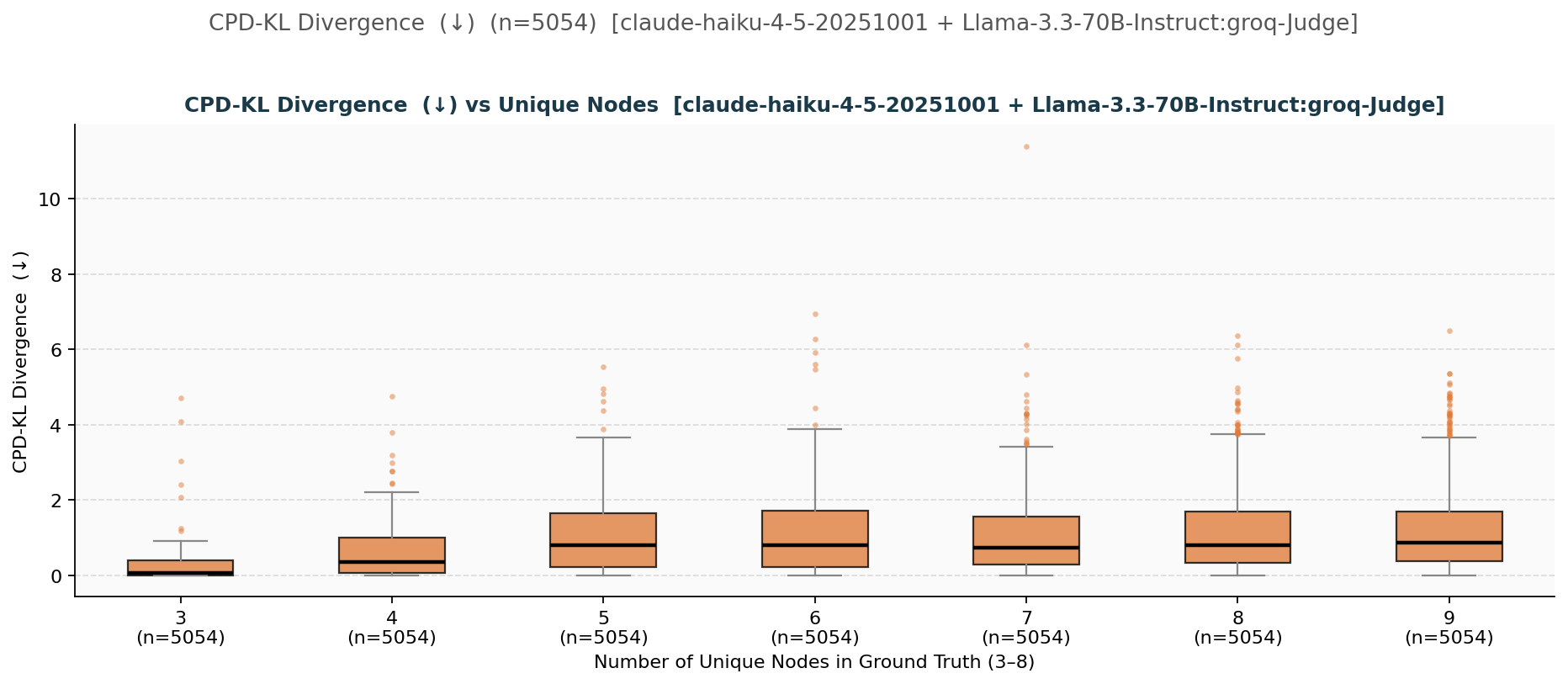}
  \caption*{Claude Haiku 4.5}\end{minipage}
\begin{minipage}{0.32\textwidth}\centering
  \includegraphics[width=\linewidth]{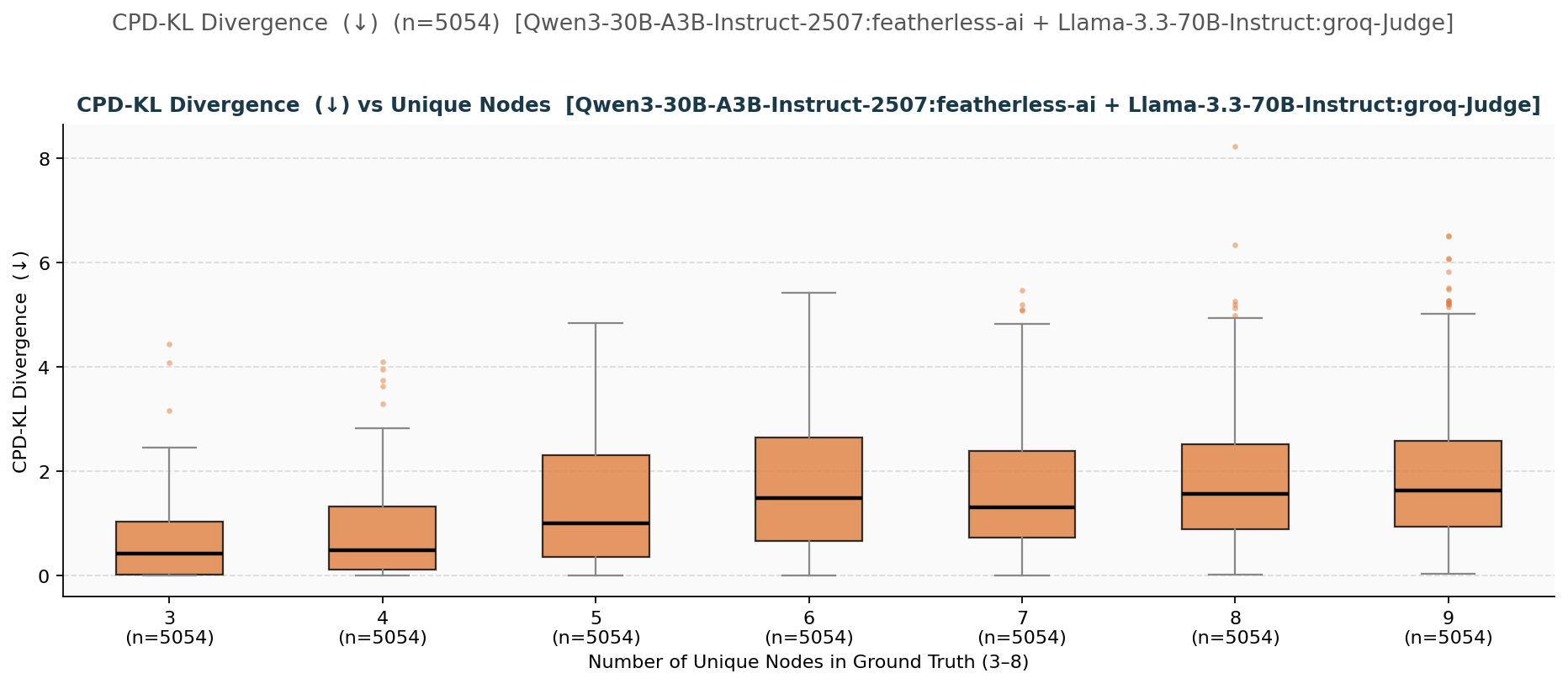}
  \caption*{Qwen3-30B-A3B}\end{minipage}\\[0.5em]
\begin{minipage}{0.32\textwidth}\centering
  \includegraphics[width=\linewidth]{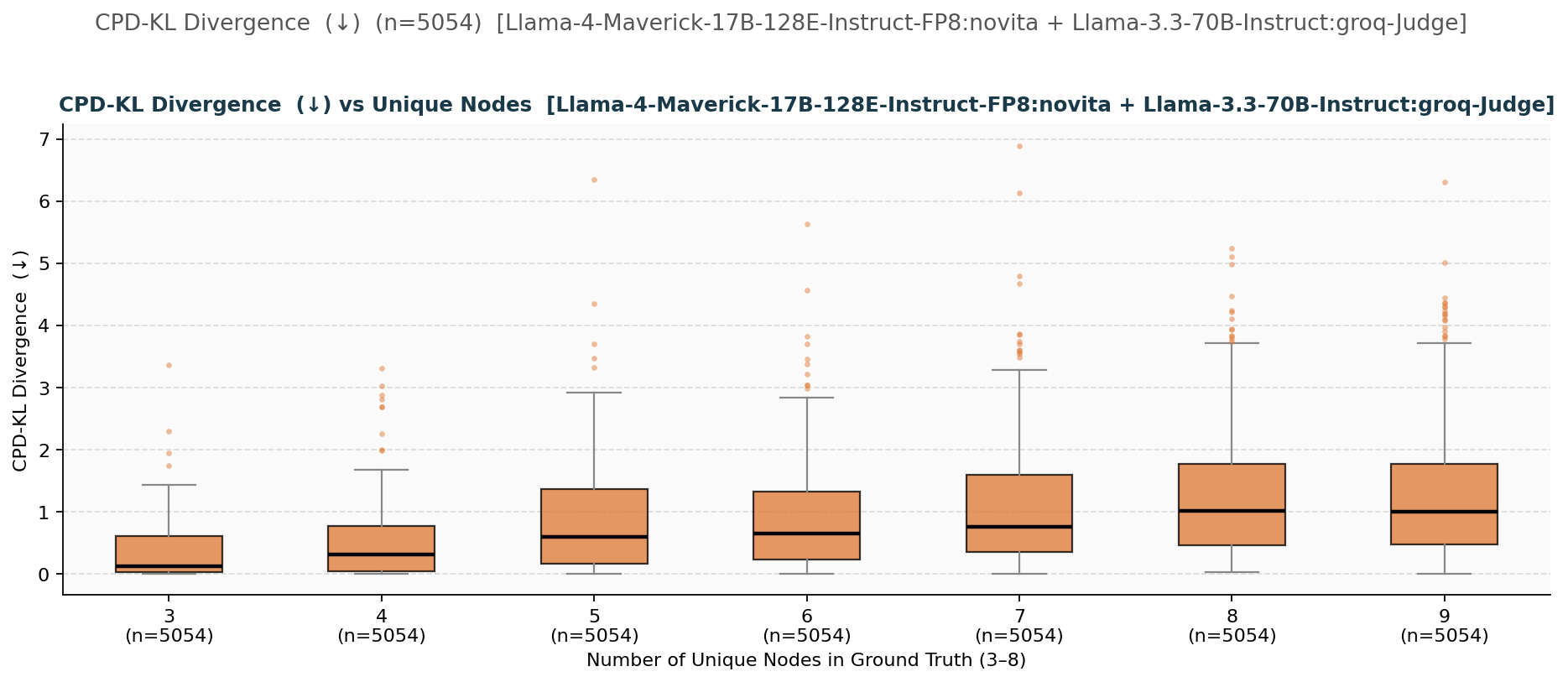}
  \caption*{Llama 4 Maverick}\end{minipage}
\begin{minipage}{0.32\textwidth}\centering
  \includegraphics[width=\linewidth]{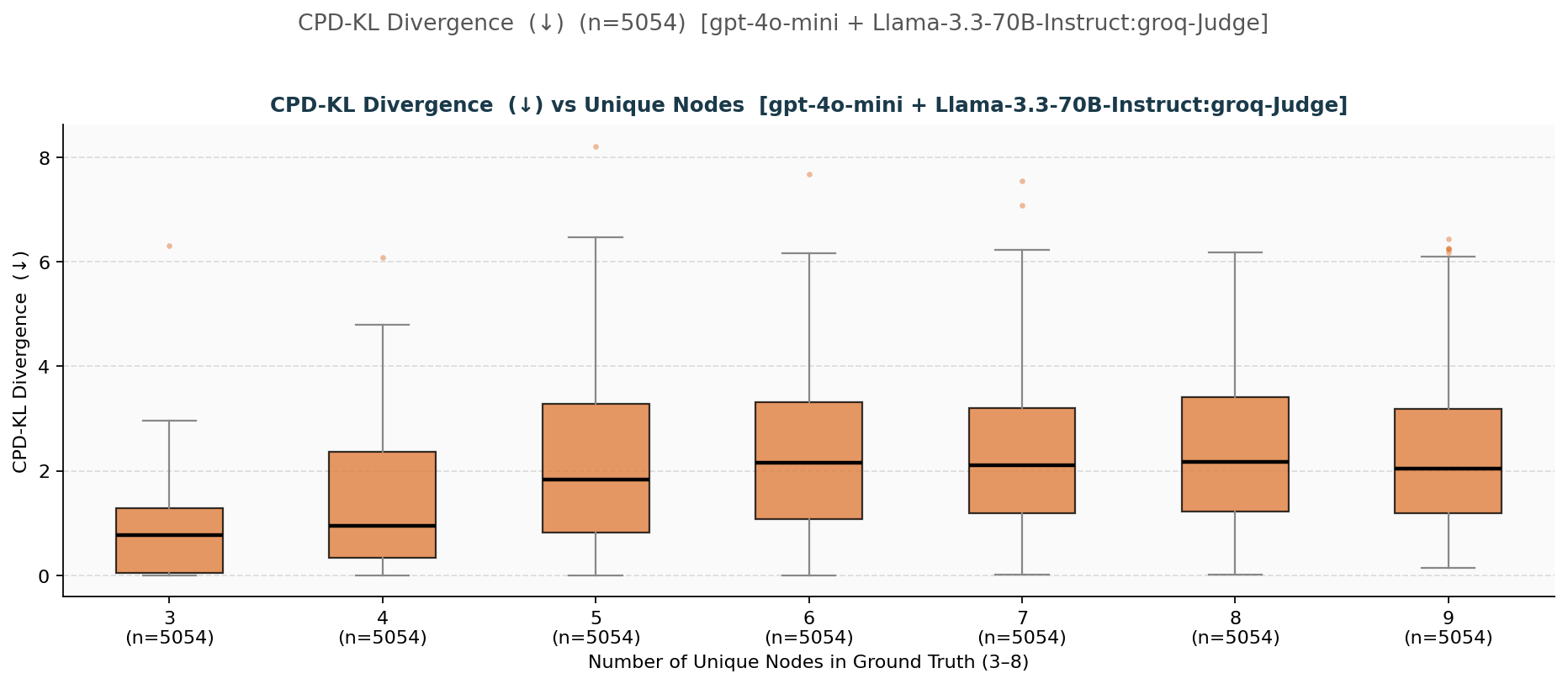}
  \caption*{GPT-4o-mini}\end{minipage}
\begin{minipage}{0.32\textwidth}\centering
  \includegraphics[width=\linewidth]{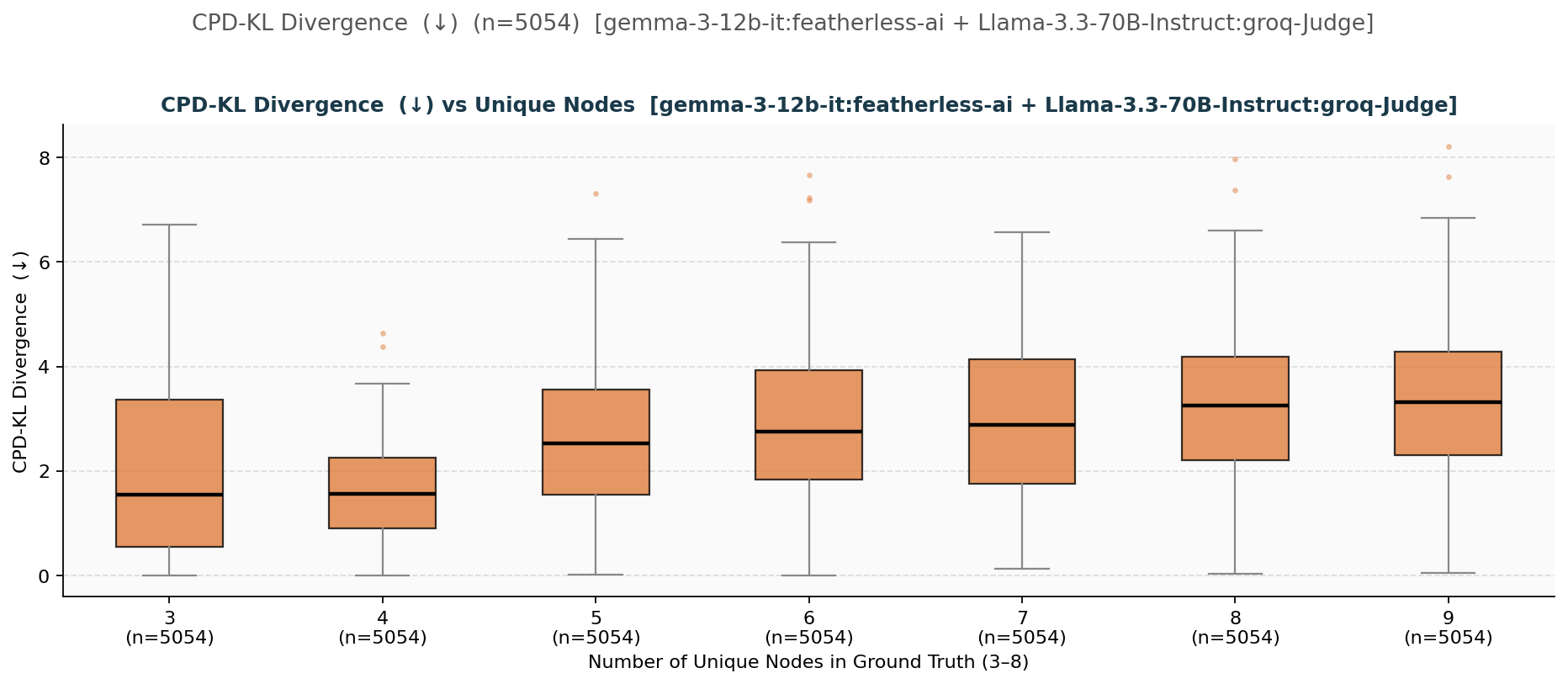}
  \caption*{Gemma 3 12B}\end{minipage}
\caption{Per-model CPD-KL divergence vs. reference node count (lower is
better). Claude Haiku 4.5 attains the lowest medians across the full range,
while Gemma 3 12B exhibits the highest medians and heaviest upper tails.}
\label{fig:appendix_kl}
\end{figure*}

\begin{figure*}[t]
\centering
\begin{minipage}{0.32\textwidth}\centering
  \includegraphics[width=\linewidth]{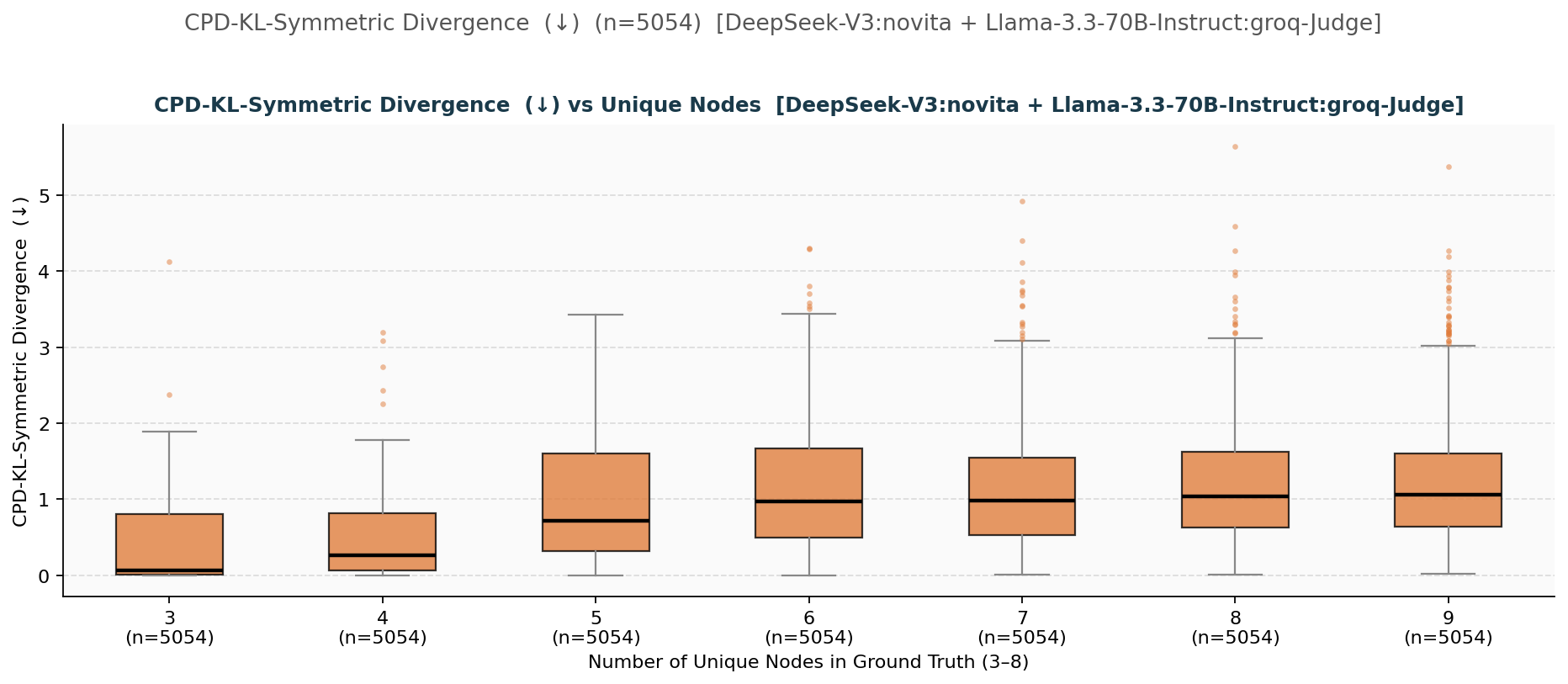}
  \caption*{DeepSeek-V3}\end{minipage}
\begin{minipage}{0.32\textwidth}\centering
  \includegraphics[width=\linewidth]{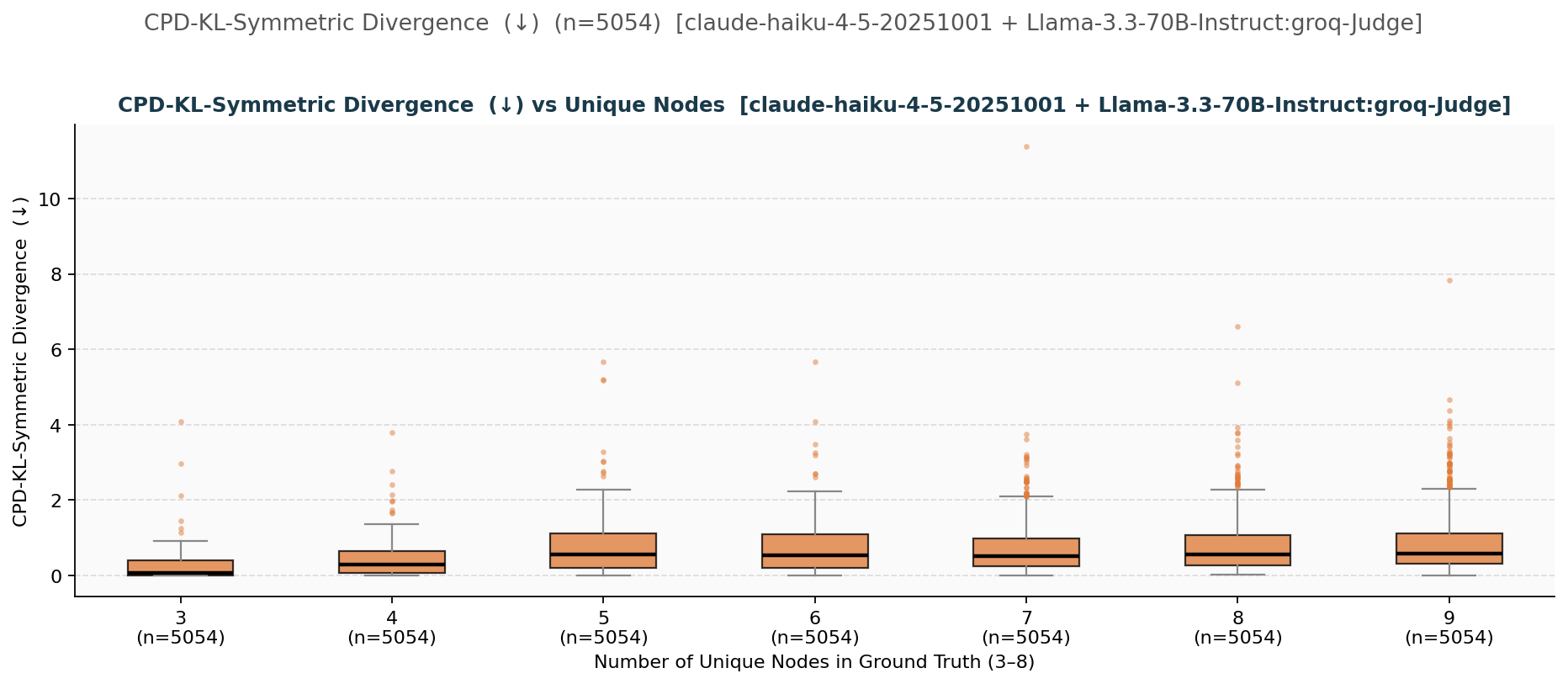}
  \caption*{Claude Haiku 4.5}\end{minipage}
\begin{minipage}{0.32\textwidth}\centering
  \includegraphics[width=\linewidth]{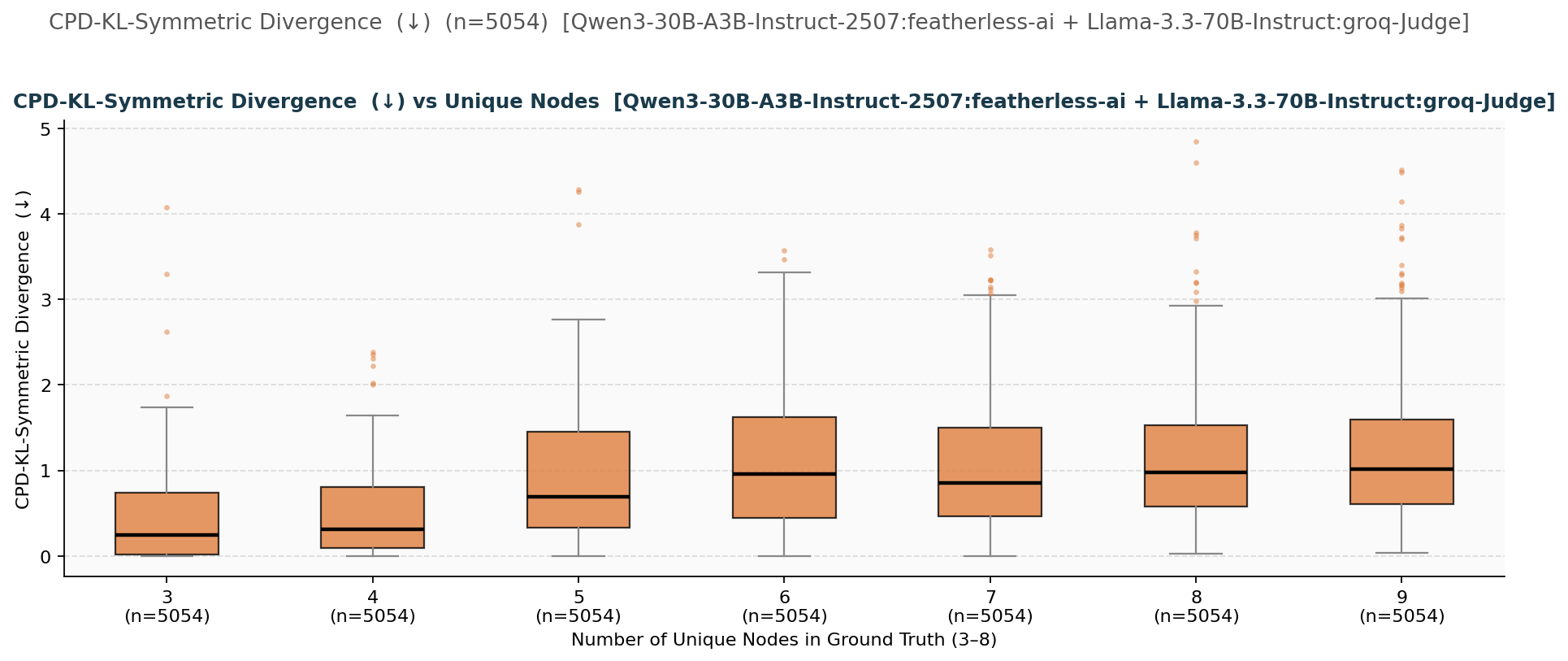}
  \caption*{Qwen3-30B-A3B}\end{minipage}\\[0.5em]
\begin{minipage}{0.32\textwidth}\centering
  \includegraphics[width=\linewidth]{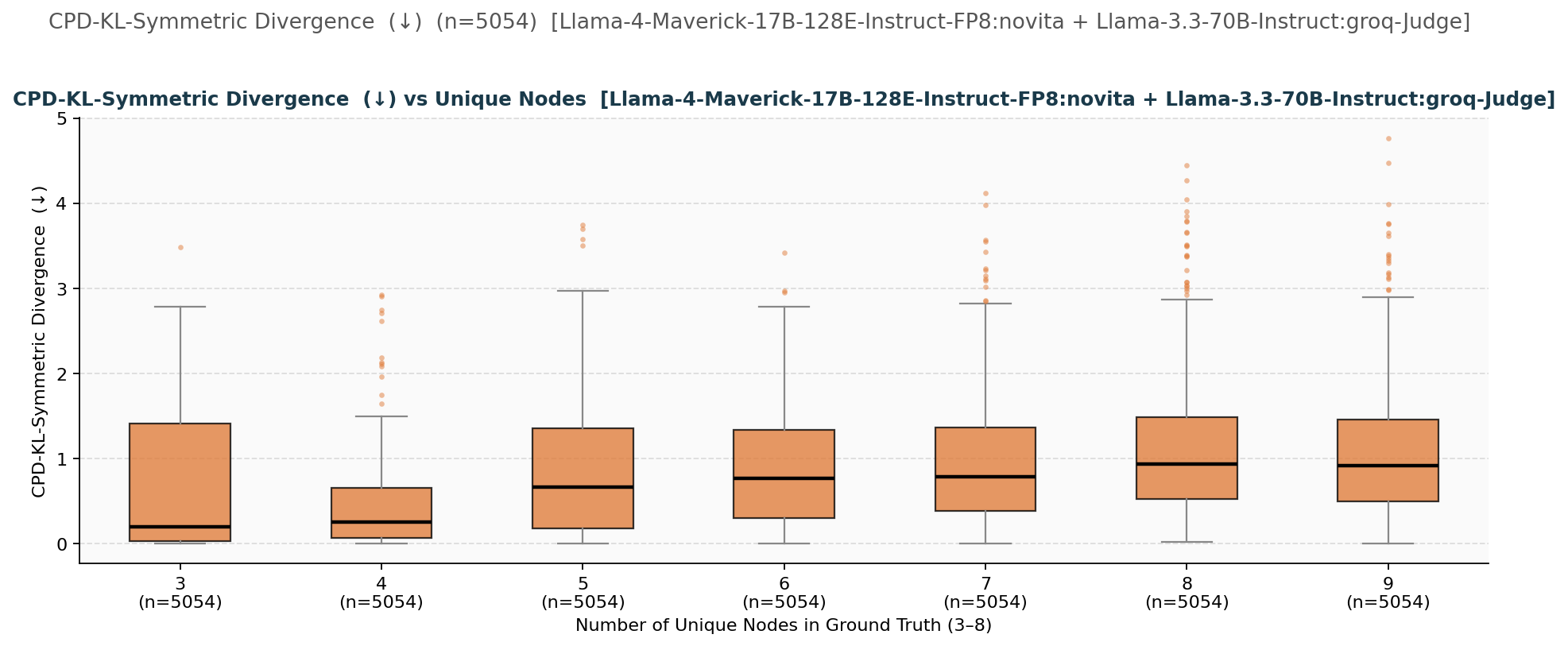}
  \caption*{Llama 4 Maverick}\end{minipage}
\begin{minipage}{0.32\textwidth}\centering
  \includegraphics[width=\linewidth]{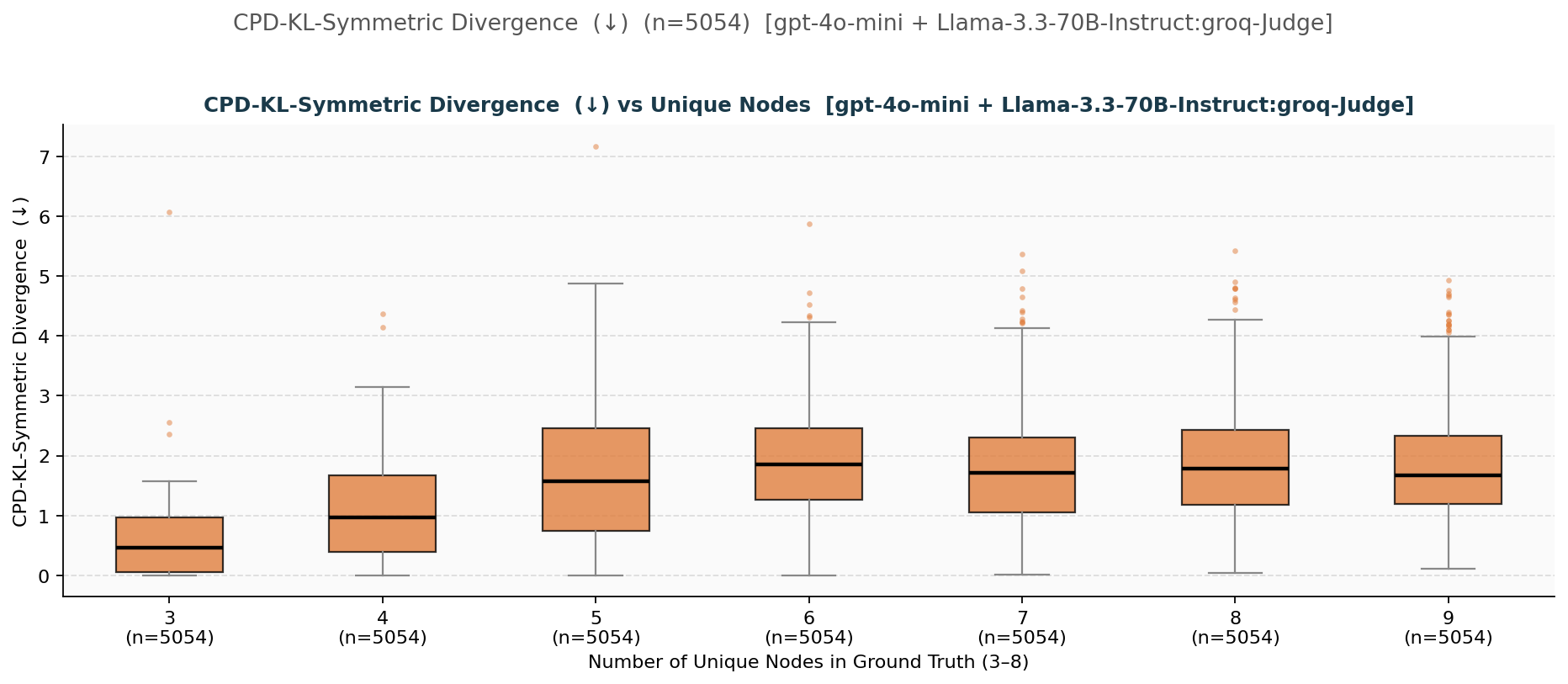}
  \caption*{GPT-4o-mini}\end{minipage}
\begin{minipage}{0.32\textwidth}\centering
  \includegraphics[width=\linewidth]{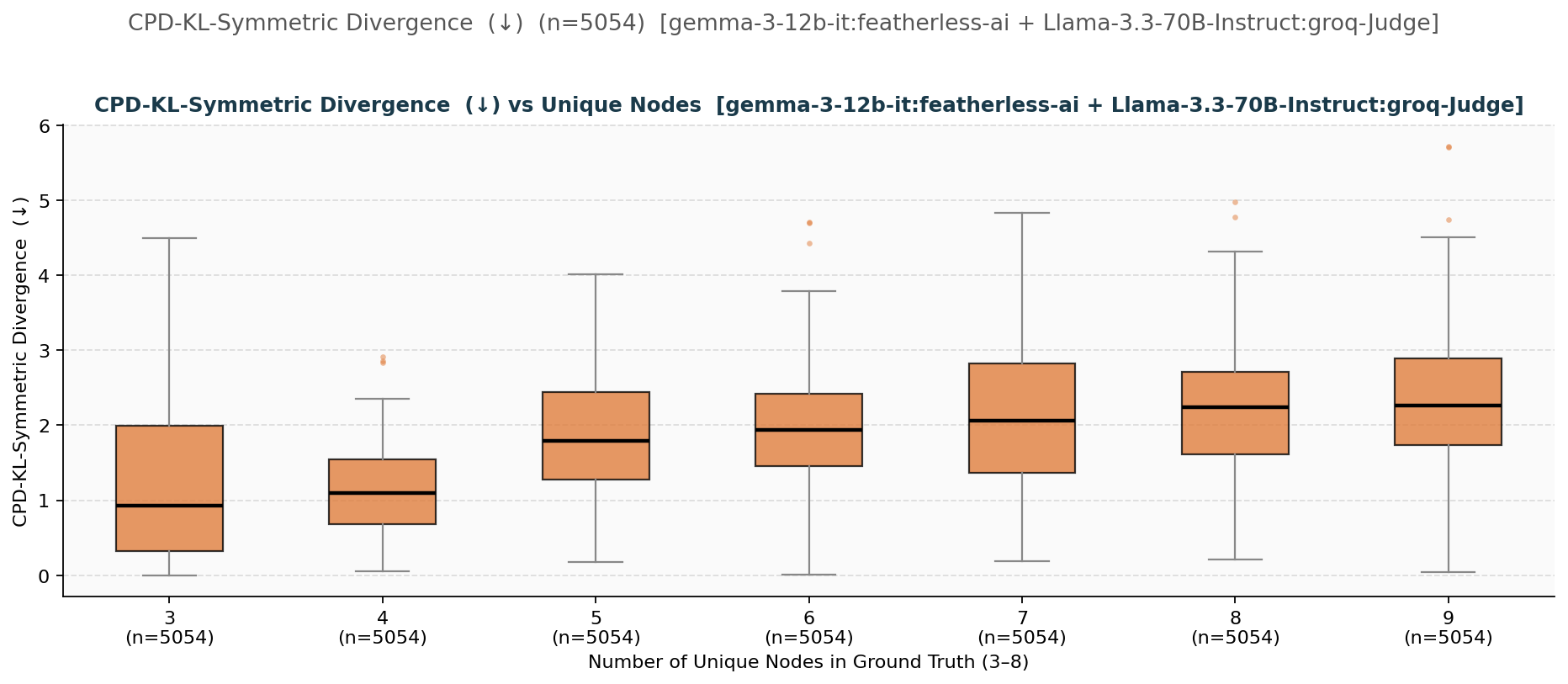}
  \caption*{Gemma 3 12B}\end{minipage}
\caption{Per-model symmetric CPD-KL divergence vs. reference node count (lower
is better). The patterns mirror the asymmetric CPD-KL plots, indicating that
model differences are driven by overall calibration rather than directional
bias.}
\label{fig:appendix_kl_sym}
\end{figure*}
\begin{table*}[t]
\centering
\caption{CPD-KL calibration references and LLM generators evaluated on the same
reference edge-CPDs. Lower is better. Calibration references make no use of
LLM-elicited conditionals.}
\label{tab:cpd-baselines}
\small
\setlength{\tabcolsep}{4pt}
\begin{tabular}{lcc}
\toprule
\textbf{Baseline / Model} & \textbf{Asym.\ KL} & \textbf{Sym.\ KL} \\
\midrule
\multicolumn{3}{l}{\textit{Calibration references}} \\
\quad Gaussian, $\sigma=0.1$        & $0.34 \pm 0.31$ & $0.23 \pm 0.18$ \\
\quad Marginal                      & $0.20 \pm 0.12$ & $0.30 \pm 0.39$ \\
\quad Uniform                       & $0.27 \pm 0.14$ & $0.40 \pm 0.47$ \\
\quad Dirichlet, $\alpha=1$         & $0.71 \pm 0.23$ & $0.77 \pm 0.50$ \\
\quad Gaussian, $\sigma=0.2$        & $1.05 \pm 0.74$ & $0.67 \pm 0.43$ \\
\quad Gaussian, $\sigma=0.3$        & $1.76 \pm 1.07$ & $1.11 \pm 0.62$ \\
\quad Gaussian, $\sigma=0.5$        & $2.81 \pm 1.47$ & $1.75 \pm 0.87$ \\
\midrule
\multicolumn{3}{l}{\textit{LLM generators}} \\
\quad Claude Haiku 4.5              & $1.11$ & $0.75$ \\
\quad Llama 4 Maverick              & $1.14$ & $1.00$ \\
\quad Qwen3-30B-A3B                 & $1.72$ & $1.08$ \\
\quad DeepSeek-V3                   & $1.52$ & $1.16$ \\
\quad GPT-4o-mini                   & $2.25$ & $1.80$ \\
\quad Gemma 3 12B                   & $3.14$ & $2.16$ \\
\bottomrule
\end{tabular}
\end{table*}
\section{CPD-KL Calibration Baselines}
\label{app:cpd-baselines}

The CPD-KL divergences reported in \S\ref{sec:results} range from 1.11 to 3.14 as shown in Table \ref{tab:cpd-baselines}
across the six evaluated generators. Because KL divergence has no universal
scale, this appendix anchors those numbers against four parameter-free
calibration references that share the PRISM-BN reference structure but make no
use of LLM-elicited conditionals. All calibration references are evaluated over
the same reference edge-CPDs used in the model evaluation.

 \begin{itemize}
    \item \textbf{Uniform.} Every column of the predicted CPD is set to the
    uniform distribution $1/|S_C|$ over child states.
    \item \textbf{Marginal.} Every column is set to the child's unconditional
    marginal $P(C{=}c)$, ignoring the parent entirely. This is a
    reference-structure-oracle, parent-ignorant predictor.
    \item \textbf{Gaussian-noise.} The reference CPD is perturbed entry-wise by
    additive Gaussian noise $\mathcal{N}(0,\sigma)$, clipped at 0, and
    column-renormalized for $\sigma\in\{0.1,0.2,0.3,0.5\}$.
    \item \textbf{Dirichlet random ($\alpha=1$).} Each column is sampled
    independently from $\mathrm{Dirichlet}(\boldsymbol{1})$ and averaged over
    three draws per edge.
\end{itemize}

\paragraph{Calibration analysis.}
All six evaluated LLMs have higher divergence than the uniform reference
predictor and the parent-ignorant marginal predictor. The marginal predictor
cannot represent how parent state shifts child distributions, so it is a
calibration reference rather than a substitute for conditional modelling. The
Gaussian-noise sweep gives a compact noise-equivalent scale, while the LLM
ranking remains useful for tracking progress within the zero-shot CPD
extraction setting.
\section{Per-Domain Breakdown of Benchmark Results}
\label{app:per-domain}

The aggregate metrics in \S\ref{sec:results} and Tables \ref{tab:per-domain-node-f1}, \ref{tab:per-domain-edge-f1} and \ref{tab:per-domain-cpd-kl} are computed over the full corpus,
in which Geopolitics and Environment together account for most examples. This
appendix recomputes Node $F_1$, Edge $F_1$, and CPD-KL within each domain.
Because Society and Technology contain fewer examples, their rows should be
interpreted with appropriate caution.

\begin{table*}[t]
\centering
\small
\caption{Per-domain Node $F_1$ $\uparrow$ for each generator. Bold marks the
best model in each column.}
\label{tab:per-domain-node-f1}
\begin{tabular}{lcccccc}
\toprule
\textbf{Model} & \textbf{Economics} & \textbf{Environment} & \textbf{Geopolitics} & \textbf{Society} & \textbf{Technology} & \textbf{ALL} \\
\midrule
DeepSeek-V3      & \textbf{0.818} & \textbf{0.829} & \textbf{0.816} & \textbf{0.838} & \textbf{0.885} & \textbf{0.826} \\
Claude Haiku 4.5 & 0.807 & 0.817 & 0.808 & 0.836 & 0.874 & 0.816 \\
Qwen3-30B-A3B    & 0.738 & 0.716 & 0.749 & 0.791 & 0.775 & 0.737 \\
Llama 4 Maverick & 0.707 & 0.590 & 0.541 & 0.738 & 0.757 & 0.618 \\
GPT-4o-mini      & 0.622 & 0.562 & 0.509 & 0.630 & 0.604 & 0.573 \\
Gemma 3 12B      & 0.620 & 0.535 & 0.486 & 0.631 & 0.640 & 0.566 \\
\bottomrule
\end{tabular}
\end{table*}

\begin{table*}[t]
\centering
\small
\caption{Per-domain Edge $F_1$ $\uparrow$ for each generator. Bold marks the
best model in each column.}
\label{tab:per-domain-edge-f1}
\begin{tabular}{lcccccc}
\toprule
\textbf{Model} & \textbf{Economics} & \textbf{Environment} & \textbf{Geopolitics} & \textbf{Society} & \textbf{Technology} & \textbf{ALL} \\
\midrule
DeepSeek-V3      & 0.952 & 0.973 & \textbf{0.961} & 0.947 & \textbf{0.973} & \textbf{0.966} \\
Claude Haiku 4.5 & 0.936 & \textbf{0.978} & 0.953 & \textbf{0.960} & 0.964 & 0.959 \\
Qwen3-30B-A3B    & 0.930 & 0.958 & 0.918 & 0.945 & 0.895 & 0.939 \\
Llama 4 Maverick & \textbf{0.961} & 0.967 & 0.944 & 0.957 & 0.933 & 0.960 \\
GPT-4o-mini      & 0.891 & 0.912 & 0.878 & 0.897 & 0.844 & 0.902 \\
Gemma 3 12B      & 0.944 & 0.925 & 0.915 & 0.938 & 0.968 & 0.932 \\
\bottomrule
\end{tabular}
\end{table*}

\begin{table*}[t]
\centering
\small
\caption{Per-domain CPD-KL $\downarrow$ for each generator. Bold marks the
lowest divergence in each column.}
\label{tab:per-domain-cpd-kl}
\begin{tabular}{lcccccc}
\toprule
\textbf{Model} & \textbf{Economics} & \textbf{Environment} & \textbf{Geopolitics} & \textbf{Society} & \textbf{Technology} & \textbf{ALL} \\
\midrule
DeepSeek-V3      & 2.052 & 1.281 & 1.460 & 2.263 & 0.907 & 1.517 \\
Claude Haiku 4.5 & 1.617 & \textbf{0.781} & \textbf{0.982} & \textbf{1.750} & \textbf{0.819} & \textbf{1.107} \\
Qwen3-30B-A3B    & 2.310 & 1.162 & 1.851 & 2.599 & 1.358 & 1.722 \\
Llama 4 Maverick & \textbf{1.408} & 0.945 & 1.052 & 2.009 & 1.407 & 1.143 \\
GPT-4o-mini      & 2.947 & 1.912 & 2.266 & 3.174 & 1.482 & 2.253 \\
Gemma 3 12B      & 3.864 & 2.724 & 3.136 & 4.183 & 2.257 & 3.139 \\
\bottomrule
\end{tabular}
\end{table*}
From Figures \ref{fig:appendix_node_f1}, \ref{fig:appendix_state_f1}  we can observe
the top structural ranking is stable across domains on Node $F_1$: DeepSeek-V3
leads in all five domains and Claude Haiku 4.5 is second in all five. Edge
$F_1$ shows minor reordering among the top models, with small absolute spreads.
In Figures \ref{fig:appendix_kl} and \ref{fig:appendix_kl_sym}, 
CPD-KL shows the largest domain effect, but the main conclusion holds in every
domain: the model that wins Node $F_1$ does not win CPD-KL. This reinforces the
need to evaluate text-to-BN systems on structure and CPDs separately.
\section{Non-Claude Reference Replication}
\label{app:gpt_ref_results}

Because the original reference BNs were generated using Claude Sonnet, the
performance of Claude Haiku could potentially be influenced by shared
model-family characteristics. To examine this possibility, we repeated the
evaluation using a separately generated GPT-5.5 reference set under the same
construction and evaluation protocol.

\begin{table*}[t]
\centering
\small
\caption{Evaluation using the independently generated GPT-5.5 reference set.
State and Edge $F_1$ are conditional metrics. Lower KL is better.}
\label{tab:gpt55-replication}
\begin{tabular}{lccccc}
\toprule
\textbf{Model} & \textbf{Node $F_1$} & \textbf{State $F_1$} & \textbf{Edge $F_1$} & \textbf{CPD-KL} & \textbf{Sym.\ KL} \\
\midrule
Claude Haiku 4.5 & \textbf{0.929} & \textbf{0.993} & \textbf{0.998} & \textbf{0.481} & \textbf{0.362} \\
DeepSeek-V3      & 0.908 & \textbf{0.993} & \textbf{0.998} & 1.133 & 0.734 \\
Llama 4 Maverick & 0.627 & 0.950 & 0.995 & 0.893 & 0.763 \\
Qwen3-30B-A3B    & 0.781 & 0.953 & 0.977 & 2.112 & 1.176 \\
GPT-4o-mini      & 0.550 & 0.964 & 0.962 & 1.552 & 1.245 \\
Gemma 3 12B      & 0.480 & 0.958 & 0.970 & 3.307 & 2.245 \\
\bottomrule
\end{tabular}
\end{table*}

Despite changes in absolute scores, the main relative trends remain broadly
consistent with the original evaluation. Claude Haiku and DeepSeek-V3 are the
two strongest models on Node $F_1$, and Claude Haiku remains the strongest CPD
extractor. This reduces, but does not eliminate, the concern that Claude
Haiku's performance is caused by sharing a model family with the original
reference generator.

Structural and probabilistic rankings are not identical. For example, Llama 4
Maverick has substantially lower Node $F_1$ than DeepSeek-V3
(0.627 versus 0.908) but better CPD-KL (0.893 versus 1.133). These results
indicate that the principal performance trends are robust across the two
reference-generation settings and that structural extraction quality does not
by itself determine probabilistic parameter quality.

\section{Human Validation and Semantic-Judge Audit}
\label{app:human}

We distinguish validation of the generated descriptions from validation of the
semantic judge used during evaluation. The former tests whether the descriptions
satisfy the intended generation constraints, whereas the latter tests whether
predicted and reference node or state names are aligned correctly.

\paragraph{Description-Coverage Audit}

The existing manual audit examines a random sample of 344 text--BN pairs and
checks whether each generated description covers the intended nodes, non-null
states, and directed edges. This audit validates the coverage constraints used
during description generation. It does not validate the Llama~3.3~70B semantic
judge used to align predicted and reference node and state names during model
evaluation.

\subsection{Human Extraction Pilot}

To assess whether the generated descriptions are interpretable by humans, two
annotators independently reconstructed ten BNs from their descriptions. The
annotators extracted the nodes, states, directed edges, and probabilistic
parameters without access to the corresponding reference BNs. Their outputs
were evaluated using the same scoring convention as the model predictions.
\begin{table*}[t]
\centering
\small
\setlength{\tabcolsep}{7pt}
\caption{Human extraction results on ten independently reconstructed BNs.
State and Edge $F_1$ are conditional on successful node alignment. Lower
full-CPD KL is better.}
\label{tab:human-extraction}
\begin{tabular}{lcccc}
\toprule
\textbf{Annotator}
& \textbf{Node $F_1$}
& \textbf{State $F_1$}
& \textbf{Edge $F_1$}
& \textbf{Full-CPD KL} \\
\midrule
Annotator 1 & 0.92 & 0.95 & 0.95 & 0.37 \\
Annotator 2 & 0.95 & 0.98 & 0.95 & 0.53 \\
\bottomrule
\end{tabular}
\end{table*}

Pairwise annotator agreement was 96\% for nodes, 93\% for states, and 98\% for
edges. Mean inter-annotator full-CPD KL was 0.14 under the same scoring
convention.

\end{document}